\documentclass[10pt,reqno]{amsart}
\usepackage{amsaddr}
\usepackage[a4paper, total={6.5in, 8.4in}]{geometry}

\usepackage[symbol]{footmisc}
\usepackage{amsmath}
\usepackage{amssymb}
\usepackage{amsthm}
\usepackage{aliascnt}
\usepackage{eurosym}
\usepackage{graphicx}
\usepackage{hyperref}
\usepackage{cleveref}
\usepackage{dsfont}
\usepackage[space]{cite}
\usepackage{caption}
\usepackage{subcaption}
\usepackage{placeins}
\usepackage{mathtools}
\usepackage{xcolor}
\usepackage[title]{appendix}

\usepackage{algorithm}
\usepackage{algorithmic}

\usepackage[displaymath,mathlines]{lineno}
\modulolinenumbers[1]

\allowdisplaybreaks

\usepackage{ifthen}
\usepackage{comment}

\makeindex
\usepackage{float}

\renewcommand{\div}{\operatorname{div}}

\newcommand{\dd}{\mathrm{d}}
\newcommand{\dif}{\mathop{}\!\mathrm{d}}
\def\leq{\leqslant}
\def\geq{\geqslant}

\numberwithin{equation}{section}

\newtheoremstyle{thmlemcorr}{10pt}{10pt}{\itshape}{}{\bfseries}{.}{10pt}{{\thmname{#1}\thmnumber{
			#2}\thmnote{ (#3)}}}
\newtheoremstyle{thmlemcorr*}{10pt}{10pt}{\itshape}{}{\bfseries}{.}\newline{{\thmname{#1}\thmnumber{
			#2}\thmnote{ (#3)}}}
\newtheoremstyle{defi}{10pt}{10pt}{\itshape}{}{\bfseries}{.}{10pt}{{\thmname{#1}\thmnumber{
			#2}\thmnote{ (#3)}}}
\newtheoremstyle{remexample}{10pt}{10pt}{}{}{\bfseries}{.}{10pt}{{\thmname{#1}\thmnumber{
			#2}\thmnote{ (#3)}}}
\newtheoremstyle{ass}{10pt}{10pt}{}{}{\bfseries}{.}{10pt}{{\thmname{#1}\thmnumber{
			A#2}\thmnote{ (#3)}}}
\usepackage{tikz}
\usetikzlibrary{shadows, positioning, calc, backgrounds, shapes.geometric, arrows.meta, fit}
\theoremstyle{thmlemcorr}

\numberwithin{theorem}{section}

\newaliascnt{lemma}{theorem}

\aliascntresetthe{lemma}
\newaliascnt{corollary}{theorem}

\aliascntresetthe{corollary}
\newaliascnt{proposition}{theorem}

\aliascntresetthe{proposition}
\newaliascnt{conjecture}{theorem}

\aliascntresetthe{conjecture}
\usepackage{booktabs}
\theoremstyle{thmlemcorr*}
\newtheorem{theorem*}{Theorem}
\newtheorem{lemma*}[theorem]{Lemma}
\newtheorem{corollary*}[theorem]{Corollary}
\newtheorem{proposition*}[theorem]{Proposition}
\newtheorem{problem*}[theorem]{Problem}
\newtheorem{conjecture*}[theorem]{Conjecture}

\theoremstyle{defi}
\newaliascnt{definition}{theorem}

\aliascntresetthe{definition}

\newtheorem{problem}{Problem}

\theoremstyle{remexample}

\newtheoremstyle{itremark}%
  {\topsep}{\topsep}% space above/below
  {\itshape}% body font (ITALIC)
  {}% indent
  {\bfseries}% head font
  {.}% head punctuation
  {.5em}% space after head
  {}% head spec

\theoremstyle{itremark}

\newaliascnt{remarknoqed}{theorem}
\newtheorem{remarknoqed}[remarknoqed]{Remark}
\aliascntresetthe{remarknoqed}

\newenvironment{remark}
  {\begin{remarknoqed}}
  {\qed\end{remarknoqed}}

\newaliascnt{example}{theorem}

\aliascntresetthe{example}
\newaliascnt{counterexample}{theorem}

\aliascntresetthe{counterexample}
\definecolor{revision_color}{RGB}{0,128,0}
\definecolor{revision_color1}{RGB}{0,128,0}
\definecolor{revision_color2}{RGB}{255,165,0}
\definecolor{mathfixgreen}{RGB}{0,128,0}
\newcommand{\mathfix}[1]{{#1}}

\newaliascnt{teo}{theorem}

\aliascntresetthe{teo}
\newaliascnt{lem}{theorem}

\aliascntresetthe{lem}
\newaliascnt{pro}{theorem}

\aliascntresetthe{pro}
\newaliascnt{cor}{theorem}

\aliascntresetthe{cor}

\crefname{theorem}{Theorem}{Theorems}
\Crefname{theorem}{Theorem}{Theorems}
\crefname{lemma}{Lemma}{Lemmas}
\Crefname{lemma}{Lemma}{Lemmas}
\crefname{corollary}{Corollary}{Corollaries}
\Crefname{corollary}{Corollary}{Corollaries}
\crefname{proposition}{Proposition}{Propositions}
\Crefname{proposition}{Proposition}{Propositions}
\crefname{conjecture}{Conjecture}{Conjectures}
\Crefname{conjecture}{Conjecture}{Conjectures}
\crefname{definition}{Definition}{Definitions}
\Crefname{definition}{Definition}{Definitions}
\crefname{remarknoqed}{Remark}{Remarks}
\Crefname{remarknoqed}{Remark}{Remarks}
\crefname{example}{Example}{Examples}
\Crefname{example}{Example}{Examples}
\crefname{counterexample}{Counterexample}{Counterexamples}
\Crefname{counterexample}{Counterexample}{Counterexamples}
\crefname{teo}{Theorem}{Theorems}
\Crefname{teo}{Theorem}{Theorems}
\crefname{lem}{Lemma}{Lemmas}
\Crefname{lem}{Lemma}{Lemmas}
\crefname{pro}{Proposition}{Propositions}
\Crefname{pro}{Proposition}{Propositions}
\crefname{cor}{Corollary}{Corollaries}
\Crefname{cor}{Corollary}{Corollaries}

\newenvironment{notations}{
    \noindent\textbf{Notations:} \it % Bold title and italicized text
}{
    \normalfont % Return to normal font
}
\theoremstyle{ass}

\newtheorem*{notations*}{Notations}

\title[Gaussian Process Policy Iteration with Additive Schwarz Acceleration]{Gaussian Process Policy Iteration with Additive Schwarz Acceleration for Forward and Inverse HJB and Mean Field Game Problems}

\author{Xianjin Yang$^{1,*, \dagger}$, Jingguo Zhang$^{2,\dagger}$}
\thanks{$^*$Corresponding author. $^\dagger$Authors are listed alphabetically and contributed equally.}

\address{$^1$Department of Computing and Mathematical Sciences, California Institute of Technology, CA, USA.}
\email{yxjmath@caltech.edu}
\address[J. Zhang]{
	$^2$Department of Mathematics and Risk Management Institute, National University of Singapore, Singapore.}
\email{e0983423@u.nus.edu}

\begin{document}

\begin{abstract}
In this paper, we propose a Gaussian Process (GP)-based policy iteration framework for addressing both forward and inverse problems in Hamilton--Jacobi--Bellman (HJB) equations and mean field games (MFGs). Policy iteration is formulated as an alternating procedure between evaluating the value function under a fixed control policy and improving the policy. In our approach, we model the unknown fields using GPs within \mathfix{a policy-iteration} framework that converts the nonlinear system into a sequence of linear PDE subproblems. Then, leveraging the linear structure, the updates for the value function and, in the MFG setting, the population density admit \mathfix{explicit representer formulas} under linear PDE collocation constraints. The policy is subsequently updated pointwise via a Legendre transform step, which involves a low-dimensional maximization over the control variable. \mathfix{This maximization is explicit for standard quadratic costs. For smooth, strictly convex costs, this pointwise maximization is solved through its first-order optimality condition, whereas in constrained or non-smooth cases, it becomes a low-dimensional constrained maximization problem.}  To improve convergence, we incorporate the additive Schwarz acceleration as a preconditioning step following each policy update.   Numerical experiments demonstrate the effectiveness of the Schwarz acceleration in improving computational efficiency.

\end{abstract}

\maketitle
\noindent\textbf{Key words:} Gaussian processes, HJB equations, mean field games, inverse problems, policy iteration, additive Schwarz acceleration
\section{Introduction}
\label{Introduction}
Optimal control problems involve designing a feedback law that minimizes a cumulative cost over a prescribed time horizon. Such problems are rigorously \mathfix{characterized} by the {Hamilton--Jacobi--Bellman} (HJB) equation, a nonlinear partial differential equation (PDE) that governs the value function of a single decision maker. When a large population of agents interacts, and each agent optimizes its own cost in response to the aggregate behavior of the group, \mathfix{mean field game (MFG) theory describes the corresponding continuum limit} \cite{lasry2006jeux1,  lasry2006jeux, lasry2007mean, huang2006large, huang2007large, huang2007nash, huang2007invariance}. A typical MFG consists of a coupled PDE system: a backward \mathfix{HJB} equation for the representative agent's value function and a forward Fokker--Planck (FP) equation for the evolution of the population density. These frameworks arise in fields ranging from robotics and economics to crowd dynamics and epidemiology \cite{gueant2011mean, gomes2015economic, lee2020mean, gao2021belief, gomes2021mean, evangelista2018first, lee2022mean}. In this paper, we propose a Gaussian Process Policy Iteration (GPPI) framework to solve both forward and inverse problems of HJB equations and MFGs. \mathfix{To accelerate convergence, we incorporate the additive Schwarz Newton method, which significantly reduces the number of iterations required.}

In the {forward} HJB/MFG problem, the model parameters (dynamics, cost functions, coupling terms) are assumed known, and the task is to compute the corresponding solution of the HJB or MFG system. Conversely, the inverse problem focuses on inferring unknown model components, such as spatial cost functions or interaction terms, from partial observations of optimal trajectories or population densities.  Inverse formulations are essential for {data-driven} calibration, enabling the recovery of hidden objectives or environmental parameters that explain observed behavior.

For clarity of exposition, we introduce the prototypical time-dependent HJB and MFG systems studied in this paper.

\subsection{Stochastic Optimal Control and HJB Equations}  Denote $\Omega=\mathbb{T}^d:=(\mathbb{R}/\mathbb{Z})^d$, the $d$-dimensional flat torus. We consider a finite-horizon stochastic optimal control problem on the time interval $[0,T]$. Let 
$x(\cdot)\colon [0,T]\to {\Omega}$ 
be a stochastic process governed by the controlled stochastic differential equation (SDE):
\[
\mathrm{d}x(s) = f\bigl(x(s), s, \boldsymbol{q}(x(s), s)\bigr)\,\mathrm{d}s + \sigma(s)\,\mathrm{d}W_s,
\qquad s\in(t,T], \qquad x(t)=x,
\]
where \( \boldsymbol{q} \colon \Omega \times [0, T] \to \mathcal{Q} \) is an admissible control with values in a compact set \( \mathcal{Q} \), \( W_s \) is a standard \( d \)-dimensional Brownian motion, \(f : \Omega \times [0,T] \times \mathcal{Q} \to \mathbb{R}^d\) represents the drift term, and \(\sigma : [0,T]\to \mathbb{R}\) represents the scalar diffusion coefficient. Here, $x(s)$ is understood modulo $1$ in each coordinate (equivalently, the coefficients are periodic on $\mathbb{R}^d$). The cost functional is defined by
\[
J\bigl(x,t; \boldsymbol{q}\bigr)
= \mathbb{E} \left[ \int_{t}^{T} \ell\bigl(x(s),s,\boldsymbol{q}(x(s),s)\bigr)\,\mathrm{d}s + \mathfix{U_T}\bigl(x(T)\bigr) \right],
\]
{where \(\ell\colon \Omega\times[0,T] \times \mathcal{Q} \to \mathbb{R} \) is the running cost, and \( \mathfix{U_T}\colon \Omega \to \mathbb{R} \) is the terminal cost.}
The {value function}
\[
u(x,t) = \inf_{\substack{\boldsymbol{q} \colon \Omega \times [0, T] \to \mathcal{Q}}} J(x,t; \boldsymbol{q}),
\]
represents the minimal expected cost-to-go starting from state \( x \) at time \( t \). By the dynamic programming principle, the value function \( u \) satisfies the time-dependent HJB equation:
\begin{equation}\label{eq:StochHJB}
-\partial_t u(x,t) - \frac{1}{2}   \sigma(t)^2 \Delta u(x,t)+ H\bigl(x,t, \nabla u(x,t)\bigr)  = 0, \quad
\ u(x,T) = \mathfix{U_T}(x),
\end{equation}
where the Hamiltonian $H$ is given by $
H(x,t,p) = \sup_{\boldsymbol{q} \in \mathcal{Q}} \left\{ -p^T f(x,t,\boldsymbol{q}) 
 - \ell(x,t,\boldsymbol{q}) \right\}$. 
The forward problem associated with \eqref{eq:StochHJB} is to solve for the value function \(u\) and to recover the optimal feedback control via
\[
\boldsymbol{q}^*(x,t) = \arg\max_{\boldsymbol{q} \in \mathcal{Q}} \left\{ -\nabla u(x,t)^T f(x,t,\boldsymbol{q}) - \ell(x,t,\boldsymbol{q}) \right\}.
\]
\mathfix{In the quadratic examples considered below, this maximizer is unique.}
The inverse HJB problem aims to identify unknown components in the dynamics \( f \), running cost \( \ell \), or terminal cost \( \mathfix{U_T} \), based on partial observations of optimal trajectories or the value function.

For the forward HJB problem, several classes of methods have been developed: finite-difference and high-order ENO/WENO schemes \cite{zhang2003high,osher1991high}, semi-Lagrangian discretizations \cite{falcone2013semi}, policy iteration algorithms \cite{howard1960dynamic,alla2015efficient}, spectral collocation approaches \cite{fornberg1998practical, boyd2001chebyshev}, and physics-informed neural networks \cite{raissi2019physics, beck2019machine}. 
The inverse HJB problem, recovering unknown running or terminal cost functions from observed optimal trajectories, has been studied extensively.  We refer readers to \cite{kalman1964linear,esteve2020inverse,golubyatnikov1995inverse}.

\subsection{Mean Field Games}
In the large population limit, \mathfix{MFG theory} \cite{lasry2006jeux1, lasry2006jeux,  lasry2007mean, huang2006large, huang2007large, huang2007nash, huang2007invariance} simplifies the interaction structure by allowing a representative agent to react to the aggregate behavior of the population, rather than modeling pairwise interactions with individual agents. The Nash equilibrium in MFGs can be \mathfix{computed via} an iterative process. First, \mathfix{for a fixed population density $m$,} the representative agent seeks an optimal control strategy by solving the associated mean-field control problem. Then, under this optimal strategy, the distribution of agents evolves and is required to match the prescribed density. More precisely, the representative agent solves
\[
\inf_{\substack{\mathfix{\boldsymbol q} \colon \Omega \times [0, T] \to \mathcal{Q}}} 
\mathbb{E} \left[ \int_{t}^{T} \Bigl[\ell\bigl(x(s),s,\boldsymbol{q}(x(s),s)\bigr) + F\bigl(m(x(s),s)\bigr)\Bigr]\,\mathrm{d}s
+ U_T(x(T))\right],
\]
subject to the controlled stochastic dynamics
$
\mathrm{d}x(s) = f\bigl(x(s),s,\mathfix{\boldsymbol q}(x(s),s)\bigr)\,\mathrm{d}s + \sigma(s)\,\mathrm{d}W_s, $ and $ x(t)=x.
$
{Here, $\Omega=\mathbb{T}^d:=(\mathbb{R}/\mathbb{Z})^d$ is the $d$-dimensional flat torus and $\ell: \Omega\times[0,T]\times\mathcal{Q}\to\mathbb{R}$ represents the running cost.} 
The state process is a trajectory $x(\cdot) : [t, T] \to \Omega$,  with the control \(\mathfix{\boldsymbol q}\) taking values in a given compact set \(\mathcal{Q}\). 
The function \( F \colon \mathbb{R}_{+} \to \mathbb{R} \) characterizes the scalar mean-field coupling appearing in the running cost, while \( U_T \colon \Omega  \to \mathbb{R} \) describes the terminal cost.

The corresponding value function
\[
u(x,t) = \inf_{\substack{\mathfix{\boldsymbol q} \colon \Omega \times [0, T] \to \mathcal{Q}}} 
\mathbb{E} \left[ \int_{t}^{T} \Bigl[\ell\bigl(x(s),s,\boldsymbol{q}(x(s),s)\bigr) + F\bigl(m(x(s), s)\bigr)\Bigr]\,\mathrm{d}s
+ U_T(x(T)) \right]
\]
satisfies a backward HJB equation coupled with \( m \). Consistency requires that the actual density \( m \) evolve under the optimal feedback control, resulting in a forward FP equation whose solution matches the optimal density prescribed from the first step. Combining these yields the time-dependent MFG system:
\begin{equation}
\label{eq:MFG}
\begin{cases}
-\,\partial_{t}u(x,t) - \frac{1}{2}\sigma(t)^2\Delta u(x,t) + H\bigl(x,t,\nabla u(x,t)\bigr)
  = F\bigl(m(x,t)\bigr), &
 u(x,T)=U_T(x),
\\[0.8ex]\partial_{t}m(x,t) -\frac{1}{2}\sigma(t)^2\Delta m(x,t)
  - \div\bigl(m(x,t)\,D_{p}H(x,t,\nabla u(x,t))\bigr)
= 0,
& m(x,0)=m_{0}(x)
.
\end{cases}
\end{equation}
Here, \( H(x,t,p) = \sup_{\boldsymbol{q} \in \mathcal{Q}} \left\{ -p^T f(x,t, \boldsymbol{q})  - \ell(x,t,\boldsymbol{q}) \right\} \), and \( D_{p}H \) denotes its gradient with respect to \( p \).
  The {forward MFG problem} is to solve \eqref{eq:MFG} for $(u,m)$ given $(f,\ell,F,U_T,m_{0})$.  MFG systems generally lack closed-form solutions. Hence, numerical approximation is indispensable.  A variety of computational schemes have been developed for the forward MFG problem, including finite-difference discretizations \cite{achdou2010mean,achdou2012iterative, gomes2020hessian}, Fourier spectral methods \cite{nurbekyan2019fourier}, splitting schemes \cite{liu2020splitting,liu2021computational}, proximal approaches \cite{briceno2018proximal,briceno2019implementation}, and GP-based approximations \cite{meng2023sparse, mou2022numerical}.  Data-driven solvers based on neural networks \cite{ruthotto2020machine,lin2020apac} and convergence analyses for diffusion-type MFGs \cite{carmona2019convergence,carmona2021convergence} have further enriched the toolkit. 
  
  {
  Inverse MFG problems seek to recover the full model configuration, including agent strategies, spatial costs, and population density, from partial observations of the corresponding system states and environmental parameters. We refer readers to \cite{liu2023inverse,klibanov2023convexification,klibanov2023holder,imanuvilov2023lipschitz,ren2024reconstructing,ren2025unique} for theoretical studies of inverse MFG problems. On the computational side, practical recovery schemes include variational methods \cite{ding2022mean}, policy iteration methods \cite{ren2024policy}, bilevel optimization formulations \cite{yu2024bilevel,huang2025joint}, operator learning approaches \cite{yang2023context,huang2025unsupervised}, and GP-based strategies \cite{guo2024decoding,zhang2025learning}. For the inverse problem, \cite{ren2024policy} proposes a policy-iteration scheme that reconstructs the obstacle function from observations of the value function, and proves a linear rate of convergence to a solution. Our approach is also based on the policy-iteration framework but differs in both methodology and data. Methodologically, we cast the forward and inverse steps in a GP formulation and accelerate the iteration with an additive Schwarz Newton method, whereas they use grid-based PDE solvers. In terms of data, our experiments use direct observations of the spatial cost together with the density, rather than value-function data.}

Policy iteration (PI) is a classical method for solving HJB equations. The algorithm \mathfix{was first formulated} in \cite{howard1960dynamic} for Markov decision processes and later extended to continuous-time control in \cite{bellman1966dynamic}.
In each iteration of PI, one alternates between policy evaluation (solving a linear PDE under a fixed control law) and policy improvement (updating the control) based on the current value function. Depending on the problem structure, PI may exhibit linear, superlinear, or even quadratic local convergence. For PDE-based settings relevant here, we refer in particular to \cite{cacace2021policy,lauriere2023policy,tang2024learning}. 
Another line of work \cite{assouli2024deep} models the unknown functions with deep neural networks and minimizes a composite PDE-residual objective. These approaches are quite flexible, but they do not leverage the linear structure in each policy-evaluation step and therefore do not offer closed-form updates.

GPs have been successfully applied to learning and solving ordinary differential equations (ODEs) \cite{hamzi2023learning,yang2024learning} and PDEs \cite{chen2021solving,chen2025sparse,yang2023mini,raissi2017machine,raissi2018numerical}.  In the MFG context, GP-based methods have been developed both for forward MFG systems \cite{mou2022numerical,meng2023sparse} and for inverse MFG problems \cite{guo2024decoding,zhang2025learning}. In this paper, we propose a GPPI algorithm with additive Schwarz Newton acceleration to address both forward and inverse HJB/MFG problems within a common \mathfix{GP} computational framework.  GPPI replaces the traditional grid-based representation of the value function with a GP surrogate. During each policy evaluation step, we sample the associated PDE at a selected set of points, fit a GP model to these samples, and thereby construct an explicit approximation of the value function. Policy improvement is then carried out by \mathfix{a pointwise Legendre update}, using the GP surrogate
to represent the value function and its derivatives. \mathfix{For standard quadratic costs, this update admits a closed-form expression. More generally, it can be computed by solving a low-dimensional constrained optimization problem.}  To accelerate the GPPI method, we incorporate an additive Schwarz Newton correction within the framework. \mathfix{This approach significantly reduces the number of iterations required for convergence without compromising accuracy.}  
Our framework differs from recent GP-based methods for MFGs in both problem formulation and algorithmic architecture.
First, unlike variational formulations that rely on a potential structure \cite{zhang2025learning}, our approach works directly with the coupled MFG PDE system and does not require rewriting the MFG as an optimization problem.
Second, and most importantly, our method differs from prior GP-based approaches (e.g., \cite{mou2022numerical, meng2023sparse, guo2024decoding}) in the choice of linearization strategy.
Existing works typically apply a Gauss--Newton linearization to the coupled system, which leads to a large, dense linear system that couples the linearized PDEs for $u$ and $m$ at every iteration.
In contrast, our method uses policy iteration, which freezes the policy and separates the update into structured linear subproblems.
In this setting, each policy evaluation step treats only one linear PDE at a time, either for $m$ or for $u$, instead of the full coupled MFG system.  This leads to a lower-dimensional computation.

Our main contributions are as follows:
\begin{itemize}
  {
  \item We introduce a \mathfix{GPPI} framework that provides a unified GP computational framework for forward and inverse problems arising from HJB equations and MFGs. 
  }
  \item Building on the GPPI framework, we propose the Gaussian process policy iteration with the additive Schwarz (GPPI-AS) algorithm, which leverages the additive Schwarz Newton correction to accelerate the GPPI iterations in practice for both forward problems and inverse reconstructions of HJBs and MFGs.
\end{itemize}

\mathfix{In this paper, the mesh-free property of GP regression allows the PDEs to be solved on nonuniform points, without being tied to a predefined computational grid.}  To validate the framework, we test it on several numerical examples involving HJB equations as well as stationary and time-dependent MFGs.

\subsection{Outline}
The remainder of the paper is organized as follows.  Section~\ref{sub:GPR:LM} reviews the fundamentals of \mathfix{GP regression}.  Section~\ref{sec:gppi_frameworks} introduces the GPPI frameworks for both forward and inverse problems of HJBs and MFGs.  Section~\ref{Newton Acceleration} presents the additive Schwarz Newton acceleration strategy to improve convergence within the GPPI algorithm.  
Section~\ref{secNumericalExpe} reports numerical experiments that demonstrate the efficiency and accuracy of the proposed methods. Section~\ref{sec:Conclusion} concludes with a discussion of our findings and outlines directions for future work. Appendix~\ref{AppendixA}  provides derivation details for the GPPI subproblems.

\begin{notations}
A real-valued vector \(\boldsymbol{v}\) is shown in boldface, except when representing a point in the physical domain. Its Euclidean norm is \(|\boldsymbol{v}|\), its transpose is \(\boldsymbol{v}^T\), and \(v_i\) denotes its \(i^{\textit{th}}\) component. \mathfix{For a scalar function \(f\) and a vector \(\boldsymbol{v}\), the notation \(f(\boldsymbol{v})\) is understood componentwise, that is, it denotes the vector \((f(v_1), \dots, f(v_N))\), where \(N\) is the length of \(\boldsymbol{v}\).} The Dirac delta function at \(x\) is denoted \(\delta_x\).

The multivariate normal distribution with covariance \(\gamma^2 I\) is written as \(\mathcal{N}(0, \gamma^2 I)\), where \(\gamma > 0\). For a normed vector space \(X\), its norm is \(\|\cdot\|_X\).

{
Let \(\mathcal{U}\) be a \mathfix{reproducing kernel} Hilbert space (RKHS) with norm
\(\|\cdot\|_{\mathcal{U}}\), and dual space \(\mathcal{U}^*\), with duality
pairing \([\cdot,\cdot]:\mathcal{U}^*\times\mathcal{U}\to\mathbb{R}\).
We assume the existence of a linear, bijective, symmetric, and positive
covariance operator \(\mathcal{K}_{\mathcal{U}}:\mathcal{U}^*\to\mathcal{U}\),
satisfying
\[
[\psi,\mathcal{K}_{\mathcal{U}}\phi]
=
[\phi,\mathcal{K}_{\mathcal{U}}\psi],
\qquad
[\phi,\mathcal{K}_{\mathcal{U}}\phi]>0
\quad \text{for all } \phi\neq 0 .
\]
The norm is given by
\[
\|u\|_{\mathcal{U}}^2
=
[\mathcal{K}_{\mathcal{U}}^{-1}u,u],
\qquad \forall u\in\mathcal{U}.
\]

 }
{
For \(\boldsymbol{\phi} = (\phi_1, \dots, \phi_P) \in {(\mathcal{U}^*)}^{P}\), we define $
[\boldsymbol{\phi}, u] := \left([\phi_1, u], \dots, [\phi_P, u]\right).$
}
Finally, for a collection of vectors \((\boldsymbol{v}_i)_{i=1}^{N_v}\), we denote by \((\boldsymbol{v}_1;\dots;\boldsymbol{v}_{N_v})\) their vertical concatenation.

\end{notations}

\section{Prerequisites for GP Regression}
\label{sub:GPR:LM}
In this section, we discuss function regression using GPs. 
{
\mathfix{Let $\Omega$ be an open subset of $\mathbb{R}^d$ or a compact smooth manifold such as $\mathbb{T}^d$, and let $\boldsymbol{f}: \Omega \to \mathbb{R}^m$ be a vector-valued random field. We call $\boldsymbol{f}$ a vector-valued GP if, for every finite collection $x_1,\dots,x_N \in \Omega$, the stacked random vector $\bigl(\boldsymbol{f}(x_1)^T,\dots,\boldsymbol{f}(x_N)^T\bigr)^T \in \mathbb{R}^{mN}$ is jointly Gaussian. The GP $\boldsymbol{f}$ is characterized by a mean function $\boldsymbol{\mu}: \Omega \to \mathbb{R}^m$ and a matrix-valued covariance kernel $K: \Omega \times \Omega \to \mathbb{R}^{m \times m}$ satisfying $\mathbb{E}(\boldsymbol{f}(x)) = \boldsymbol{\mu}(x)$ and $\operatorname{Cov}(\boldsymbol{f}(x), \boldsymbol{f}(x')) = K(x, x')$ for all $x, x' \in \Omega$. We write $\boldsymbol{f} \sim \mathcal{GP}(\boldsymbol{\mu}, K)$.}
}

The goal of learning vector-valued functions is to generate a GP estimator $\boldsymbol{f}^\dagger$ from a training set $(x_i, \boldsymbol{Y}_i)_{i=1}^N$, where each $\boldsymbol{Y}_i\in \mathbb{R}^m$. Assuming $\boldsymbol{f} \sim \mathcal{GP}(\boldsymbol{0}, K)$, we define $\boldsymbol{f}^\dagger$ as the mean of the posterior distribution of $\boldsymbol{f}$ conditional on the training data, i.e., 
{
$\boldsymbol{f}^\dagger = \mathbb{E}[\boldsymbol{f} \mid \boldsymbol{f}(x_i) = \boldsymbol{Y}_i, i = 1, \dots, N]$. Define $\boldsymbol{Y}:=[\boldsymbol{Y}_1\ \cdots\ \boldsymbol{Y}_N]\in\mathbb{R}^{m\times N}$, and let $\overset{\rightarrow}{\boldsymbol{Y}}$ be the vector formed by concatenating these columns. The estimator $\boldsymbol{f}^\dagger(x)$ is then expressed as 
}
$\boldsymbol{f}^\dagger(x) = K(x, \boldsymbol{x})K(\boldsymbol{x}, \boldsymbol{x})^{-1}\overset{\rightarrow}{\boldsymbol{Y}}$, 
where $K(x, \boldsymbol{x})$ consists of $m$ rows and $N \times m$ columns, formed by concatenating $K(x, x_i)$ for $i=1,\dots, N$. The block matrix $K(\boldsymbol{x}, \boldsymbol{x})$ is defined as:
\begin{align*}
K(\boldsymbol{x}, \boldsymbol{x})=\begin{bmatrix}
K(x_1, x_1) & \cdots & K(x_1, x_N)\\
\vdots & \ddots & \vdots \\
K(x_N, x_1) & \cdots & K(x_N, x_N)
\end{bmatrix}.
\end{align*}
Equivalently, $\boldsymbol{f}^\dagger$ can be obtained in the vector-valued RKHS $\mathcal{U}$ associated with $K$ by solving the optimal recovery problem
\begin{align*}
\begin{cases}
\min\limits_{\boldsymbol{f} \in \mathcal{U}} \|\boldsymbol{f}\|_{\mathcal{U}}^2\\
\operatorname{s.t.} \boldsymbol{f}(x_i) = \boldsymbol{Y}_i, \quad \forall i \in \{1, \dots, N\}.
\end{cases}
\end{align*}
Following the operator-theoretic viewpoint of \cite{owhadi2019operator}, we now allow
more general linear observations.  For ease of presentation, we frame the discussion in terms of scalar-valued GPs.

Let $\mathcal{U}$ be the scalar RKHS associated with a covariance kernel $K$ and
let $f\sim\mathcal{GP}(0,K)$.
Choose bounded linear functionals $\{\phi_i\}_{i=1}^N\subset\mathcal{U}^*$ and define
\[
[\boldsymbol{\phi},f]
:= \bigl(\phi_1(f),\dots,\phi_N(f)\bigr)^T \in\mathbb{R}^N,
\quad
\boldsymbol{y}\in\mathbb{R}^N.
\]
Set
$K(\boldsymbol{\phi},\boldsymbol{\phi}) = \bigl(K(\phi_i,\phi_j)\bigr)_{i,j}$
with $K(\phi_i,\phi_j):=\mathbb{E}[\phi_i(f)\phi_j(f)]$ and
$K(x,\boldsymbol{\phi})=(K(x,\phi_j))_j$ with
$K(x,\phi_j):=\mathbb{E}[f(x)\phi_j(f)]$.
Then the posterior mean of $f$ conditional on $[\boldsymbol{\phi},f]=\boldsymbol{y}$ is
\cite[Sec.~17.8]{owhadi2019operator}
\begin{equation}
\label{goprs}
f^\dagger(x)
= K(x,\boldsymbol{\phi})\,K(\boldsymbol{\phi},\boldsymbol{\phi})^{-1}\boldsymbol{y}. 
\end{equation}
\mathfix{Here $K(\boldsymbol{\phi},\boldsymbol{\phi})^{-1}$ denotes the inverse of $K(\boldsymbol{\phi},\boldsymbol{\phi})$, regularized by a very small nugget term $\eta I$ whenever necessary to ensure numerical stability in the numerical experiment.}
Moreover, $f^\dagger$ is the unique minimizer of
\[
\min_{f\in\mathcal{U}} \|f\|_{\mathcal{U}}^2
\quad\text{s.t.}\quad
[\boldsymbol{\phi},f]=\boldsymbol{y}.
\]
In our MFG applications, each operator $\phi_i$ represents a point evaluation, a derivative, or an integral observation of the unknown fields. 
We refer readers to \cite{williams2006gaussian, owhadi2019operator} for a comprehensive treatment of GPs in machine learning.

The parameters that control the shape of a kernel $K$ are called \emph{hyperparameters}. For example, consider the Gaussian kernel given by
\[
{
K(x,x')=\exp\left(-\frac{|x-x'|^2}{2\mathfix{\ell_{\mathrm K}}^{\,2}}\right).
}
\]
Here, the \emph{length-scale} $\mathfix{\ell_{\mathrm K}}$ determines the rate at which $K(x,x')$ decays as the distance $\lvert x-x' \rvert$ increases: a smaller $\mathfix{\ell_{\mathrm K}}$ yields rapid decay, while a larger $\mathfix{\ell_{\mathrm K}}$ yields slower decay, corresponding to smoother behavior.

\section{GP Policy Iteration Frameworks} 
\label{sec:gppi_frameworks}

PI methods for solving HJB equations and MFGs alternate between solving the HJB equation and, in the MFG case, the FP equation, and updating the feedback control \mathfix{until a prescribed stopping criterion is met}. In this section, we introduce the GPPI method, which represents each unknown as a GP. By exploiting the linearity of GP regression under linear PDE collocation constraints, each iteration reduces to a linear PDE subproblem that admits an explicit representer update. This update is applied to the value function, and, in the MFG setting, also to the population density. In contrast to neural network-based methods, which do not yield explicit solutions or exact minimizations at each step, our approach admits closed-form updates. 

\subsection{A GP Policy Iteration Framework for HJB Equations}
\label{HJB Equation Inverse}
In this subsection, we present our GP framework to address both forward and inverse problems arising from the HJB equation in stochastic control. The HJB equation is fundamental in determining the optimal control strategy and the associated value function. In particular, we consider the time-dependent HJB equation on the $d$-dimensional torus $\mathbb{T}^d$:
\begin{equation}
\begin{cases}
-\partial_t U(x, t)- \frac{1}{2} \sigma(t)^2 \Delta U(x, t) + \sup_{\boldsymbol{q} \in \mathcal{Q}} \Big\{-\nabla U(x, t)^T f(x, t, \boldsymbol{q}) -\ell(x, t, \boldsymbol{q})\Big\} = 0, &\forall  (x, t)\in \mathbb{T}^d \times(0, T), \\[2mm]
U(x, T) = U_T(x),  & \forall x\in \mathbb{T}^d,
\end{cases}
\label{HJBtimedepend}
\end{equation}
where \(U\) represents the value function, and the supremum is taken over the set \(\mathcal{Q}\) of admissible controls, which is typically a compact subset of \(\mathbb{R}^d\). The function \(\ell\) is the running cost associated with the control action \(\boldsymbol{q}\), while \(f\) governs the system dynamics, and \(\sigma\) denotes the scalar volatility. The terminal cost is prescribed by the function \(U_T(x)\). \mathfix{For concreteness,} we adopt the following separable structure for the running cost:
\[
\ell(x,t,\boldsymbol{q}) = V(x,t) + G(t,\boldsymbol{q}),
\]
where \(V\) represents a \mathfix{cost field, possibly time-dependent in the general formulation}, and \(G\) quantifies the cost associated with the control \(\boldsymbol{q}\). This formulation is widely used in applications, such as control-affine systems.

The forward problem thus consists of solving for the value function \(U\) (and consequently the optimal policy) given complete system data. \mathfix{In the inverse setting treated in this paper, we focus on recovering the {cost field} \(V\) from partial noisy observations, while \(f\), \(G\), \(\sigma\), and \(U_T\) are regarded as known inputs.} Specifically, we are interested in the following inverse problem.
\begin{problem}
\label{st_probHJB}
\mathfix{Let \(V^*\) be a cost field, and suppose} that the components \((\sigma, f, G, U_T)\) are known.  Assume that, for a given \(V^*\), the time-dependent HJB equation \eqref{HJBtimedepend} admits a unique classical solution \(U^*\).  In practice, we only have partial, noisy observations of \(U^*\) and \(V^*\), and wish to recover both functions over the entire domain.  

To that end, we assume the following data model:

\begin{enumerate}
  \item \textbf{Partial noisy observations of \(U^*\).}  
    There is a collection of linear observation operators \(\{\phi_j^{o}\}_{j=1}^{N_u}\) and corresponding measurements
      $\boldsymbol{U}^o
      = \bigl([ \phi_1^{o},U^*],\dots,[ \phi_{N_u}^{o},U^*]\bigr)
        + \boldsymbol\epsilon_u
      $, $     \boldsymbol\epsilon_u\sim\mathcal{N}(0,\gamma_u^2 I).$
    We denote \(\boldsymbol\phi^{o}=(\phi_1^{o},\dots,\phi_{N_u}^{o})\).  For example, if we observe \(U^*\) at a finite set of collocation points, each \(\phi_j^{o}\) is a Dirac delta at the corresponding location.

  \item \textbf{Partial noisy observations of \(V^*\).}  
    In a similar fashion, let \(\{\psi_j^{o}\}_{j=1}^{N_v}\) be observation operators for \(V^*\), with data
      $\boldsymbol{V}^o
      = \bigl([ \psi_1^{o},V^*],\dots,[ \psi_{N_v}^{o},V^*]\bigr)
        + \boldsymbol\epsilon_v
      $, $\boldsymbol\epsilon_v\sim\mathcal{N}(0,\gamma_v^2 I)$.
    We write \mathfix{$\Psi=(\psi_1^{o}, \dots, \psi_{N_v}^{o})$}. 
\end{enumerate}

Here, \(\gamma_u\) and \(\gamma_v\) denote the standard deviations of the observation noise for \(U^*\) and \(V^*\), respectively.

\medskip

\textbf{Inverse Problem Statement.}  
The goal is to construct a regularized reconstruction of a pair \((U,V)\) consistent with the time-dependent HJB equation \eqref{HJBtimedepend} and the noisy data \(\boldsymbol{U}^o\), \(\boldsymbol{V}^o\).
\end{problem}
{Recovering the cost field \(V\)} in the HJB framework provides insight into the preferences or objectives that shape optimal behavior. In a robotic navigation scenario, for instance, recovering \(V\) from observed trajectories can reveal areas the robot avoids or favors, reflecting factors such as safety, efficiency, or task relevance.

The PI method for solving the HJB equation begins by initializing a candidate feedback control. In each iteration, the HJB equation is solved to update the value function, which is then used to revise the control policy via a maximization step over the admissible control set. \mathfix{This iterative process is run until a prescribed stopping criterion is met.} The GPPI method adopts this classical strategy and refines it into the following two main steps.

\textbf{Step 1}. We solve the HJB equation. Assume that the value function \( U \) lies in the RKHS \(\mathcal{U}\) associated with the kernel \(K_u\), and that the auxiliary function \( V \) belongs to the RKHS \(\mathcal{V}\) corresponding to the kernel \(K_v\). Observational data for \( V \) are obtained via a set of linear operators, denoted by \(\Psi\), with the corresponding measurement vector \(\boldsymbol{V}^o\). We select \(M\) collocation points \(\{(x_i,t_i)\}_{i=1}^M \subset \mathbb{T}^d \times (0,T]\), where the first \(M_\Omega\) points lie in the interior \(\mathbb{T}^d \times (0,T)\) and the remaining \(M - M_\Omega\) lie on the terminal slice \(\mathbb{T}^d \times \{T\}\).

\mathfix{Given the current policy \(\boldsymbol{q}^{(k)}\), we write the inverse HJB subproblem as the following joint minimization over \((U,V)\):}
\begin{align}
\begin{cases}
\inf_{(U,V) \in \mathcal{U} \times \mathcal{V}} & \alpha_u\|U\|_{\mathcal{U}}^2 +\alpha_v \|V\|_{\mathcal{V}}^2+ \alpha_{v^o}| [\Psi, V] - \boldsymbol{V}^o|^2+\alpha_{u^o}|\left[\boldsymbol{\phi}^o, U\right]-\boldsymbol{U}^{o}|^2 \\
\text{s.t.} & \mathfix{-\partial_t U(x_i,t_i)- \frac{1}{2} \sigma( t_i)^2 \Delta U(x_i, t_i)- \nabla U(x_i, t_i)\cdot f(x_i, t_i, \boldsymbol{q}^{(k)}(x_i, t_i))} \\
&\quad\quad\quad\quad\quad\quad\mathfix{-V(x_i, t_i)-G(t_i, \boldsymbol{q}^{(k)}(x_i, t_i))  = 0}, \quad \forall i = 1,\dots, M_\Omega, \\
& U(x_i, T) = U_T(x_i), \quad \forall i=M_\Omega+1, \dots, M.  
\end{cases}
\label{OptHJBt}
\end{align}
\mathfix{Here, $\alpha_u>0$ and $\alpha_v,\alpha_{v^o},\alpha_{u^o}\ge 0$ are regularization weights. For the forward problem, the cost \(V\) is prescribed and is therefore not an optimization variable: one fixes \(V\) and minimizes over \(U\).  The joint optimization over \((U,V)\) is used only in the inverse problem. The problem \eqref{OptHJBt} can be transformed into a finite-dimensional, linearly constrained quadratic minimization problem. 
Assume that the constraint set is nonempty and that the quadratic objective is strictly convex on this set. 
Then \eqref{OptHJBt} admits a unique solution, for which explicit formulas can be derived. 
We refer the reader to Appendix~\ref{Appendix5} for the derivation of these formulas and a discussion of uniqueness.} \mathfix{The minimizer of \eqref{OptHJBt} is denoted by $(U^{(k)}, V^{(k)})$ in the inverse case, and by $U^{(k)}$ in the forward case where $V$ is prescribed.}

\textbf{Step 2}. 
 \mathfix{Next, we update the policy $\boldsymbol{q}$, which is a $d$-dimensional vector-valued function. At each interior collocation point, we compute the improved policy value by maximizing the Hamiltonian. We then define the updated policy on the whole domain as the posterior mean of a vector-valued GP conditioned on these computed values.} More precisely, let $\chi = \{(x_1, t_1), \dots, (x_{M_\Omega}, t_{M_\Omega})\}$ be the collection of collocation points on $\mathbb{T}^d \times (0, T)$. Then, we compute 
\begin{align*}
\boldsymbol{\mathfrak{q}}^{k+1, i} :=\underset{\boldsymbol{q} \in \mathcal{Q}}{\arg \max }\{-\nabla U^{(k)}(x_i, t_i) \cdot f(x_i, t_i,  \boldsymbol{q})-\ell(x_i,t_i, \boldsymbol{q})\}, \quad \forall i = 1,\dots, M_\Omega.
\end{align*}
\mathfix{In the inverse case, $\ell(x_i,t_i, \boldsymbol{q}) = V^{(k)}(x_i,t_i) + G(t_i, \boldsymbol{q})$. Since $V^{(k)}$ is independent of $\boldsymbol{q}$, it does not affect the pointwise maximizer.}
{Consider a vector-valued GP \(\boldsymbol{\xi}_{\boldsymbol{q}} : \mathbb{T}^d \times (0, T) \to \mathbb{R}^d\), for all \((x, t) \in \mathbb{T}^d \times (0, T)\), with zero mean, that is, \(\mathbb{E}[\boldsymbol{\xi}_{\boldsymbol{q}}(x, t)] = \mathbf{0}\). Its covariance is described by a matrix-valued kernel \(\boldsymbol{K}_{\boldsymbol{q}}((x, t), (y, s)) \in \mathbb{R}^{d \times d}\), for all \((x, t), (y, s) \in \mathbb{T}^d \times (0, T)\), where each block \(\boldsymbol{K}_{\boldsymbol{q}}((x, t), (y, s))\) encodes both the variances and cross-covariances between the \(d\) components of \(\boldsymbol{\xi}_{\boldsymbol{q}}(x, t)\) and \(\boldsymbol{\xi}_{\boldsymbol{q}}(y, s)\). Given the set \(\chi = \{(x_1, t_1), \dots, (x_{M_{\Omega}}, t_{M_{\Omega}})\}\) of collocation points, one assembles \(\boldsymbol{K}_{\boldsymbol{q}}(\chi, \chi) \in \mathbb{R}^{d{M_{\Omega}} \times d{M_{\Omega}}}\) by placing each \(\boldsymbol{K}_{\boldsymbol{q}}((x_i, t_i), (x_j, t_j)) \in \mathbb{R}^{d \times d}\) as the \((i, j)\)-th block in a grid, and forms \(\boldsymbol{K}_{\boldsymbol{q}}((x, t), \chi) \in \mathbb{R}^{d \times (d{M_{\Omega}})}\) by horizontally concatenating \(\boldsymbol{K}_{\boldsymbol{q}}((x, t), (x_j, t_j))\) for \(j = 1, \dots, {M_{\Omega}}\).}

Suppose that \(\boldsymbol{\xi}_{\boldsymbol{q}}(x_i, t_i) = \boldsymbol{\mathfrak{q}}^{k+1, i} \in \mathbb{R}^d\) are the observed outputs at the collocation points \((x_i, t_i)\), for \(i = 1, \dots, M_{\Omega}\). We build a vector \(\boldsymbol{\mathfrak{q}}^{k+1} \in \mathbb{R}^{dM_{\Omega}}\) by stacking each \(\boldsymbol{\mathfrak{q}}^{k+1, i} \in \mathbb{R}^d\) vertically, i.e., 
\begin{align}
\label{eq:defqvec}
\boldsymbol{{\mathfrak{q}}}^{k+1} = \begin{pmatrix}
\boldsymbol{\mathfrak{q}}^{k+1, 1}\\ \vdots\\  
\boldsymbol{\mathfrak{q}}^{k+1, {M_{\Omega}}}
\end{pmatrix}.
\end{align}
To obtain the updated policy over the entire domain, we approximate \(\boldsymbol{q}\) using the posterior mean of \(\boldsymbol{\xi}_{\boldsymbol{q}}\), that is, $
\boldsymbol{q}^{(k+1)} = \mathbb{E}\left[\boldsymbol{\xi}_{\boldsymbol{q}} \,\middle|\, \boldsymbol{\xi}_{\boldsymbol{q}}(x_i, t_i) = \boldsymbol{\mathfrak{q}}^{k+1, i}, \; i = 1, \dots, M_{\Omega} \right]$. Thus,
\begin{align}
\label{eq:qk1}
\boldsymbol{q}^{(k+1)}(x, t) = \boldsymbol{K}_{\boldsymbol{q}}((x, t), \chi) \boldsymbol{K}_{\boldsymbol{q}}(\chi, \chi)^{-1} \boldsymbol{\mathfrak{q}}^{k+1}, \quad \forall (x, t)\in \mathbb{T}^d \times(0, T).
\end{align}
{Since admissibility is imposed only at the collocation points, the GP reconstruction in \eqref{eq:qk1} is not automatically guaranteed to satisfy 
$\boldsymbol{q}^{(k+1)}(x,t)\in\mathcal{Q}$ throughout the domain. 
In this paper, we restrict attention to settings in which the reconstructed policy remains admissible. 
One possible alternative is to enforce admissibility by projecting the GP mean pointwise onto $\mathcal{Q}$, but we do not consider this variant here.}

\mathfix{The above procedure is repeated until the prescribed stopping criterion is met. Specifically, in \Cref{secNumericalExpe}, we terminate the iterations when the change in the policy drops below a prescribed tolerance.}

\subsection{A GP Policy Iteration Framework for Stationary MFGs}
\label{MFG Forward Problem Stationary}
In this subsection, we present a unified GPPI framework to solve forward and inverse stationary MFG problems. 
For brevity, we illustrate our method using the following prototypical MFG system on the $d$-dimensional torus $\mathbb{T}^d$:
\begin{equation}
\begin{cases}
-\nu\Delta u + H(x, \nabla u) +\lambda = F(m)+V(x), & \quad \forall x\in \mathbb{T}^d, \\ 
-\nu \Delta m - \operatorname{div}\left( D_p H(x, \nabla u) \, m \right) = 0, & \quad \forall x\in \mathbb{T}^d, \\
\int_{\mathbb{T}^d} u \, \dif x  = 0, \quad \int_{\mathbb{T}^d} m \, \dif x  = 1. 
\end{cases}
\label{eqonedimension1forward}
\end{equation}
{Here, $u$ denotes the value function, $m$ the agent distribution, $H$ the Hamiltonian, $F$ the coupling term, $V$ the spatial cost, $\nu$ the viscosity coefficient, and $\lambda$  is a real constant. In a typical MFG forward problem, one seeks to solve $(u, m, \lambda)$, which encodes the Nash equilibrium, given the environmental configuration $(H, \nu, F, V)$. }

For the inverse problem, we aim to infer both the agents' strategies and the environmental parameters from partial, noisy observations of the agents' distribution and the environment. More precisely, we seek to solve the following inverse problem.
\begin{problem}
\label{st_prob}
Let \(V^*\) be a spatial cost function and suppose that the components \((H,\nu,F)\) are known.  Assume that, for a given \(V^*\), the stationary MFG system \eqref{eqonedimension1forward} admits a unique classical solution \((u^*,m^*,\lambda^*)\).  In practice, we only observe noisy, partial measurements of \(m^*\) and \(V^*\), and our goal is to compute a regularized reconstruction of the configuration \((u^*,m^*,\lambda^*,V^*)\).

To formalize, suppose we have:

\begin{enumerate}
  \item \textbf{Partial noisy observations of \(m^*\).}  
    There is a collection of linear observation operators \mathfix{\(\{\phi_j^o\}_{j=1}^{N_m}\)} and related data $
      \boldsymbol{m}^o
      = \bigl([\phi_1^o,m^*],\dots,[\phi_{N_m}^o,m^*]\bigr)
        + \boldsymbol\epsilon_m$, $  \boldsymbol\epsilon_m\sim\mathcal{N}(0,\gamma_m^2 I)$. 
    Here \(\boldsymbol\phi^o=(\phi_1^o,\dots,\phi_{N_m}^o)\), and \(\gamma_m\) denotes the standard deviation of the measurement noise for \(m^*\).  

  \item \textbf{Partial noisy observations of \(V^*\).}  
    Similarly, let \(\{\psi_j^{o}\}_{j=1}^{N_v}\) be observation operators for \(V^*\), with measurements
      $\boldsymbol{V}^o
      = \bigl([\psi_1^{o},V^*],\dots,[\psi_{N_v}^{o},V^*]\bigr)
        + \boldsymbol\epsilon_v$, $\boldsymbol\epsilon_v\sim\mathcal{N}(0,\gamma_v^2 I)$,
    where \(\gamma_v\) is the noise standard deviation for observations of \(V^*\).
\end{enumerate}

\medskip

\textbf{Inverse Problem Statement.}  
Given the noisy observations \(\boldsymbol{m}^o\) and \(\boldsymbol{V}^o\), and the stationary MFG system \eqref{eqonedimension1forward}, we seek to construct a regularized reconstruction of \((u,m,\lambda,V)\) over the entire domain.
\end{problem}

Before introducing the GPPI method, we recall the standard PI method \cite{cacace2021policy}. The PI method solves \eqref{eqonedimension1forward} by introducing the feedback control $
\boldsymbol{Q}(x)=D_p H\bigl(x,\nabla u\bigr).$
Starting with an initial guess \(\boldsymbol{Q}^{(0)}\), at each iteration \(k\) the PI method first solves the linear FP equation corresponding to \(\boldsymbol{Q}^{(k)}\), 
\begin{align}
\label{eq:linear_st_fk}
\left\{\begin{array}{l}
-\nu \Delta m^{(k)}(x)-\operatorname{div}\bigl(m^{(k)} \boldsymbol{Q}^{(k)}\bigr)(x)=0, \quad \forall x\in \mathbb{T}^d ,\\
\int_{\mathbb{T}^d} m^{(k)} \,\dif x=1, \quad m^{(k)} \geq 0 ,
\end{array}\right.
\end{align}
to obtain the density $m^{(k)}$ corresponding to the current policy. Next, given $m^{(k)}$, the PI algorithm solves the HJB equation 
\begin{equation}
\left\{\begin{array}{l}
-\nu \Delta u^{(k)}(x) + \boldsymbol{Q}^{(k)}(x) \cdot \nabla u^{(k)}(x) + \lambda^{(k)} = L(x, \boldsymbol{Q}^{(k)}(x)) + V(x) + F(m^{(k)}(x)), \quad \forall x\in \mathbb{T}^d, \\
\int_{\mathbb{T}^d} u^{(k)} \,\dif x = 0,
\end{array}\right.
\label{eq:mfg_systeU1}
\end{equation}
\mathfix{where $L$ is the Lagrangian, equivalently the convex dual associated with  $H$. Finally, the policy is updated pointwise by setting}
\[
\boldsymbol{Q}^{(k+1)}(x)= \arg \max_{\|\boldsymbol{q}\| \le \mathfix{R_{\mathrm{pol}}}} \Bigl\{ \boldsymbol{q}\cdot \nabla u^{(k)}(x) - L(x, \boldsymbol{q}) \Bigr\}, \quad \forall x \in \mathbb{T}^d,
\]
where $\mathfix{R_{\mathrm{pol}}}$ is chosen sufficiently large to ensure that the policy remains bounded and does not diverge.

Here, we employ GPs to approximate the unknown functions $m$, $u$, and $\boldsymbol{Q}$, while modeling $\lambda$ as a Gaussian random variable. In contrast to finite difference methods \cite{lauriere2023policy, cacace2021policy, tang2024learning, ren2024policy}, our framework naturally integrates uncertainty quantification (UQ) into each step of the PI because each iteration involves solving a linear PDE. The solution to this linear PDE can be interpreted as a maximum a posteriori (MAP) estimate under linear observations. With GP priors, the resulting posterior remains a GP. This inherent property facilitates error estimation and optimal experimental design (e.g., selecting sample points for the next iteration). We defer a detailed study of UQ to future work. 

Moreover, compared to neural network-based methods \cite{assouli2024deep}, the linearity of GPs and the underlying PDEs allows each GPPI iteration to admit an explicit reduced formulation. In particular, the policy-evaluation subproblems are solved through linear GP collocation systems, while the policy-improvement step is explicit in the quadratic case and otherwise reduces to a low-dimensional local solve rather than to a large-scale iterative minimization over the full coupled PDE system.

More precisely, let \(\{x_i\}_{i=1}^M\) be collocation points on \(\mathbb{T}^d\). The GPPI method proceeds in three steps.

\textbf{Step 1}. Assume that the solution $m$ is in the RKHS $\mathcal{M}$ associated with the kernel $K_m$. Let $\boldsymbol{\phi}^o$ denote the linear observation operator and $\boldsymbol{m}^o$ the corresponding observation data for $m$, as defined in Problem~\ref{st_prob}. 
  Given the current policy $\boldsymbol{Q}^{(k)}$, we approximate the solution $m^{(k)}$ of \eqref{eq:linear_st_fk} by solving the following minimization problem
\begin{align}
\begin{cases}
\inf_{m \in \mathcal{M}} & \alpha_m\|m\|_{\mathcal{M}}^2+ \alpha_{m^o} |[\boldsymbol{\phi}^o, m] - \boldsymbol{m}^o|^2 \\
\text{s.t.} &  - \nu \Delta m(x_i)- \operatorname{div}(m \boldsymbol{Q}^{(k)})(x_i) = 0, \quad \forall i = 1, \ldots, M, \\
& \int_{\mathbb{T}^d} m \, \dif x  = 1,
\end{cases}
\label{OptGPtdProb_Cforwardm}
\end{align}
where $\alpha_m$ and $\alpha_{m^o}$ are \mathfix{nonnegative} real numbers serving as penalization parameters. We choose $(\alpha_m, \alpha_{m^o}) = (\frac{1}{2}, 0)$ for the forward problem (i.e., when there are no observations for $m$).  \mathfix{For the inverse problem, we typically set \(\alpha_m=1/2\) and treat
\(\alpha_{m^o}\) as a regularization hyperparameter controlling the
relative weight between the RKHS prior and the data fidelity term.
If one adopts a strictly MAP-calibrated Gaussian noise model and writes
the data term as \(\alpha_{m^o}\|[\boldsymbol{\phi}^o,m]-\boldsymbol{m}^o\|^2\), then the choice
\(\alpha_{m^o}=1/(2\gamma_m^2)\) corresponds to observation noise variance
\(\gamma_m^2\). More generally, for arbitrary weights
\(\alpha_m,\alpha_{m^o}>0\), the same objective can be interpreted as a
MAP estimator with an effective noise variance
$
\gamma_{\rm eff}^2=\frac{\alpha_m}{\alpha_{m^o}}.
$
In the numerical experiments, the observation weights can be selected
empirically to balance data fidelity and regularization.} 
\mathfix{The problem \eqref{OptGPtdProb_Cforwardm} is a linearly constrained quadratic minimization problem. The derivation of the explicit solution formula for \eqref{OptGPtdProb_Cforwardm} is given in Appendix~\ref{Appendix3}.}

The above formulation admits a natural probabilistic interpretation. Specifically, we model the unknown function $m$ as a GP with prior
$m \sim \mathcal{GP}(0, K_m)$, 
where $K_m$ is the covariance kernel for $m$. Observations are obtained via a linear operator $\boldsymbol{\phi}^o$, yielding
$\boldsymbol{m}^o = [\boldsymbol{\phi}^o, m] + \boldsymbol\epsilon_m$, $ \boldsymbol\epsilon_m \sim \mathcal{N}(0,\mathfix{\gamma_m^2} I)$,
so that the likelihood is given by
\mathfix{$p(\boldsymbol{m}^o \mid m) \propto \exp\Bigl(-\frac{1}{2\gamma_m^2}\|[\boldsymbol{\phi}^o, m]-\boldsymbol{m}^o\|^2\Bigr)$}. 
We also require that the FP equation constraint and the mass conservation condition hold exactly. In particular, at each collocation point $x_i$, for $i=1,\dots,M$, we impose 
$FP(x_i) \equiv -\nu\,\Delta m(x_i) - \div\bigl(m\boldsymbol{Q}^{(k)}\bigr)(x_i) = 0$,
and  enforce 
$\int_{\mathbb{T}^d} m(x)\,\dif x = 1.$
Thus, the posterior may be written conditionally as
\[
p\Bigl(m \,\Big|\, \boldsymbol{m}^o,\,\{FP(x_i)=0\}_{i=1}^M,\,\int_{\mathbb{T}^d}m(x)\,\dif x=1\Bigr)
\propto p(\boldsymbol{m}^o \mid m)\,p(m)\, \prod_{i=1}^M \delta\Bigl(FP(x_i)\Bigr)\, \delta\!\Bigl(\int_{\mathbb{T}^d} m(x)\,\dif x-1\Bigr),
\]
where $\delta$ denotes the Dirac measure. \mathfix{With the normalization $\alpha_m=1/2$ and $\alpha_{m^o}=1/(2\gamma_m^2)$, \eqref{OptGPtdProb_Cforwardm} is exactly the MAP estimate}
\[
m^{(k)} = \arg\max_m \ln p\Bigl(m \,\Big|\, \boldsymbol{m}^o,\,\{FP(x_i)=0\}_{i=1}^M,\,\int_{\mathbb{T}^d} m(x)\,\dif x=1\Bigr),
\]
\mathfix{in which the GP prior and the data fidelity term are balanced subject to these hard constraints. For other choices of $\alpha_m$ and $\alpha_{m^o}$, the same optimization should be read as a regularized MAP-type estimator with tunable prior and data-fidelity weights.}

Furthermore, since the observations of $m$ are imposed via the linear operator $\boldsymbol{\phi}^o$ and the PDE constraints are linear, a Gaussian prior on $m$ yields a Gaussian posterior that can be sampled efficiently.

 \textbf{Step 2}.
 Suppose that the value function $u$ is in an RKHS $\mathcal{U}$ associated with the kernel $K_u$, and that $V$ is a function in an RKHS $\mathcal{V}$ associated with the kernel $K_v$. Let $\Psi$ denote the collection of observation operators for $V$, with corresponding data $\boldsymbol{V}^o$. Given the fixed policy $\boldsymbol{Q}^{(k)}$ and the density function $m^{(k)}$ obtained in Step 1, we approximate the solution $u^{(k)}$ of the linear equation \eqref{eq:mfg_systeU1} by solving the minimization problem
\begin{align}
\begin{cases}
\inf _{(u,\lambda, V) \in\mathcal{U}\times \mathbb{R}\times \mathcal{V}}&\alpha_{u}\|u\|_{\mathcal{U}}^2+\alpha_{\lambda}|\lambda|^2+\alpha_v\|V\|_{\mathcal{V}}^2+ \alpha_{v^o}| [\Psi, V] - \boldsymbol{V}^o|^2 \\
\text{s.t.} &  -\nu \Delta u(x_i) + \boldsymbol{Q}^{(k)}(x_i) \cdot \nabla u(x_i) + \lambda\\
& \quad\quad\quad\quad= L(x_i, \boldsymbol{Q}^{(k)}(x_i)) + V(x_i) + F(m^{(k)}(x_i)), \quad \forall i = 1, \ldots, M,  \\
& 
\int_{\mathbb{T}^d} u \,\dif x= 0, 
\end{cases}
\label{eq:mfg_systeUforward}
\end{align}
where $\alpha_u$, $\alpha_{\lambda}$, $\alpha_v$, and $\alpha_{v^o}$ are \mathfix{nonnegative} regularization parameters. Analogously to \eqref{OptGPtdProb_Cforwardm}, we set $\alpha_u = \frac{1}{2}$, $\alpha_{\lambda} = \frac{1}{2}$, $\alpha_v = 0$, and $\alpha_{v^o} = 0$ for the forward problem (i.e., when $V$ is given). \mathfix{In the forward problem, $V$ is prescribed and is therefore not an optimization variable: it is fixed at its known value, and the minimization is taken only over $(u,\lambda)$, the joint optimization over $(u,\lambda,V)$ being used only in the inverse problem.} For the inverse problem, a typical choice is $\alpha_u = \frac{1}{2}$, $\alpha_{\lambda} = \frac{1}{2}$, $\alpha_v = \frac{1}{2}$. \mathfix{If the objective is written as $\alpha_{v^o}\|[\Psi, V] - \boldsymbol{V}^o\|^2$, then we may choose $\alpha_{v^o} =\frac{1}{2\gamma_v^2} $ for Gaussian noise with variance $\gamma_v^2$. 
}\mathfix{Meanwhile, \eqref{eq:mfg_systeUforward} is a quadratic optimization problem subject to linear constraints. For a discussion of the explicit formula associated with \eqref{eq:mfg_systeUforward}, see Appendix~\ref{Appendix3}.}

\mathfix{Analogous to Step 1, we adopt a probabilistic interpretation for the unknowns $u$, $\lambda$, and $V$: $u$ and $V$ are assigned GP priors, while $\lambda$ is modeled by a Gaussian prior. The observation constraints on $V$ are imposed via a Gaussian likelihood, and the HJB PDE together with the zero-mean normalization of $u$ are enforced by Dirac measures, restricting the posterior to the set of $(u,\lambda,V)$ that satisfy the corresponding equations. Thus, solving the optimization problem in \eqref{eq:mfg_systeUforward} is equivalent to computing the MAP estimate under these priors and linear observations when the objective weights match the corresponding Gaussian prior and noise normalizations.}

\textbf{Step 3}.
Next, we proceed to update the policy \(\boldsymbol{Q}\), a \(d\)-dimensional vector-valued function. To determine \(\boldsymbol{Q}\), we initially update its values at the designated collocation points, and then derive \(\boldsymbol{Q}^{(k+1)}\) as the GP mean, conditioned on the latest observations of \(\boldsymbol{Q}\). Specifically, let \(X = \{x_1, \dots, x_M\}\) represent the set of collocation points on \(\mathbb{T}^d\). We then perform the following computation: 
\begin{align*}
\mathbf{q}^{k+1, i}=\arg \max _{\|\boldsymbol{q}\| \le \mathfix{R_{\mathrm{pol}}}}\left\{\boldsymbol{q} \cdot \nabla u^{(k)}(x_i)-L(x_i,\boldsymbol{q})\right\}, \quad \forall i = 1, \dots, M.
\end{align*}
\mathfix{Here, \(R_{\mathrm{pol}}\) bounds the pointwise maximization at the collocation points.}

Consider a vector-valued GP $\boldsymbol{\xi}_{\boldsymbol{Q}} : \mathbb{T}^d \to \mathbb{R}^d$ with zero mean, that is, $\mathbb{E}[\boldsymbol{\xi}_{\boldsymbol{Q}}(x)] = \mathbf{0}$ for all $x\in \mathbb{T}^d$. Its covariance is described by a matrix-valued kernel $\boldsymbol{K}_{\boldsymbol{Q}}(x,y)\in \mathbb{R}^{d\times d}$ for all $x, y \in \mathbb{T}^d$, where each block $\boldsymbol{K}_{\boldsymbol{Q}}(x,y)$ encodes both the variances  and cross-covariances between the $d$ components of $\boldsymbol{\xi}_{\boldsymbol{Q}}(x)$ and $\boldsymbol{\xi}_{\boldsymbol{Q}}(y)$. Given the set  $X$ of collocation points, one assembles $\boldsymbol{K}_{\boldsymbol{Q}}(X,X)\in \mathbb{R}^{dM\times dM}$ by placing each $\boldsymbol{K}_{\boldsymbol{Q}}(x_i,x_j)\in \mathbb{R}^{d\times d}$ as the $(i,j)$-th block in a grid and forms $\boldsymbol{K}_{\boldsymbol{Q}}(x,X)\in \mathbb{R}^{d\times(dM)}$ by horizontally concatenating $\boldsymbol{K}_{\boldsymbol{Q}}(x,x_j)$ for $j=1,\dots,M$.

Suppose that $\boldsymbol{\xi}_{\boldsymbol{Q}}(x_i) = \mathbf{q}^{k+1,i}\in\mathbb{R}^d$ are the observed outputs at the collocation points $\{x_i\}_{i=1}^M$. Define $$\mathbf{q}^{k+1} =
\begin{pmatrix}
\mathbf{q}^{k+1,1}\\ \vdots\\  
\mathbf{q}^{k+1,M}
\end{pmatrix}.$$ 
Then, we approximate $\boldsymbol{Q}$ by the posterior mean of $\boldsymbol{\xi}_{\boldsymbol{Q}}$, i.e., $\boldsymbol{Q}^{(k+1)} = \mathbb{E}[\boldsymbol{\xi}_{\boldsymbol{Q}}\mid \boldsymbol{\xi}_{\boldsymbol{Q}}(x_i) = \mathbf{q}^{k+1, i},\ i=1, \dots, M]$. Thus, 
\begin{align}
\label{eq:stQk1}
\boldsymbol{Q}^{(k+1)}(x)
\;=\;
\boldsymbol{K}_{\boldsymbol{Q}}(x,X)\,
\boldsymbol{K}_{\boldsymbol{Q}}(X,X)^{-1}\,
\mathbf{q}^{k+1},\quad \forall x\in \mathbb{T}^d. 
\end{align}
After obtaining $\boldsymbol{Q}^{(k+1)}$ in \eqref{eq:stQk1}, we iterate Steps 1 to 3. In Step 1, when computing the divergence of $\boldsymbol{Q}^{(k+1)}$, we differentiate the explicit expression given in \eqref{eq:stQk1}.  \mathfix{The procedure is run until the prescribed stopping criterion is met. Specifically, in \Cref{secNumericalExpe}, we terminate the iterations when the policy update drops below a prescribed tolerance.}

\subsection{A GP Policy Iteration Framework for Time-Dependent MFGs}
\label{MFG Forward Problem Time Dependent}
In this subsection, we present a unified GPPI framework for solving time-dependent MFG forward and inverse problems. Our approach leverages GP models to approximate the unknown functions in the MFG system, thereby yielding an explicit, tractable formulation at every iteration. In the forward problem, the objective is to compute the Nash equilibrium of the system, while in the inverse problem, the goal is to recover the underlying system parameters from partial, noisy observations. For ease of exposition, we focus on the following time-dependent MFG system on the $d$-dimensional torus $\mathbb{T}^d$:
\begin{equation}
\begin{cases}
-\partial_t u-\nu \Delta u+H(x,t, \nabla u) =F(m)+ V(x,t), & \forall (x, t)\in \mathbb{T}^d \times(0, T), \\
\partial_t m-\nu \Delta m-\operatorname{div}\left(m D_p H(x,t, \nabla u)\right)=0, & \forall (x, t)\in \mathbb{T}^d \times(0, T), \\
m(x, 0)=m_0(x), \quad u(x, T)=U_T(x), & \forall x\in \mathbb{T}^d.
\end{cases}
\label{eq:MFG_Systemt}
\end{equation}
Here, $u$ denotes the value function, $m$ the agent distribution, $\nu$ the viscosity coefficient, $H$ the Hamiltonian, $V$ the \mathfix{cost field}, $F$ the coupling function, $m_0$ the initial distribution, and $U_T$ the terminal cost. In the forward problem, one solves for $(u, m)$ given the functions $H$, $V$, $F$, $m_0$, and $U_T$. On the other hand, our inverse problem aims to recover the true solution components $u^*$, $m^*$, and $V^*$ based on partial, noisy observations of $m^*$ and a subset of $V^*$. Specifically, we consider the following inverse problem.
\begin{problem}
\label{td_prob}
\mathfix{Let \(V^*\) be a cost field, and suppose} that the components \((H,\nu,F, m_0, U_T)\) are known.  Assume that, for a given \(V^*\), the time-dependent MFG system \eqref{eq:MFG_Systemt} admits a unique classical solution \((u^*,m^*)\).  In practice, we only have access to noisy, partial measurements of \(m^*\) and \(V^*\), and we aim to compute a regularized reconstruction of the configuration \((u^*,m^*,V^*)\).

Concretely, we assume:

\begin{enumerate}
  \item \textbf{Partial noisy observations of \(m^*\).}  
    Let  \mathfix{\(\{\phi_j^o\}_{j=1}^{N_m}\)} be linear observation operators, and denote their measurements by $
      \boldsymbol{m}^o
      = \bigl([\phi_1^o, m^*], \dots, [\phi_{N_m}^o, m^*]\bigr)
        + \boldsymbol\epsilon_m,
      \
      \boldsymbol\epsilon_m \sim \mathcal{N}(0,\gamma_m^2 I).$
    Here \(\boldsymbol\phi^o=(\phi_1^o,\dots,\phi_{N_m}^o)\), and \(\gamma_m\) is the standard deviation of the observation noise for \(m^*\).  

  \item \textbf{Partial noisy observations of \(V^*\).}  
    Likewise, let  \mathfix{\(\{\psi_j^o\}_{j=1}^{N_v}\)} be observation operators for \(V^*\), with data
$      \boldsymbol{V}^o
      = \bigl([\psi_1^o, V^*], \dots, [\psi_{N_v}^o, V^*]\bigr)
        + \boldsymbol\epsilon_v,
      \
      \boldsymbol\epsilon_v \sim \mathcal{N}(0,\gamma_v^2 I),$
    where \(\gamma_v\) denotes the noise standard deviation for \(V^*\) observations.
\end{enumerate}

\medskip

\textbf{Inverse Problem Statement.}  
Given the noisy data \(\boldsymbol{m}^o\) and \(\boldsymbol{V}^o\) together with the MFG system \eqref{eq:MFG_Systemt}, construct a regularized reconstruction of \((u,m,V)\).
\end{problem}

We now briefly recall the standard PI method \cite{cacace2021policy} for solving the forward problem of \eqref{eq:MFG_Systemt}. The PI method first introduces the feedback control $
\boldsymbol{Q}(x,t) = D_pH(x,t,\nabla u).$
Beginning with an initial guess \(\boldsymbol{Q}^{(0)}\), the method iteratively refines the solution through three primary steps. In the first step, with the current control \(\boldsymbol{Q}^{(k)}\), we solve the linear FP equation
\begin{equation}
\label{eq:linear_timedepend_fk}
\left\{
\begin{aligned}
\partial_t m^{(k)} - \nu \Delta m^{(k)} - \operatorname{div}\bigl(m^{(k)} \boldsymbol{Q}^{(k)}\bigr) &= 0, \quad\quad\,\quad \forall \, (x,t) \in \mathbb{T}^d \times (0,T),\\[1mm]
m^{(k)}(x,0) &= m_0(x), \quad  \forall\, x \in \mathbb{T}^d,
\end{aligned}
\right.
\end{equation}
thereby updating the density  \(m^{(k)}\). In the second step, given \(m^{(k)}\) and \(\boldsymbol{Q}^{(k)}\), we solve the HJB equation
\begin{equation}
\label{eq:mfg_systimedependU1}
\left\{
\begin{aligned}
-\partial_t u^{(k)} - \nu \Delta u^{(k)} + \boldsymbol{Q}^{(k)}\cdot \nabla u^{(k)} &= L\bigl(x, t,  \boldsymbol{Q}^{(k)}(x,t)\bigr) + V(x,t) + F\bigl(m^{(k)}(x,t)\bigr), \quad \forall (x,t) \in \mathbb{T}^d \times (0,T),\\[1mm]
u^{(k)}(x,T) &= U_T(x), \quad \forall x \in \mathbb{T}^d,
\end{aligned}
\right.
\end{equation}
\mathfix{which in turn updates the value function \(u^{(k)}\), where $L$ is the Lagrangian associated with $H$. In the third step, the control is updated by computing}
\[
\boldsymbol{Q}^{(k+1)}(x,t)= \arg\max_{\|\boldsymbol{q}\| \le \mathfix{R_{\mathrm{pol}}}}\Bigl\{\boldsymbol{q}\cdot \nabla u^{(k)}(x,t) - L(x, t, \boldsymbol{q})\Bigr\}, \quad \forall (x,t) \in \mathbb{T}^d \times (0,T).
\]
Here, \(\mathfix{R_{\mathrm{pol}}}\) is chosen sufficiently large to bound the policy and prevent its divergence.
This systematic iteration refines the solution and is designed so that the MFG system is increasingly well satisfied at the collocation points.

Similar to the stationary case, our approach approximates the unknown functions using GPs. Consequently, solving the FP and HJB equations reduces to quadratic minimization problems that combine a regularization term (imposed by the GP prior) with a data fidelity term, all under linear PDE constraints. Notably, each step admits a unique explicit solution. Moreover, a natural probabilistic interpretation emerges: these optimization problems correspond to computing the MAP estimates under GP priors.

\textbf{Step 1}.
Assume that the solution \( m \) lies in the RKHS \(\mathcal{M}\) associated with the kernel \(K_m\). Let \(\boldsymbol{\phi}^o\) denote the observation operator and \(\boldsymbol{m}^o\) the corresponding observation data for \( m \), as defined in Problem~\ref{td_prob}.
We choose \mathfix{\( M^{\mathrm{FP}} \)} collocation points \mathfix{\(\{(x_i, t_i)\}_{i=1}^{M^{\mathrm{FP}}} \subset \mathbb{T}^d \times [0,T) \)}, where the first \mathfix{\( M_\Omega^{\mathrm{FP}} \)} points lie in the interior domain \(\mathbb{T}^d \times (0, T)\), and the remaining \mathfix{\( M^{\mathrm{FP}} - M_\Omega^{\mathrm{FP}} \)} points lie on the initial time slice \(\mathbb{T}^d \times \{0\} \). 
Given the current policy \(\boldsymbol{Q}^{(k)}\), we approximate the solution \( m^{(k)} \) of \eqref{eq:linear_timedepend_fk} by solving the minimization problem
\begin{align}
\label{eq:opt:td:fp}
\begin{cases}
\inf_{m \in \mathcal{M}} & \alpha_m\|m\|_{\mathcal{M}}^2+\alpha_{m^o} |[\boldsymbol{\phi}^o, m] - \boldsymbol{m}^o|^2\\
\text{s.t.} & \partial_t m(x_i, t_i) - \nu \Delta m(x_i, t_i) - \operatorname{div}(m \boldsymbol{Q}^{(k)})(x_i, t_i) = 0, \quad \forall i = 1, \dots, \mathfix{M_{\Omega}^{\mathrm{FP}}},\\
& m(x_i,0) = m_0(x_i),\quad \forall i = \mathfix{M_{\Omega}^{\mathrm{FP}}+1}, \dots, \mathfix{M^{\mathrm{FP}}}.
\end{cases}
\end{align}
Here, \(\alpha_m\) and \(\alpha_{m^o}\) are  regularization parameters. For the forward problem (i.e., when no observations of \( m \) are available), we set \((\alpha_m, \alpha_{m^o}) = (\frac{1}{2},0)\). For the inverse problem, a common choice is to set \(\alpha_m = \frac{1}{2}\). \mathfix{ If the data fidelity term is written as
\(\alpha_{m^o}\|[\boldsymbol{\phi}^o,m]-\boldsymbol{m}^o\|^2\), then the noise-calibrated MAP choice under independent Gaussian noise with variance \(\gamma_m^2\) is
\(\alpha_{m^o}=1/(2\gamma_m^2)\). 
In the numerical experiments,  \(\alpha_{m^o}\) can be treated as a tunable regularization hyperparameter.}

It is important to note that the optimization problem above is a quadratic minimization under linear constraints, which guarantees the existence of a unique explicit solution \mathfix{if the feasible affine space is nonempty}. For a detailed derivation of this explicit formula, we refer the reader to Appendix~\ref{Appendix4}.

\textbf{Step 2}.
Similarly, we choose \mathfix{\( M^{\mathrm{HJB}} \)} collocation points \mathfix{\(\{(x_j, t_j)\}_{j=1}^{M^{\mathrm{HJB}}} \subset \mathbb{T}^d \times (0,T] \)}, where the first \mathfix{\( M_\Omega^{\mathrm{HJB}} \)} points are the interior HJB collocation points and the remaining \mathfix{\( M^{\mathrm{HJB}} - M_\Omega^{\mathrm{HJB}} \)} points lie on the terminal time slice \(\mathbb{T}^d \times \{T\} \). \mathfix{We distinguish these HJB counts from the FP counts \(M^{\mathrm{FP}}\) and \(M_\Omega^{\mathrm{FP}}\) used in Step~1.}
Suppose the value function \(u\) resides in an RKHS \(\mathcal{U}\) associated with kernel \(K_u\), and the unknown function \(V\) lies in an RKHS \(\mathcal{V}\) associated with kernel \(K_v\). Let \(\Psi\) represent the collection of observation operators corresponding to data \(\boldsymbol{V}^o\). Given the current policy \(\boldsymbol{Q}^{(k)}\) and the density \(m^{(k)}\) computed in Step 1, we approximate the solution  \(u^{(k)}\) of the linear equation \eqref{eq:mfg_systimedependU1} by solving the following optimization problem:
\begin{align}
\begin{cases}
\inf_{(u, V) \in \mathcal{U} \times \mathcal{V}} & \alpha_u\|u\|_{\mathcal{U}}^2 +\alpha_v \|V\|_{\mathcal{V}}^2+ \alpha_{v^o}| [\Psi, V] - \boldsymbol{V}^o|^2 \\
\text{s.t.} & -\partial_t u(x_j, t_j) - \nu \Delta u(x_j, t_j) + \boldsymbol{Q}^{(k)}(x_j, t_j) \cdot \nabla u(x_j, t_j)\\
&\quad\quad\quad\quad = L(x_j, t_j, \boldsymbol{Q}^{(k)}(x_j, t_j)) + V(x_j, t_j) + F(m^{(k)}(x_j, t_j)), \quad \forall j = 1, \dots, \mathfix{M_{\Omega}^{\mathrm{HJB}}}, \\
& u(x_j, T) = U_T(x_j), \quad \forall j = \mathfix{M_{\Omega}^{\mathrm{HJB}}+1}, \dots, \mathfix{M^{\mathrm{HJB}}}. 
\end{cases}
\label{OptHJBtdProb_Cptu}
\end{align}
Here, \(\alpha_u\), \(\alpha_v\), and \(\alpha_{v^o}\) are \mathfix{nonnegative} regularization parameters. Analogous to the stationary case, we select \((\alpha_u, \alpha_v, \alpha_{v^o})=(\frac{1}{2},0,0)\) for the forward problem, where \(V\) is known exactly. \mathfix{As in the stationary case, $V$ is then prescribed rather than optimized: it is fixed at its known value, and the minimization is taken only over $u$, the joint optimization over $(u,V)$ being used only in the inverse problem.} For the inverse problem, a common choice is \(\alpha_u=\frac{1}{2}\), \(\alpha_v=\frac{1}{2}\). \mathfix{If the objective is written as $\alpha_{v^o}\|[\Psi, V] - \boldsymbol{V}^o\|^2$, then $\alpha_{v^o} = \frac{1}{2\gamma_v^2}$ for Gaussian noise with variance $\gamma_v^2$, or we can treat it as a tunable regularization hyperparameter. 
}

\mathfix{Furthermore, the optimization problem in \eqref{OptHJBtdProb_Cptu} is quadratic with linear constraints. The explicit formula for the solution is given in Appendix~\ref{Appendix4}.}

\textbf{Step 3}. \mathfix{Next, we update the policy $\boldsymbol{Q}$, which is a $d$-dimensional vector-valued function. To find $\boldsymbol{Q}$, we first update the values of $\boldsymbol{Q}$ at the collocation points and get $\boldsymbol{Q}^{(k+1)}$ by means of the GP conditioned on the observations of $\boldsymbol{Q}$ at the new values of $\boldsymbol{Q}$.} More precisely, let $\chi = \mathfix{\{(x_1, t_1), \dots, (x_{M_\Omega^{\mathrm{HJB}}}, t_{M_\Omega^{\mathrm{HJB}}})\}}$ be the collection of HJB interior collocation points on $\mathbb{T}^d \times (0, T)$. Then, we compute 
\begin{align*}
\mathbf{q}^{k+1, i}=\arg \max _{\|\boldsymbol{q}\| \le \mathfix{R_{\mathrm{pol}}}}\left\{\boldsymbol{q} \cdot \nabla u^{(k)}(x_i, t_i)-L(x_i, t_i, \boldsymbol{q})\right\},  \quad \forall i=1, \dots, \mathfix{M_{\Omega}^{\mathrm{HJB}}}.
\end{align*}
{Consider a vector-valued GP \(\boldsymbol{\xi}_{\boldsymbol{Q}} : \mathbb{T}^d \times (0, T) \to \mathbb{R}^d\), for all \((x, t) \in \mathbb{T}^d \times (0, T)\), with zero mean, that is, \(\mathbb{E}[\boldsymbol{\xi}_{\boldsymbol{Q}}(x, t)] = \mathbf{0}\). Its covariance is described by a matrix-valued kernel \(\boldsymbol{K}_{\boldsymbol{Q}}((x, t), (y, s)) \in \mathbb{R}^{d \times d}\), for all \((x, t), (y, s) \in \mathbb{T}^d \times (0, T)\), where each  \(\boldsymbol{K}_{\boldsymbol{Q}}((x, t), (y, s))\) encodes both the variances of and cross-covariances between the \(d\) components of \(\boldsymbol{\xi}_{\boldsymbol{Q}}(x, t)\) and \(\boldsymbol{\xi}_{\boldsymbol{Q}}(y, s)\). 
Given the set \(\chi = \mathfix{\{(x_1, t_1), \dots, (x_{M_{\Omega}^{\mathrm{HJB}}}, t_{M_{\Omega}^{\mathrm{HJB}}})\}}\) of collocation points, one assembles \(\boldsymbol{K}_{\boldsymbol{Q}}(\chi, \chi) \in \mathbb{R}^{dM_{\Omega}^{\mathrm{HJB}} \times dM_{\Omega}^{\mathrm{HJB}}}\) by placing each \(\boldsymbol{K}_{\boldsymbol{Q}}((x_i, t_i), (x_j, t_j)) \in \mathbb{R}^{d \times d}\) as the \((i, j)\)-th block in a grid, and forms \(\boldsymbol{K}_{\boldsymbol{Q}}((x, t), \chi) \in \mathbb{R}^{d \times (dM_{\Omega}^{\mathrm{HJB}})}\) by horizontally concatenating \(\boldsymbol{K}_{\boldsymbol{Q}}((x, t), (x_j, t_j))\) for \(j = 1, \dots, M_{\Omega}^{\mathrm{HJB}}\).}

Suppose that \(\boldsymbol{\xi}_{\boldsymbol{Q}}(x_i, t_i) = \mathbf{q}^{k+1, i} \in \mathbb{R}^d\) are the observed outputs at the collocation points \((x_i, t_i)\), for \(i = 1, \dots, \mathfix{M_{\Omega}^{\mathrm{HJB}}}\). We define $ 
\mathbf{q}^{k+1} = \begin{pmatrix}
\mathbf{q}^{k+1, 1}\\ \vdots\\  
\mathbf{q}^{k+1, M_{\Omega}^{\mathrm{HJB}}}
\end{pmatrix}.$
Then, we update \(\boldsymbol{Q}\) by the posterior mean of \(\boldsymbol{\xi}_{\boldsymbol{Q}}\), i.e., \(\boldsymbol{Q}^{(k+1)} = \mathbb{E}[\boldsymbol{\xi}_{\boldsymbol{Q}} | \boldsymbol{\xi}_{\boldsymbol{Q}}(x_i, t_i) = \mathbf{q}^{k+1, i}, i = 1, \dots, M_{\Omega}^{\mathrm{HJB}}]\). Thus,
\begin{align}
\label{eq:Qk1}
\boldsymbol{Q}^{(k+1)}(x, t) = \boldsymbol{K}_{\boldsymbol{Q}}((x, t), \chi) \boldsymbol{K}_{\boldsymbol{Q}}(\chi, \chi)^{-1} \mathbf{q}^{k+1}, \quad \forall (x, t)\in \mathbb{T}^d \times(0, T).
\end{align}
In Step 1, when computing the divergence of \(\boldsymbol{Q}^{(k+1)}\), we differentiate the explicit expression given in \eqref{eq:Qk1}.   \mathfix{The procedure is run until the prescribed stopping criterion is met. In the numerical implementation, we terminate the iterations when the policy stabilizes.}

\section{GPPI Frameworks with the additive Schwarz Newton Acceleration}
\label{Newton Acceleration}
In this section, we incorporate the additive Schwarz Newton acceleration method into the GPPI framework to accelerate solvers for both forward and inverse HJB and MFG problems. PI methods typically exhibit linear or superlinear convergence. The Newton method proposed in \cite{cacace2021policy} reduces the number of iterations compared to classical PI, but its convergence can be non-monotone and may fail if the initial guess is far from the true solution. Moreover, directly extending Newton's method to solve inverse problems in HJBs and MFGs is not straightforward. To address these limitations, we adopt recent nonlinear preconditioning techniques, specifically the additive Schwarz Newton approach \cite{cai2002nonlinearly,dolean2016nonlinear}, within the GPPI frameworks proposed in Section~\ref{sec:gppi_frameworks}. The resulting unified iterative scheme is designed to accelerate the outer iterations in practice.

\subsection{The Additive Schwarz Newton Method} For clarity of presentation, we adopt an abstract formulation for the forward problem of time-dependent MFGs. Analogous constructions apply to HJB problems, stationary MFGs, and the corresponding inverse formulations. Conceptually, the proposed framework reformulates the policy iteration as a nonlinear fixed-point mapping and employs the additive Schwarz Newton method to solve the corresponding fixed-point equation. For illustration, Figure~\ref{fig:gppi_as_arch_final} provides a schematic overview of this architecture. We now proceed to detail the implementation of each component.

\begin{figure}[htbp]
\centering
\resizebox{0.95\textwidth}{!}{
\begin{tikzpicture}[
    font=\sffamily,
    >=Stealth,
    node distance=1.4cm, 
    var/.style = {
        circle, 
        draw=black!80, 
        thick, 
        fill=white, 
        minimum size=1.4cm, 
        inner sep=0pt, 
        text centered,
        font=\Large\bfseries, 
        drop shadow={opacity=0.1, shadow xshift=1pt, shadow yshift=-1pt}
    },
    submodule/.style = {
        rectangle, 
        rounded corners=4pt, 
        draw=blue!40, 
        fill=blue!10, 
        thick, 
        minimum height=1.3cm, 
        minimum width=2.0cm, 
        align=center, 
        font=\normalsize\bfseries, 
        text=blue!60!black
    },
    gppi_frame/.style = {
        rectangle, 
        rounded corners=8pt, 
        draw=blue!50, 
        fill=blue!5, 
        thick, 
        inner sep=0.5cm 
    },
    as_solver/.style = {
        rectangle, 
        rounded corners=8pt, 
        draw=orange!40, 
        fill=orange!5, 
        thick, 
        minimum height=2.0cm, 
        minimum width=4.8cm, 
        align=center,
        font=\large, 
        drop shadow={opacity=0.15}
    },
    op/.style = {
        circle, 
        draw=gray!50, 
        fill=white, 
        thick, 
        inner sep=0pt, 
        minimum size=0.8cm, 
        font=\huge\bfseries, 
        text=gray!70
    },
    decision/.style = {
        diamond,
        aspect=2.2,
        draw=red!50,
        fill=red!5,
        thick,
        align=center,
        font=\normalsize, 
        inner sep=1pt,
        minimum width=3.5cm 
    },
    term/.style = {
        rectangle,
        rounded corners=4pt, 
        draw=green!60!black,
        fill=green!5, 
        thick,
        align=center,
        font=\normalsize\bfseries, 
        minimum height=1.1cm,
        minimum width=2.2cm,
        drop shadow={opacity=0.1}
    },
    link/.style = {
        ->, 
        draw=black!70, 
        line width=1.5pt, 
        rounded corners=10pt
    },
    txt/.style = {
        font=\normalsize, 
        text=black!80, 
        midway,
        align=center,
        fill=white, 
        inner sep=2pt
    }
]

    \node (wk) [var] {$\boldsymbol{w}^{(k)}$};

    \node (fp) [submodule, right=2.8cm of wk] {Update\\ $\boldsymbol m$};
    \node (hjb) [submodule, right=0.6cm of fp] {Update\\ $\boldsymbol u$};
    \node (pol) [submodule, right=0.6cm of hjb] {Update\\ $\mathbf{q}$};
    
    \node (diff) [op, right=1.6cm of pol] {$-$};
    \node [below=0.1cm of diff, font=\small, text=black!60] {Residual};

    \node (as) [as_solver, below=1.8cm of diff] {
        \textbf{Newton Step}\\
        $\Delta \boldsymbol{w} = -\left( \frac{\dd \Phi}{\dd \boldsymbol{w}} \right)^{-1} \Phi(\boldsymbol{w})$
    };

    \node (check) [decision, left=1.2cm of as] {Converged?\\ $\|\Delta \boldsymbol{w}_i\| < \text{tol}$};

    \node (add) [op, left=1.2cm of check] {$+$};
    \node [below=0.1cm of add, font=\small, text=black!60] {Update};

    \node (wk1) [var] at (wk |- as) {$\boldsymbol{w}^{(k+1)}$};

    \node (stop) [term, below=0.8cm of check] {Output\\ $\boldsymbol{w}^* \approx \boldsymbol{w}^{(k)}$};

    \begin{scope}[on background layer]
        \node (gppi_box) [gppi_frame, fit=(fp) (pol), 
            label={[anchor=south, yshift=0pt, black, font=\large\bfseries]north:Component-wise Solvers $\mathcal{L}(\cdot)$}] {};
    \end{scope}

    \draw [link] (wk) -- node[above, txt, yshift=10pt, xshift=-4pt] {$\boldsymbol{w} \equiv (\boldsymbol m, \boldsymbol u, \mathbf{q})$} (gppi_box.west);
    
    \draw [link] (gppi_box.east) -- node[above, txt] {$\mathcal{L}(\boldsymbol{w})$} (diff);

    \draw [link] (diff) -- node[right, txt] {$\Phi(\boldsymbol{w})$} (as);

    \draw [link] (as) -- node[above, txt] {$\Delta \boldsymbol{w}$} (check);

    \draw [link] (check.west) -- node[above, txt] {No} (add.east);

    \draw [link] (check.south) -- node[right, txt] {Yes} (stop.north);

    \draw [link] (add) -- (wk1);

    \draw [link] (wk.north) -- ++(0, 1.4) -| node[pos=0.25, above, txt] {Identity Map} (diff.north);

    \draw [link] (wk.south) -- ++(0, -1.0) -| (add.north);

    \draw [link, dashed, draw=black!70, line width=2pt] (wk1.west) -- ++(-0.6, 0) |- node[pos=0.5, left, font=\large, text=black!80] {Next Iteration} (wk.west);

\end{tikzpicture}
}
\caption{
{The architecture of the additive Schwarz Newton framework applied to MFGs. The independent component-wise solvers first generate the mapping $\mathcal{L}(\boldsymbol{w})$ in parallel. Solving the forward or the inverse problem corresponds to finding a fixed point of $\mathcal{L}$, which is reformulated as finding the zero of the residual $\Phi(\boldsymbol{w}) := \boldsymbol{w} - \mathcal{L}(\boldsymbol{w})$. This root-finding problem is then solved using Newton's method. Convergence is determined by checking if the increment of specific components of $\boldsymbol{w}$ (i.e., $\boldsymbol m$, $\boldsymbol u$, or $\mathbf{q}$) falls below a tolerance.}
}
\label{fig:gppi_as_arch_final}
\end{figure}

{Let $\boldsymbol m$, $\boldsymbol u$ and $\mathbf{q}$ denote, respectively, the vectors of values of certain linear operators acting on the functions $m$, $u$, and $\boldsymbol{Q}$ at the collocation points. For instance, when solving the time-dependent MFG system \eqref{eq:MFG_Systemt}, 
one may take points 
$\{(x_i,t_i)\}_{i=1}^{M^{\mathrm{FP}}}$ and 
$\{(x_j,t_j)\}_{j=1}^{M^{\mathrm{HJB}}}$ for the FP and HJB equations, respectively, and define}
\[
{
\mathbf{q} = \bigl([\delta_{(x_j,t_j)},\,\boldsymbol{Q}]\bigr)_{j=1}^{M_{\Omega}^{\mathrm{HJB}}},
}
\]
\[
\begin{aligned}
\boldsymbol m
={}&\left([\delta_{(x_i,t_i)},\,m])_{i=1}^{M^{\mathrm{FP}}},
([\delta_{(x_i,t_i)}\circ\partial_t,\,m])_{i=1}^{M_\Omega^{\mathrm{FP}}},
([\delta_{(x_i,t_i)}\circ\nabla,\,m])_{i=1}^{M_\Omega^{\mathrm{FP}}},\right.\\
&\left.([\delta_{(x_i,t_i)}\circ\Delta,\,m])_{i=1}^{M_\Omega^{\mathrm{FP}}},
\mathfix{([\delta_{(x_j,t_j)},\,m])_{j=1}^{M_{\Omega}^{\mathrm{HJB}}}}\right)\!,
\end{aligned}
\]
\[
\boldsymbol u
=\left
([\delta_{(x_j,t_j)},\,u])_{j=1}^{M^{\mathrm{HJB}}}, 
([\delta_{(x_j,t_j)}\circ\partial_t,\,u])_{j=1}^{M_{\Omega}^{\mathrm{HJB}}}, 
([\delta_{(x_j,t_j)}\circ\nabla,\,u])_{j=1}^{M_{\Omega}^{\mathrm{HJB}}}, 
([\delta_{(x_j,t_j)}\circ\Delta,\,u])_{j=1}^{M_{\Omega}^{\mathrm{HJB}}}
\right).
\]
{A concrete construction of these vectors for the time-dependent MFG system is given in \Cref{Appendix4}. Building on these observations, the representer theorem \cite{owhadi2019operator} yields explicit GP updates for $m$ and $u$. The policy update for $\boldsymbol{Q}$ is performed pointwise through a low-dimensional maximization, which is explicit for standard quadratic costs and otherwise reduces to an inexpensive local solve.  Let \(R_1\), \(R_2\), and \(R_3\) denote the first-order optimality systems for the optimization problems associated with solving the FP equation, the HJB equation, and the policy-map equation, respectively.}

 Denote \(\boldsymbol w = (\boldsymbol m, \boldsymbol u, \mathbf{q})\) and
$
R(\boldsymbol w) = (R_1(\boldsymbol w), R_2(\boldsymbol w), R_3(\boldsymbol w))
$. We therefore introduce three update maps $\mathcal L_1$, $\mathcal L_2$, and $\mathcal L_3$ such that 
\begin{align}
\label{eq:solutioneq}
R_1\bigl(\mathcal L_1(\boldsymbol w),\,\boldsymbol u,\,\mathbf{q}\bigr)=0,\quad
R_2\bigl(\boldsymbol m,\,\mathcal L_2(\boldsymbol w),\,\mathbf{q}\bigr)=0,\quad
R_3\bigl(\boldsymbol m,\,\boldsymbol u,\,\mathcal L_3(\boldsymbol w)\bigr)=0,
\end{align}
where $\mathcal L_1$, $\mathcal L_2$, and $\mathcal L_3$ each solve for one updated vector while holding the other two fixed. Thus, the PI is to find the fixed point of the equation:
\begin{align}
\label{PolicyFixedPoint}
\Phi(\boldsymbol{w}) = \boldsymbol{w} - (\mathcal{L}_1(\boldsymbol{w}), \mathcal{L}_2(\boldsymbol{w}), \mathcal{L}_3(\boldsymbol{w})). 
\end{align} 
For instance, after eliminating the linear equality constraints, the optimization problem~\eqref{eq:opt:td:fp} can be abstracted as the quadratic program 
\begin{align}
\label{eq:abstct:opt:FP}
\min_{\boldsymbol{m}}\; \big(\Xi(\mathbf{q})\boldsymbol{m} + \boldsymbol{y}(\mathbf{q})\big)^{T}\Gamma^{-1}\big(\Xi(\mathbf{q})\boldsymbol{m} + \boldsymbol{y}(\mathbf{q})\big),
\end{align}
\mathfix{where the matrix $\Xi(\mathbf{q})$ and the vector $\boldsymbol{y}(\mathbf{q})$ depend only on $\mathbf{q}$ after this reduction. The weighting matrix \(\Gamma\) is block-diagonal. For forward problems, \(\Gamma\) reduces to the covariance of the unknowns. For inverse problems, it is augmented by the data-noise covariance. Using the associated first-order optimality condition of \eqref{eq:abstct:opt:FP}, we define  }
\begin{align}
\label{eq:R1:L1:opt}
R_1(\boldsymbol{w}) &:= \Xi(\mathbf{q})^{T}\Gamma^{-1}\Xi(\mathbf{q})\,\boldsymbol{m} + \Xi(\mathbf{q})^{T}\Gamma^{-1}\boldsymbol{y}(\mathbf{q}),\\
\mathcal{L}_1(\boldsymbol{w}) &:= -\big(\Xi(\mathbf{q})^{T}\Gamma^{-1}\Xi(\mathbf{q})\big)^{-1}\Xi(\mathbf{q})^{T}\Gamma^{-1}\boldsymbol{y}(\mathbf{q}). 
\end{align}
Analogous arguments apply to \(R_2\) and \(R_3\).

{To accelerate this PI, we use Newton's method to solve \eqref{PolicyFixedPoint}. The corresponding Newton step satisfies}
\begin{align}
\label{PolicyNewton}
\boldsymbol{w}^{(k+1)} = \boldsymbol{w}^{(k)} + \Delta \boldsymbol{w}^{(k)}, \quad -\frac{\mathrm{d} \Phi(\boldsymbol{w}^{(k)})}{\mathrm{d} \boldsymbol{w}}\Delta \boldsymbol{w}^{(k)} = \Phi(\boldsymbol{w}^{(k)}).  
\end{align}
{where $\frac{\mathrm{d}\Phi(\boldsymbol w)}{\mathrm{d}\boldsymbol w}$ denotes the Jacobian of $\Phi$, namely}
\begin{align*}
\frac{\mathrm{d} \Phi(\boldsymbol{w})}{\mathrm{d} \boldsymbol{w}} = I - \begin{pmatrix}
\frac{\mathrm{d} \mathcal{L}_1(\boldsymbol{w})}{\mathrm{d} \boldsymbol{w}} \\ \frac{\mathrm{d} \mathcal{L}_2(\boldsymbol{w})}{\mathrm{d} \boldsymbol{w}} \\ \frac{\mathrm{d} \mathcal{L}_3(\boldsymbol{w})}{\mathrm{d} \boldsymbol{w}} 
\end{pmatrix}
=I-
\begin{pmatrix}
\frac{\partial \mathcal{L}_1}{\partial \boldsymbol{m}} & \frac{\partial \mathcal{L}_1}{\partial \boldsymbol{u}} & \frac{\partial \mathcal{L}_1}{\partial \mathbf{q}} \\ 
\frac{\partial \mathcal{L}_2}{\partial \boldsymbol{m}} & \frac{\partial \mathcal{L}_2}{\partial \boldsymbol{u}} & \frac{\partial \mathcal{L}_2}{\partial \mathbf{q}} \\ 
\frac{\partial \mathcal{L}_3}{\partial \boldsymbol{m}} & \frac{\partial \mathcal{L}_3}{\partial \boldsymbol{u}} & \frac{\partial \mathcal{L}_3}{\partial \mathbf{q}}
\end{pmatrix}.
\end{align*}

\noindent It remains to compute the Jacobian of \(\mathcal{L}_i\) for each \(i\). Define the half-step states
{
\[
\widehat{\boldsymbol{w}}_1(\boldsymbol{w})=(\mathcal L_1(\boldsymbol w),\boldsymbol u,\mathbf{q}), \qquad
\widehat{\boldsymbol{w}}_2(\boldsymbol{w})=(\boldsymbol m,\mathcal L_2(\boldsymbol w),\mathbf{q}), \qquad
\widehat{\boldsymbol{w}}_3(\boldsymbol{w})=(\boldsymbol m,\boldsymbol u,\mathcal L_3(\boldsymbol w)).\]}

\noindent Differentiating the first equation in \eqref{eq:solutioneq} at \(\widehat{\boldsymbol{w}}_1(\boldsymbol{w})\), we obtain
\begin{align*}
\frac{\mathrm{d} R_1}{\mathrm{d} \boldsymbol{m}}\bigl(\widehat{\boldsymbol{w}}_1(\boldsymbol{w})\bigr)\frac{\mathrm{d} \mathcal{L}_1}{\mathrm{d} \boldsymbol{w}} + \frac{\mathrm{d} R_1}{\mathrm{d} \boldsymbol{u}}\bigl(\widehat{\boldsymbol{w}}_1(\boldsymbol{w})\bigr)\frac{\mathrm{d} \boldsymbol{u}}{\mathrm{d} \boldsymbol{w}} + \frac{\mathrm{d} R_1}{\mathrm{d} \mathbf{q}}\bigl(\widehat{\boldsymbol{w}}_1(\boldsymbol{w})\bigr)\frac{\mathrm{d} \mathbf{q}}{\mathrm{d} \boldsymbol{w}} = 0. 
\end{align*}
Solving for the derivative of \(\mathcal{L}_1\) and 
using the identities \(\frac{\mathrm{d} \boldsymbol{u}}{\mathrm{d} \boldsymbol{w}} = [0, I, 0]\) and \(\frac{\mathrm{d} \mathbf{q}}{\mathrm{d} \boldsymbol{w}} = [0, 0, I]\), we have
\[
\frac{\mathrm{d} \mathcal{L}_1}{\mathrm{d} \boldsymbol{w}} = \left[0,\ - \left(\frac{\mathrm{d} R_1}{\mathrm{d} \boldsymbol{m}}\bigl(\widehat{\boldsymbol{w}}_1(\boldsymbol{w})\bigr)\right)^{-1} \frac{\mathrm{d} R_1}{\mathrm{d} \boldsymbol{u}}\bigl(\widehat{\boldsymbol{w}}_1(\boldsymbol{w})\bigr),\ - \left(\frac{\mathrm{d} R_1}{\mathrm{d} \boldsymbol{m}}\bigl(\widehat{\boldsymbol{w}}_1(\boldsymbol{w})\bigr)\right)^{-1} \frac{\mathrm{d} R_1}{\mathrm{d} \mathbf{q}}\bigl(\widehat{\boldsymbol{w}}_1(\boldsymbol{w})\bigr) \right].
\]
Similarly, we obtain
\[
\frac{\mathrm{d} \mathcal{L}_2}{\mathrm{d} \boldsymbol{w}} = \left[- \left(\frac{\mathrm{d} R_2}{\mathrm{d} \boldsymbol{u}}\bigl(\widehat{\boldsymbol{w}}_2(\boldsymbol{w})\bigr)\right)^{-1} \frac{\mathrm{d} R_2}{\mathrm{d} \boldsymbol{m}}\bigl(\widehat{\boldsymbol{w}}_2(\boldsymbol{w})\bigr),\ 0,\ - \left(\frac{\mathrm{d} R_2}{\mathrm{d} \boldsymbol{u}}\bigl(\widehat{\boldsymbol{w}}_2(\boldsymbol{w})\bigr)\right)^{-1} \frac{\mathrm{d} R_2}{\mathrm{d} \mathbf{q}}\bigl(\widehat{\boldsymbol{w}}_2(\boldsymbol{w})\bigr) \right],
\]
\[
\frac{\mathrm{d} \mathcal{L}_3}{\mathrm{d} \boldsymbol{w}} = \left[- \left(\frac{\mathrm{d} R_3}{\mathrm{d} \mathbf{q}}\bigl(\widehat{\boldsymbol{w}}_3(\boldsymbol{w})\bigr)\right)^{-1} \frac{\mathrm{d} R_3}{\mathrm{d} \boldsymbol{m}}\bigl(\widehat{\boldsymbol{w}}_3(\boldsymbol{w})\bigr),\ - \left(\frac{\mathrm{d} R_3}{\mathrm{d} \mathbf{q}}\bigl(\widehat{\boldsymbol{w}}_3(\boldsymbol{w})\bigr)\right)^{-1} \frac{\mathrm{d} R_3}{\mathrm{d} \boldsymbol{u}}\bigl(\widehat{\boldsymbol{w}}_3(\boldsymbol{w})\bigr),\ 0 \right].
\]
Combining the above calculations, the Jacobian of \(\Phi(\boldsymbol{w})\) is given by
\begin{align}
\label{eq:as_update}
\frac{\mathrm{d} \Phi(\boldsymbol{w})}{\mathrm{d} \boldsymbol{w}}
=&
\left(\underbrace{
\begin{bmatrix}
\frac{\mathrm{d} R_1}{\mathrm{d} \boldsymbol{m}}\bigl(\widehat{\boldsymbol{w}}_1(\boldsymbol{w})\bigr) & 0 & 0 \\
0 & \frac{\mathrm{d} R_2}{\mathrm{d} \boldsymbol{u}}\bigl(\widehat{\boldsymbol{w}}_2(\boldsymbol{w})\bigr) & 0 \\
0 & 0 & \frac{\mathrm{d} R_3}{\mathrm{d} \mathbf{q}}\bigl(\widehat{\boldsymbol{w}}_3(\boldsymbol{w})\bigr)
\end{bmatrix}
}_{J_{\mathrm{AS}}}\right)^{\mathfix{-1}}\mkern-7mu
\underbrace{
\begin{bmatrix}
\frac{\mathrm{d} R_1}{\mathrm{d} \boldsymbol{m}}\bigl(\widehat{\boldsymbol{w}}_1(\boldsymbol{w})\bigr) &
\frac{\mathrm{d} R_1}{\mathrm{d} \boldsymbol{u}}\bigl(\widehat{\boldsymbol{w}}_1(\boldsymbol{w})\bigr) &
\frac{\mathrm{d} R_1}{\mathrm{d} \mathbf{q}}\bigl(\widehat{\boldsymbol{w}}_1(\boldsymbol{w})\bigr) \\[2pt]
\frac{\mathrm{d} R_2}{\mathrm{d} \boldsymbol{m}}\bigl(\widehat{\boldsymbol{w}}_2(\boldsymbol{w})\bigr) &
\frac{\mathrm{d} R_2}{\mathrm{d} \boldsymbol{u}}\bigl(\widehat{\boldsymbol{w}}_2(\boldsymbol{w})\bigr) &
\frac{\mathrm{d} R_2}{\mathrm{d} \mathbf{q}}\bigl(\widehat{\boldsymbol{w}}_2(\boldsymbol{w})\bigr) \\[2pt]
\frac{\mathrm{d} R_3}{\mathrm{d} \boldsymbol{m}}\bigl(\widehat{\boldsymbol{w}}_3(\boldsymbol{w})\bigr) &
\frac{\mathrm{d} R_3}{\mathrm{d} \boldsymbol{u}}\bigl(\widehat{\boldsymbol{w}}_3(\boldsymbol{w})\bigr) &
\frac{\mathrm{d} R_3}{\mathrm{d} \mathbf{q}}\bigl(\widehat{\boldsymbol{w}}_3(\boldsymbol{w})\bigr)
\end{bmatrix}
}_{R'_{\mathrm{AS,row}}}
\nonumber\\
=&\,
J_{\mathrm{AS}}^{-1}R'_{\mathrm{AS,row}}.
\end{align}
{Hence, according to \eqref{PolicyNewton}, at each step we solve}
\begin{align*}
\mathfix{-R'_{\mathrm{AS,row}}(\boldsymbol w)\,\Delta \boldsymbol{w}
=
J_{\mathrm{AS}}(\boldsymbol w) \Phi(\boldsymbol{w}).}
\end{align*}
{
The procedure for the additive Schwarz Newton method is outlined in Algorithm~\ref{alg:example}. In the numerical experiments, we observe that the additive Schwarz Newton method requires fewer iterations than the GPPI method. In practice, since each iteration admits explicit update formulas, as discussed in the previous section, assembling the components of the row-wise matrices \(J_{\mathrm{AS}}^k\) and \(R_{\mathrm{AS,row}}^{\prime,k}\) is straightforward. The main computational bottleneck lies in assembling and solving the resulting linear system. The key structural simplification is that several blocks of the assembled Jacobian vanish. For example, in the FP case, \(\mathcal{L}_1\) in \eqref{eq:R1:L1:opt} depends only on \(\mathbf{q}\), so
\[
\frac{\partial \mathcal{L}_1}{\partial \boldsymbol{m}}=\frac{\partial \mathcal{L}_1}{\partial \boldsymbol{u}}=\boldsymbol{0}.
\]
}

\begin{algorithm}[h]
\caption{Additive Schwarz Newton Method}
\label{alg:example}
\begin{algorithmic}  % The "[1]" is to number each line
\REQUIRE Input parameters, $\boldsymbol{m}^{(0)}$, $\boldsymbol{u}^{(0)}$, and $\mathbf{q}^{(0)}$, and the number of iterations $\widetilde{K}$
\ENSURE Output $\boldsymbol{m}^{(\widetilde{K})}, \boldsymbol{u}^{(\widetilde{K})}, \mathbf{q}^{(\widetilde{K})}$
\STATE Initialize variables and parameters
\FOR{{$k = 0, \ldots, \widetilde{K}-1$}}
    \STATE Given $\boldsymbol{w}^{(k)} = (\boldsymbol{m}^{(k)}, \boldsymbol{u}^{(k)}, \mathbf{q}^{(k)})$, we want to solve $R_1(\boldsymbol{m}, \boldsymbol{u}^{(k)}, \mathbf{q}^{(k)})=0$,  $R_2(\boldsymbol{m}^{(k)}, \boldsymbol{u}, \mathbf{q}^{(k)})=0$, and $R_3(\boldsymbol{m}^{(k)}, \boldsymbol{u}^{(k)}, \mathbf{q})=0$ to obtain $(\boldsymbol{m}^{(k+1/2)}, \boldsymbol{u}^{(k+1/2)}, \mathbf{q}^{(k+1/2)})$, compute $\Phi(\boldsymbol{w}^{(k)}) = \boldsymbol{w}^{(k)} - (\boldsymbol{m}^{(k+1/2)}, \boldsymbol{u}^{(k+1/2)}, \mathbf{q}^{(k+1/2)})$, and form
    \begin{align*}
    {J_{\mathrm{AS}}^k =} \begin{bmatrix}
    \frac{\mathrm{d} R_1}{\mathrm{d} \boldsymbol{m}}(\boldsymbol{m}^{(k+1/2)}, \boldsymbol{u}^{(k)}, \mathbf{q}^{(k)}) & 0 & 0 \\
    0 & \frac{\mathrm{d} R_2}{\mathrm{d} \boldsymbol{u}}(\boldsymbol{m}^{(k)}, \boldsymbol{u}^{(k+1/2)}, \mathbf{q}^{(k)}) & 0 \\
    0 & 0 & \frac{\mathrm{d} R_3}{\mathrm{d} \mathbf{q}}(\boldsymbol{m}^{(k)}, \boldsymbol{u}^{(k)}, \mathbf{q}^{(k+1/2)})
    \end{bmatrix}
    \end{align*}
    \begin{align*}
    {\mathfix{R_{\mathrm{AS,row}}^{\prime,k} =}} \begin{bmatrix}
    \frac{\mathrm{d} R_1}{\mathrm{d} \boldsymbol{m}}(\boldsymbol{m}^{(k+1/2)}, \boldsymbol{u}^{(k)}, \mathbf{q}^{(k)}) & \frac{\mathrm{d} R_1}{\mathrm{d} \boldsymbol{u}}(\boldsymbol{m}^{(k+1/2)}, \boldsymbol{u}^{(k)}, \mathbf{q}^{(k)}) & \frac{\mathrm{d} R_1}{\mathrm{d} \mathbf{q}}(\boldsymbol{m}^{(k+1/2)}, \boldsymbol{u}^{(k)}, \mathbf{q}^{(k)}) \\
    \frac{\mathrm{d} R_2}{\mathrm{d} \boldsymbol{m}}(\boldsymbol{m}^{(k)}, \boldsymbol{u}^{(k+1/2)}, \mathbf{q}^{(k)}) & \frac{\mathrm{d} R_2}{\mathrm{d} \boldsymbol{u}}(\boldsymbol{m}^{(k)}, \boldsymbol{u}^{(k+1/2)}, \mathbf{q}^{(k)}) & \frac{\mathrm{d} R_2}{\mathrm{d} \mathbf{q}}(\boldsymbol{m}^{(k)}, \boldsymbol{u}^{(k+1/2)}, \mathbf{q}^{(k)}) \\
    \frac{\mathrm{d} R_3}{\mathrm{d} \boldsymbol{m}}(\boldsymbol{m}^{(k)}, \boldsymbol{u}^{(k)}, \mathbf{q}^{(k+1/2)}) & \frac{\mathrm{d} R_3}{\mathrm{d} \boldsymbol{u}}(\boldsymbol{m}^{(k)}, \boldsymbol{u}^{(k)}, \mathbf{q}^{(k+1/2)}) & \frac{\mathrm{d} R_3}{\mathrm{d} \mathbf{q}}(\boldsymbol{m}^{(k)}, \boldsymbol{u}^{(k)}, \mathbf{q}^{(k+1/2)})
    \end{bmatrix}.
    \end{align*}
    {
    \STATE Solve $-R_{\mathrm{AS,row}}^{\prime,k} \Delta \boldsymbol{w}^{(k)} = J_{\mathrm{AS}}^k \Phi(\boldsymbol{w}^{(k)})$. Then, $\boldsymbol{w}^{(k+1)} = \boldsymbol{w}^{(k)} + \Delta \boldsymbol{w}^{(k)}$ 
    }
\ENDFOR
\RETURN $\boldsymbol{w}^{(\widetilde{K})}$ 
\end{algorithmic}
\end{algorithm}

Likewise, by the update rules for the HJB equation and the policy map, $\mathcal{L}_2$ depends only on $(\boldsymbol{m},\mathbf{q})$, whereas $\mathcal{L}_3$ depends only on $\boldsymbol{u}$. Consequently, the Jacobian has the block form
\begin{align*}
\frac{\mathrm{d}\Phi(\boldsymbol{w})}{\mathrm{d}\boldsymbol{w}}
&= I -
\begin{pmatrix}
\displaystyle \frac{\partial \mathcal{L}_1}{\partial \boldsymbol{m}} &
\displaystyle \frac{\partial \mathcal{L}_1}{\partial \boldsymbol{u}} &
\displaystyle \frac{\partial \mathcal{L}_1}{\partial \mathbf{q}} \\
\displaystyle \frac{\partial \mathcal{L}_2}{\partial \boldsymbol{m}} &
\displaystyle \frac{\partial \mathcal{L}_2}{\partial \boldsymbol{u}} &
\displaystyle \frac{\partial \mathcal{L}_2}{\partial \mathbf{q}} \\
\displaystyle \frac{\partial \mathcal{L}_3}{\partial \boldsymbol{m}} &
\displaystyle \frac{\partial \mathcal{L}_3}{\partial \boldsymbol{u}} &
\displaystyle \frac{\partial \mathcal{L}_3}{\partial \mathbf{q}}
\end{pmatrix}
= I -
\begin{pmatrix}
\boldsymbol{0} & \boldsymbol{0} & \displaystyle \frac{\partial \mathcal{L}_1}{\partial \mathbf{q}} \\
\displaystyle \frac{\partial \mathcal{L}_2}{\partial \boldsymbol{m}} & \boldsymbol{0} & \displaystyle \frac{\partial \mathcal{L}_2}{\partial \mathbf{q}} \\
\boldsymbol{0} & \displaystyle \frac{\partial \mathcal{L}_3}{\partial \boldsymbol{u}} & \boldsymbol{0}
\end{pmatrix}.
\end{align*}
Thus, only four of the nine block entries need to be assembled. Moreover, the matrix \(J_{\mathrm{AS}}\) used in Algorithm~\ref{alg:example} appears both in forming these Jacobian blocks and in evaluating \(\Phi\). Thus, it can be cached and reused. 

{Overall, the additional computational overhead introduced by the additive Schwarz acceleration lies in assembling and solving the Jacobian linear system. The dimension of this system grows linearly with the number of collocation points. Crucially, the additive Schwarz construction yields a reduced block-coupling structure: only four of the nine block derivatives are nonzero in the present formulation. Individual nonzero blocks may still be dense because of the global kernel couplings in the GP representer formulas, but the reduced block pattern can still be exploited by block linear solvers, cached factorizations, or matrix-free Jacobian-vector products. The main difference between GPPI and GPPI-AS is the way the variables are updated at each iteration.}
In GPPI, the updates for $\boldsymbol{m}$, $\boldsymbol{u}$, and $\mathbf{q}$ are carried out sequentially within each outer iteration.
In contrast, GPPI-AS first computes the three half-step updates independently and then applies the resulting Newton correction, which makes the assembly stage amenable to parallel implementation.
This structural difference implies that the convergence behavior of GPPI-AS is distinct from that of GPPI and suggests the potential for additional wall-clock time gains if the row-wise Jacobian assembly were implemented in a fully parallel computing environment.

\begin{remark}[Convergence requirements and observed behavior]
We do not claim a global convergence theorem for the GPPI and GPPI-AS algorithms
on the forward and inverse problems. Convergence rests on two ingredients. First,
the underlying exact policy iteration must converge, which holds under standard
regularity of the coefficients and monotonicity of the coupling. We refer to
\cite{cacace2021policy} and \cite{ren2024policy} for convergence analyses of
related policy-iteration schemes for the forward and inverse problems of MFGs,
respectively. Second, the GP interpolation error must vanish as the collocation
points become dense. This requires the solutions to be sufficiently regular
relative to the RKHSs of the chosen kernels, together with a fixed positive
diffusion coefficient that ensures uniform ellipticity. 

Empirically, convergence is robust across the cases with smooth solutions reported in the next section.  As the diffusion coefficient becomes small, however, the problem becomes more challenging, and the solutions may become less regular.  In such regimes, policy iteration itself may require more iterations to converge, and the GPPI approximation may also be affected by the use of fixed collocation points and fixed kernels throughout the iteration.  More collocation points, or points refined in regions where the solution has reduced regularity, may be needed to maintain accuracy.  Adaptive refinement of the collocation points and adaptive kernel selection across the iterations are promising directions for handling these cases more efficiently, which we leave to future work.
\end{remark}

\section{Numerical Experiments}
\label{secNumericalExpe}
 This section details numerical experiments conducted on various MFG forward and inverse problems, as well as HJB inverse problems, to validate our proposed frameworks. 
In \Cref{HJBequationinverseproblem}, we address the inverse problem associated with the HJB equation. \Cref{ForwardProblemGP} focuses on the forward problem for stationary MFGs. In \Cref{GPinverseproblem}, we apply the proposed approach to the inverse problem of stationary MFGs. Lastly, \Cref{Timedependentexperiment} considers the inverse problem in the time-dependent MFG setting. 

{
All reported errors are evaluated on uniform reference grids, while inverse GP reconstructions use nonuniform collocation and observation points. If \(u_\alpha\) and \(v_\alpha\) 
denote the values at the grid point indexed by \(\alpha\in\mathcal{I}\), we define
\begin{align}
\mathcal{E}_1(u,v)
&:= \sum_{\alpha\in\mathcal{I}} w_\alpha |u_{\alpha}-v_{\alpha}|,\notag\\
\mathcal{E}_2(u,v)
&:= \left(\sum_{\alpha\in\mathcal{I}} w_\alpha |u_{\alpha}-v_{\alpha}|^2\right)^{1/2},
\label{eq:l2disc}\\
\mathcal{E}_\infty(u,v)
&:= \max_{\alpha\in\mathcal{I}} |u_{\alpha}-v_{\alpha}|.\notag
\end{align}
Here \(\mathcal{I}\) is the grid index set. The weight \(w_\alpha\) is the grid spacing in the corresponding dimension: 
\(w_\alpha=h_x\) in one dimension, \(w_\alpha=h_xh_y\) in two spatial dimensions, 
and \(w_\alpha=h_xh_t\) on a space-time grid.} In all experiments, the algorithms terminate when the $L^2$ norm of the policy increment falls below the threshold $\text{tol}=10^{-5}$, as illustrated in \Cref{fig:gppi_as_arch_final}.

Moreover, all experiments use Python 3.11.3 with the JAX library and run on a 2023 Mac mini with an Apple M2 processor and 8 GB of RAM.

\subsection{The Inverse Problem for the HJB Equation
}
\label{HJBequationinverseproblem}
In this section, we solve the \mathfix{inverse problem for the HJB equation} using the GPPI framework in Section~\ref{HJB Equation Inverse}.

{We consider a periodic HJB equation with compact controls
and a subquadratic control cost given by}
\begin{equation}
\begin{cases}
-\partial_t U- \frac{1}{2} \sigma^2 \Delta U + \sup_{\boldsymbol{q}\in \mathcal{Q}} \{-\nabla U \cdot (A\sin(2\pi x) + B \boldsymbol{q}) - V(x) - (\boldsymbol{q}^T\mathfix{R_{\mathrm{c}}}\boldsymbol{q})^{\frac{2}{3}} \}   = 0, & \forall (x, t) \in \mathbb{T}^d\times (0,T), \\
U(x, T) =U_T(x), & \forall x \in \mathbb{T}^d,
\end{cases}
\label{eqHJB}
\end{equation}
Given noisy observations of \(U\) and \(V\), our goal is to recover the value function \(U\) and the spatial cost function \(V\) using the GPPI framework.

\textbf{Experimental Setup.}
We consider the one-dimensional case with parameters  \( d = 1 \), \( \mathfix{R_{\mathrm{c}} = (0.4)^{\frac{3}{2}}} \), \( A = 0.1 \), \( B = 0.5 \), and \( \sigma = \sqrt{0.1} \), in the domain $\mathbb{T}\times (0,T]$ identified with $[-0.5, 0.5)\times (0,1]$. {The admissible control set is $\mathcal{Q}=\{q\in\mathbb{R}:|q|\le 100\}$.} In this setting, the true spatial cost function is given by \( V(x) = \mathfix{\frac{3}{16}}\bigl(1-\cos(2\pi x)\bigr) \) and the terminal cost function is given by $U_T(x)=0.5 + \frac{1}{8}\bigl(1-\cos(2\pi x)\bigr)$. The reference solution for \(U\) is obtained using the finite difference method.

\begin{figure}[h]
    \centering    
    \begin{subfigure}[b]{0.23\textwidth}
        \centering
        {\includegraphics[width=\linewidth,page=1]{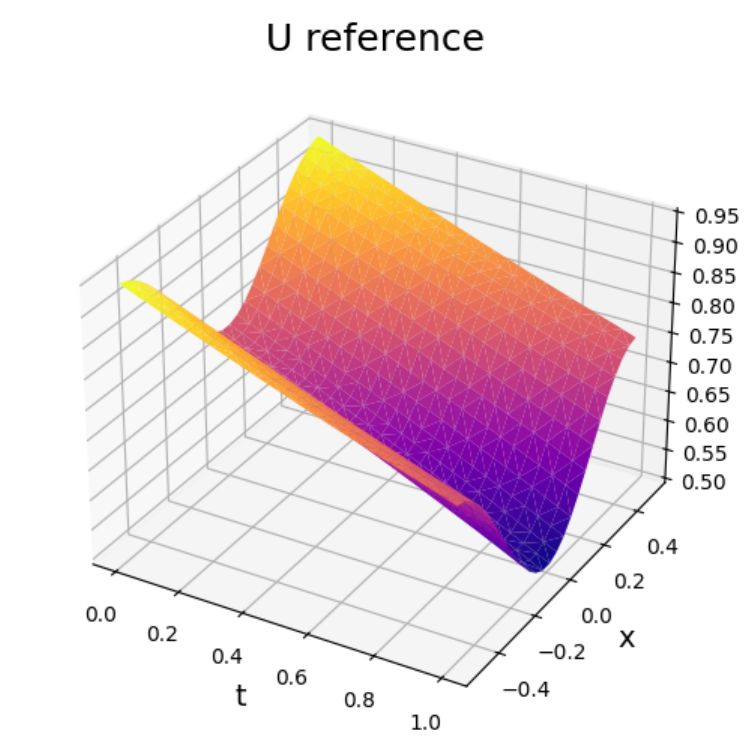}}
        \caption{$U$ reference}
        \label{HJBeqfig1}
    \end{subfigure}%
    \hspace{1mm}
    \begin{subfigure}[b]{0.23\textwidth}
        \centering       {\includegraphics[width=\linewidth,page=2]{figures_combined.pdf}}
        \caption{Recovered $U$ via GPPI}
        \label{HJBeqfig2}
    \end{subfigure}%
    \hspace{1mm}
    \begin{subfigure}[b]{0.23\textwidth}
        \centering
        {\includegraphics[width=\linewidth,page=3]{figures_combined.pdf}}
        \caption{$V$ reference}
        \label{HJBeqfig5}
    \end{subfigure}   
    \hspace{1mm}   
    \begin{subfigure}[b]{0.23\textwidth}
        \centering
        {\includegraphics[width=\linewidth,page=4]{figures_combined.pdf}}
        \caption{Recovered $V$ via GPPI}
        \label{HJBeqfig6}
    \end{subfigure}

    \vspace{2mm}
    
    \begin{subfigure}[b]{0.23\textwidth}
        \centering
        {\includegraphics[width=\linewidth,page=5]{figures_combined.pdf}}
        \caption{Absolute pointwise error contour of $U$ via GPPI}
        \label{HJBeqfig3}
    \end{subfigure}
    \hspace{1mm}
    \begin{subfigure}[b]{0.23\textwidth}
        \centering
        {\includegraphics[width=\linewidth,page=6]{figures_combined.pdf}}
        \caption{Absolute pointwise error of $V$ via GPPI}
        \label{HJBeqfig7}
    \end{subfigure} 
    \hspace{1mm}
    \begin{subfigure}[b]{0.23\textwidth}
        \centering
       {\includegraphics[width=\linewidth,page=7]{figures_combined.pdf}}
        \caption{Samples \& observations for $U$}
        \label{HJBeqfig4}
    \end{subfigure}
    
    \caption{Numerical results for the HJB equation in \eqref{eqHJB}: (a) and (c) show the reference solutions for $U$ and $V$; (b) and (d) show the recovered $U$ and $V$ via the GPPI method under  Gaussian  noise; (e) and (f) show the absolute pointwise errors of $U$ and $V$ via the GPPI method; and (g) shows the sample and observation points for $U$.}
    \label{Numerical results for HJequation}
\end{figure}

{To demonstrate the mesh-free property of the GPPI method, 676 sample points for \mathfix{the HJB equation and the value function \(U\)} are randomly generated in the space-time domain \(\mathbb{T} \times (0,T]\), together with \(M_T=50\) 
  terminal collocation points on
\(\mathbb T\times\{T\}\). From these sample points, 42 points are randomly selected as observations for \(U\), and their observed values are obtained by interpolating from a finer reference solution (computed on a uniform $40 \times 40$ grid by the finite difference method) using cubic interpolation. Meanwhile, 6 observation points for \(V\) are generated in the spatial domain, independent of the sample points.  We set  \(\alpha_{u^o} =  \mathfix{10^5}\), \(\alpha_{v^o} = \mathfix{10^5}\), \(\alpha_{v} = 13 \), \(\alpha_{u} = 0.5 \). In the main reconstruction experiment shown in Figure~\ref{Numerical results for HJequation},  Gaussian noise \(\mathcal{N}(0,\gamma^2 I)\) with \(\gamma = 10^{-3}\) is added to the observations. To evaluate the reconstruction error, the GP solution is interpolated back onto the uniform reference grid via the GP posterior mean and compared with the reference solution. The plotted error panels, therefore, report the absolute pointwise error on that grid.}
We choose the following kernel for the GPs of $U$ and $\boldsymbol{q}$: $K\left((x, t), (x', t'); \boldsymbol{\sigma}\right)
= \mathfix{\exp\left(\frac{\cos\left(2\pi(x-x')\right)-1}{\sigma_1^2}\right)}  \exp\left(-\frac{(t - t')^2}{\sigma_2^2}\right),$
where \(\boldsymbol{\sigma} = (\sigma_1, \sigma_2)=(0.9,0.9)\) are the kernel parameters.
When approximating $V$, we use the periodic kernel \(\mathfix{K\left(x,x^{\prime}; \bar\sigma\right)}=\exp\left(\frac{\cos\left(2\pi(x-x^{\prime})\right)-1}{\bar{\sigma}^{2}}\right)\) with parameter $\bar\sigma=0.6$ for the GP. 

{
\textbf{Experimental Results.}
Figure~\ref{Numerical results for HJequation} presents the experimental results for reconstructing \(U\) and \(V\) for the HJB equation in \eqref{eqHJB}. In this experiment, the observations are perturbed by Gaussian noise \(\mathcal{N}(0,\gamma^2 I)\), and the plotted error fields are the absolute pointwise errors \(|U-U^*|\) and \(|V-V^*|\).

Table~\ref{HJBnoise} reports the relative \(L^1\), \(L^2\), and \(L^\infty\) errors of the reconstructed \(U\) under different noise levels with the same observation points as in the experimental setup. In the experiments in Table~\ref{HJBnoise}, given a clean observation vector \(\boldsymbol{f}\), we perturb it by a noise vector
\begin{equation}
\label{eq:relnoise}
\boldsymbol{\epsilon} \sim \mathcal{N}\!\left(\boldsymbol{0},\, \frac{\tilde{\gamma}^{2}\lVert \boldsymbol{f} \rVert_2^{2}}{n}\, I\right),
\end{equation}
where \(n\) denotes the dimension of the observation vector \(\boldsymbol{f}\). This normalization gives $\frac{\left(\mathbb{E}\lVert \boldsymbol{\epsilon} \rVert_2^2\right)^{1/2}}
{\lVert \boldsymbol{f} \rVert_2}
=
\tilde{\gamma}$.
The relative error is defined using the discrete norms above as
\begin{equation}
\label{eq:relerr}
\mathcal{E}_{\mathrm{rel}}^{p}(f, f^*) =
\frac{\mathcal{E}_{p}(f, f^*)}{\mathcal{E}_{p}(f^*, 0)},
\quad p = 1, 2, \infty,
\end{equation}
where \(f\) denotes the reconstruction and \(f^*\) denotes the reference solution. This normalization factors out the scale of the reference signal when comparing reconstructions across different noise levels. Table~\ref{HJBnoise} shows that the reconstruction accuracy degrades moderately as \(\tilde{\gamma}\) increases.
}

 \begin{table}[H]
\centering
\caption{Relative recovery errors of $U$ under different noise levels $\tilde{\gamma}$ for the HJB equation in \eqref{eqHJB}, under the  Gaussian noise model \eqref{eq:relnoise}.}
\begin{tabular}{lccccc}
\toprule
 $\tilde{\gamma}$ & 0e+00 & 1e-03 & 5e-03 & 1e-02 & 5e-02 \\
\midrule
Relative $L^1$ error     & 5.227e-03 & 5.280e-03 & 5.494e-03 & 5.761e-03 & 7.912e-03 \\
Relative $L^2$ error     & 6.273e-03 & 6.332e-03 & 6.571e-03 & 6.874e-03 & 9.453e-03 \\
Relative $L^\infty$ error & 1.320e-02 & 1.324e-02 & 1.340e-02 & 1.361e-02 & 1.599e-02 \\
\bottomrule
\end{tabular}
\label{HJBnoise}
\end{table}

\subsection{The Stationary MFG Forward Problem}
\label{ForwardProblemGP}
In this subsection, we address the forward problem for stationary MFGs within the framework introduced in Section \ref{MFG Forward Problem Stationary}.  Specifically, we consider the MFG system
\begin{equation}
\begin{cases}
-\nu\,\Delta u + H\bigl(x,\nabla u\bigr) + \lambda = F(m) + V(x), 
& \forall x \in \mathbb{T}^d, \\
-\nu\,\Delta m - \operatorname{div}\bigl(D_p H(x,\nabla u)\,m\bigr) = 0, 
& \forall x \in \mathbb{T}^d, \\
\int_{\mathbb{T}^d} u \,\dif x = 0, 
\quad 
\int_{\mathbb{T}^d} m \,\dif x = 1.
\end{cases}
\label{eq:stationaryMFG}
\end{equation}
Given \(\nu\), \(H\), \(V\), and \(F\), we solve for the distribution \(m\), the value function \(u\), and the constant \(\lambda\).

\textbf{Experimental Setup.} We set \(d = 1\) and identify \(\mathbb{T}\) with the interval \([0, 1)\). The Hamiltonian is chosen as \mathfix{$H(x,p) = \frac{1}{2}|p|^2$}.
 The spatial cost function is \( V(x) = 2(\sin^2(\pi x) + \cos(6\pi x)) \) and the coupling function is \( F(m) = m^4 \). The viscosity coefficient is set to $\nu=0.5$. The grid resolution is set to \( h = \frac{1}{100} \).  The Gaussian regularization coefficients \( \alpha_m,\alpha_u,\alpha_{\lambda} \) are all set to 0.5. The initial values for \(u\), \(\boldsymbol{Q}\), and \(\lambda\) are set to \(0\), while the initial value for \(m\) is set to \(1\). We impose independent GP priors on the unknown functions \(m\), \(u\), and \(\boldsymbol{Q}\), each defined over the torus \(\mathbb{T}^d\).  Specifically, we use the stationary periodic covariance kernel $
K(x,x') = \exp\!\Bigl(-\frac{2\sin^2\!\bigl(\pi(x - x')\bigr)}{\mathfix{\ell_{\mathrm K}}^{\,2}}\Bigr),$
where \(\mathfix{\ell_{\mathrm K}}>0\) is the length-scale.  This choice enforces periodicity and respects the boundary conditions.

\begin{figure}[h]
    \centering
    \begin{subfigure}[b]{0.23\textwidth}
        \centering
        {\includegraphics[width=\linewidth,page=8]{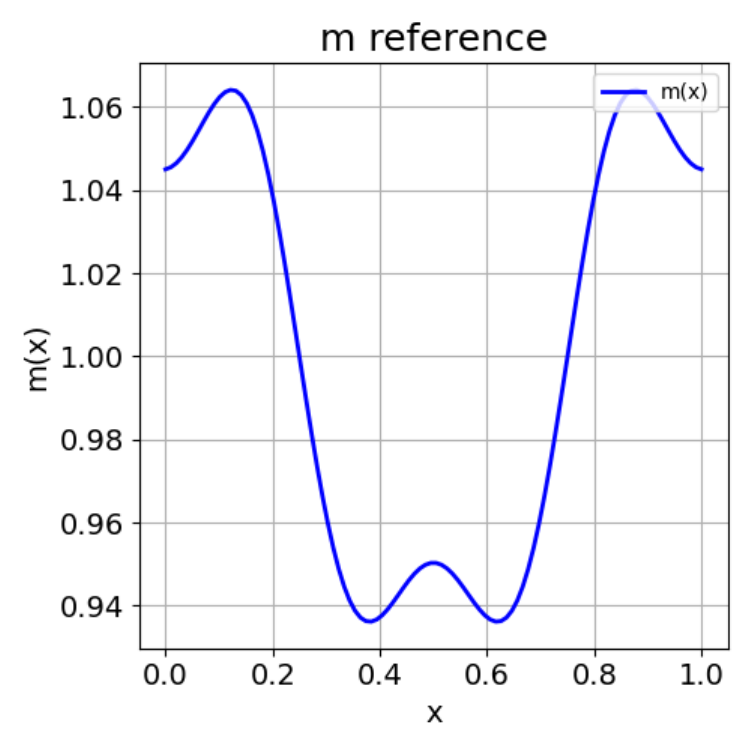}}
        \caption{$m$ reference}
        \label{OnedimensionForwardfig1}
    \end{subfigure}%
    \hspace{1mm}
    \begin{subfigure}[b]{0.23\textwidth}
        \centering
        {\includegraphics[width=\linewidth,page=9]{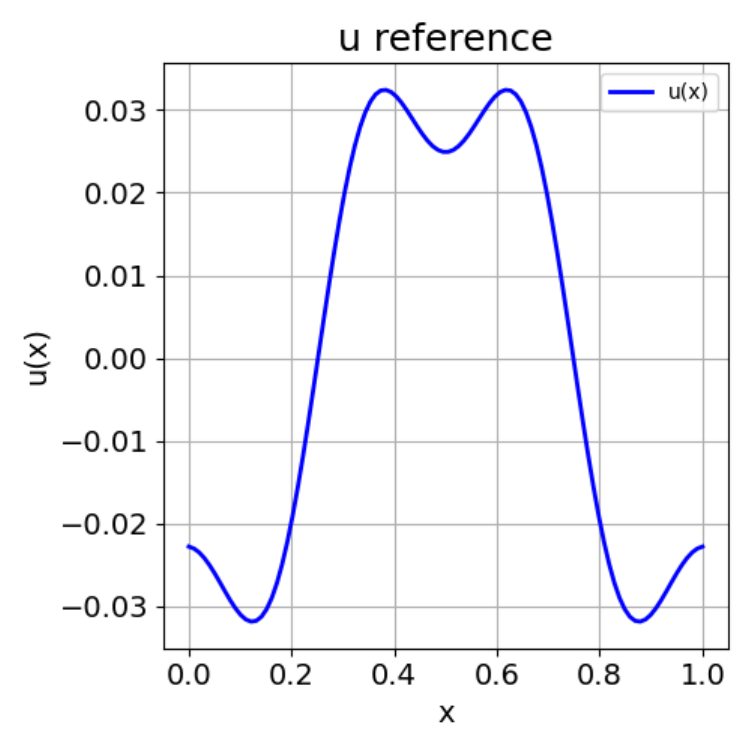}}
        \caption{$u$ reference}
        \label{OnedimensionForwardfig2}
    \end{subfigure}%
    \hspace{1mm}
    \begin{subfigure}[b]{0.23\textwidth}
        \centering
        {\includegraphics[width=\linewidth,page=10]{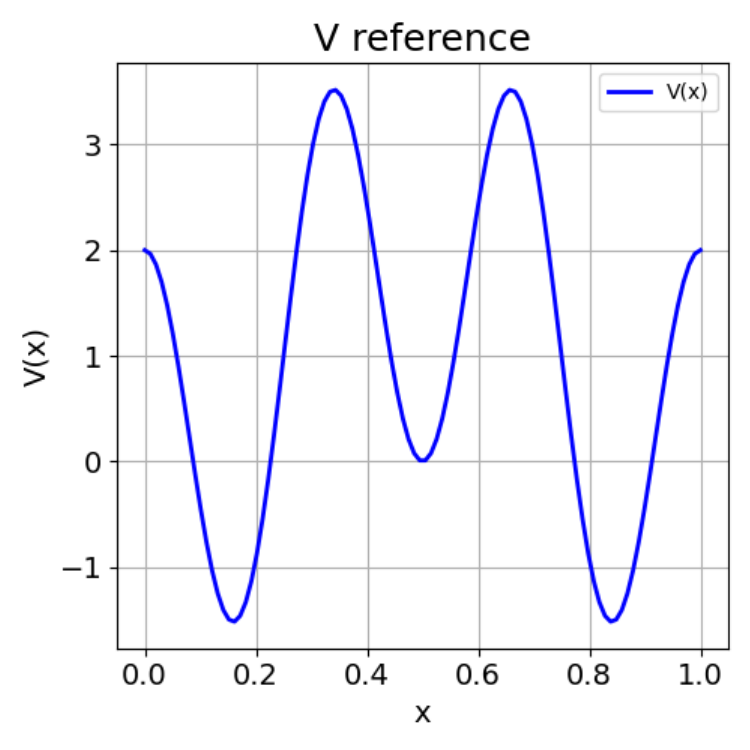}}
        \caption{$V$ reference}
        \label{OnedimensionForwardfig3}
    \end{subfigure}%
    \hspace{1mm}
    \begin{subfigure}[b]{0.23\textwidth}
        \centering
        {\includegraphics[width=\linewidth,page=11]{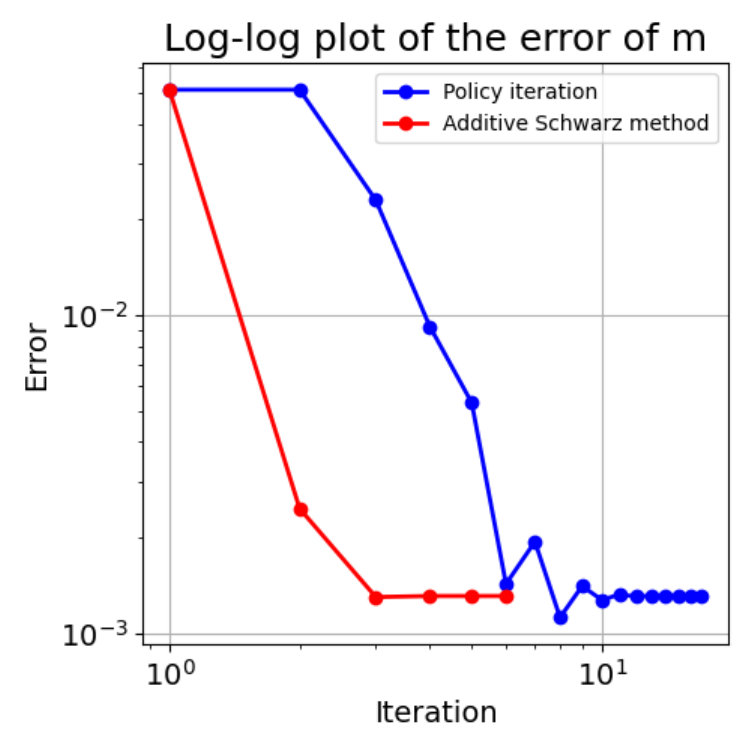}}
        \caption{Log-log error of $m$}
        \label{OnedimensionForwardfig4}
    \end{subfigure}
    
    \begin{subfigure}[b]{0.23\textwidth}
        \centering
        {\includegraphics[width=\linewidth,page=12]{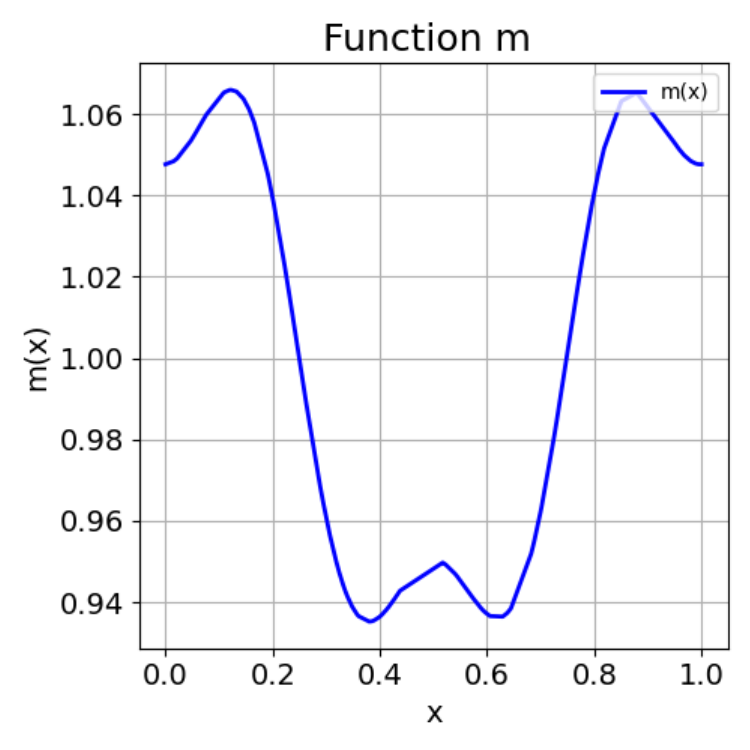}}
        \caption{Recovered $m$ via GPPI}
        \label{OnedimensionForwardfig5}
    \end{subfigure}%
    \hspace{1mm}
    \begin{subfigure}[b]{0.23\textwidth}
        \centering
        {\includegraphics[width=\linewidth,page=13]{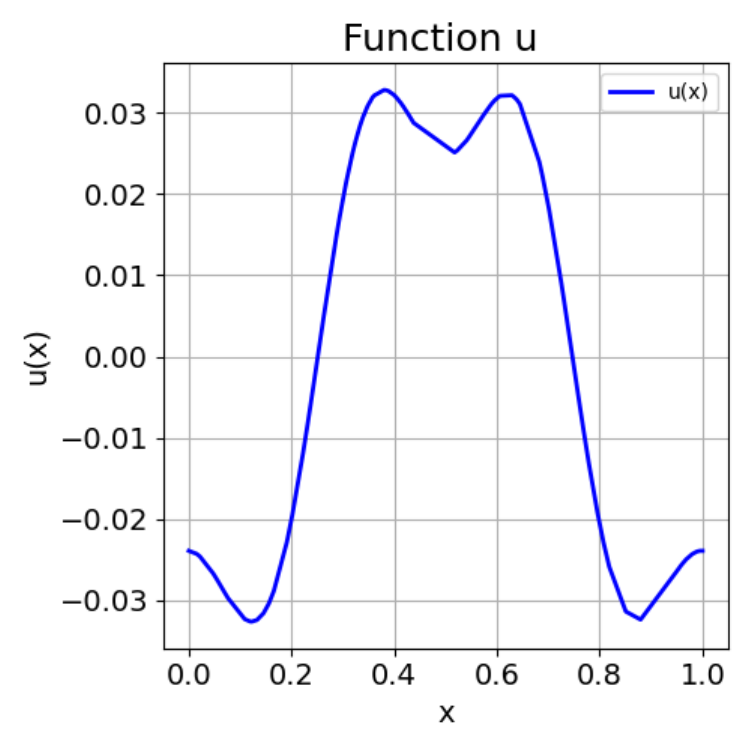}}
        \caption{Recovered $u$ via GPPI}
        \label{OnedimensionForwardfig6}
    \end{subfigure}%
    \hspace{1mm}
    \begin{subfigure}[b]{0.23\textwidth}
        \centering
        {\includegraphics[width=\linewidth,page=14]{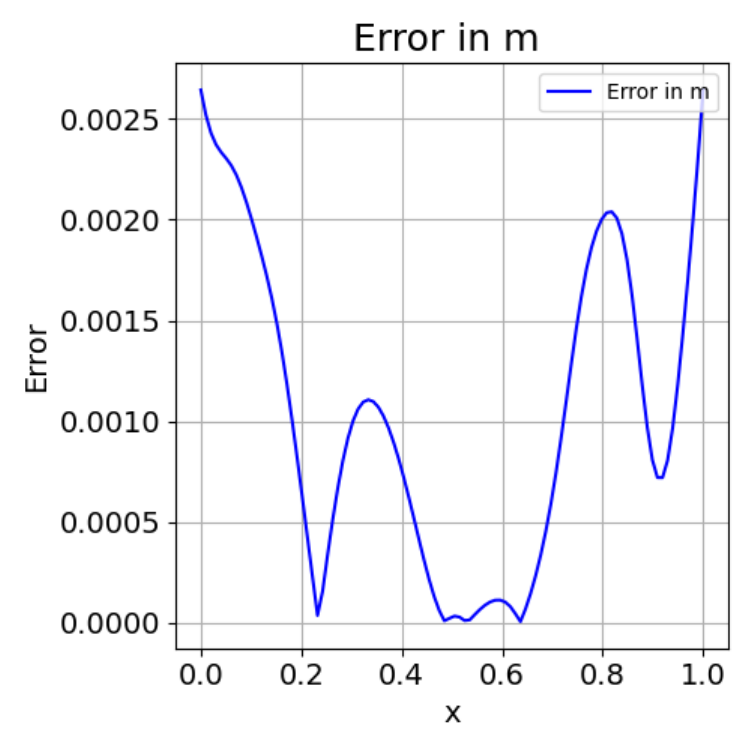}}
        \caption{Error of $m$ via GPPI}
        \label{OnedimensionForwardfig7}
    \end{subfigure}
    \hspace{1mm}
    \begin{subfigure}[b]{0.23\textwidth}
        \centering
        {\includegraphics[width=\linewidth,page=15]{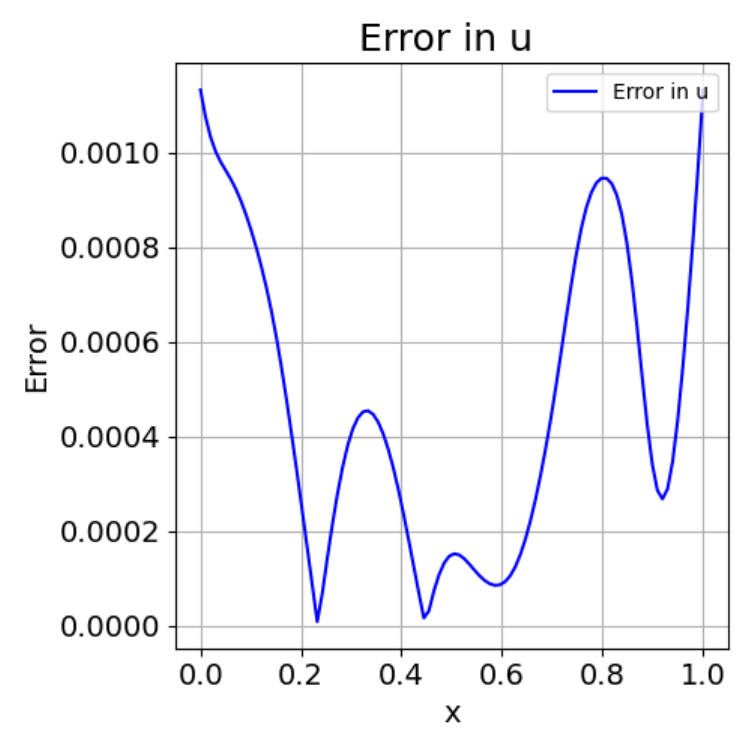}}
        \caption{Error of $u$ via GPPI}
        \label{OnedimensionForwardfig8}
    \end{subfigure}
    
    \begin{subfigure}[b]{0.23\textwidth}
        \centering
        {\includegraphics[width=\linewidth,page=16]{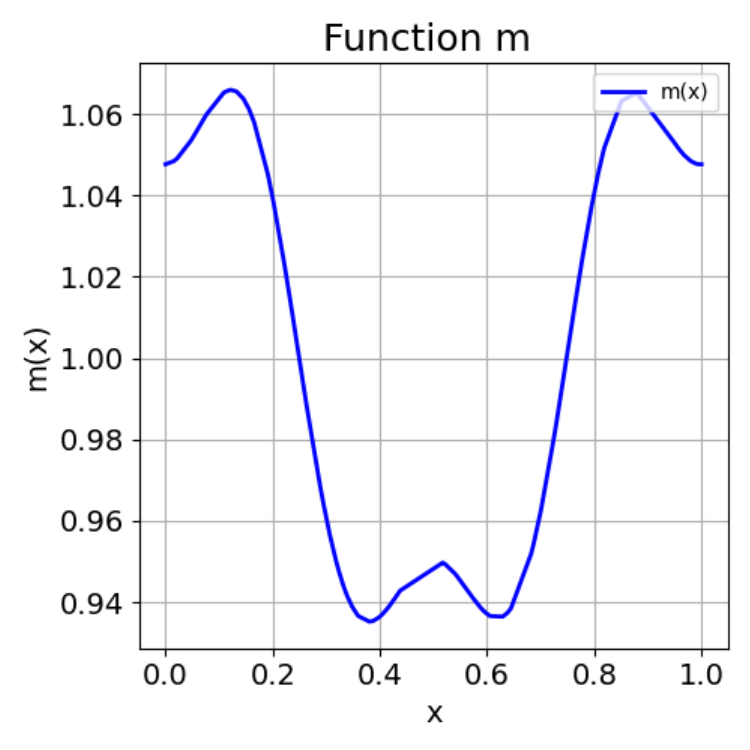}}
        \caption{Solution \(m\) via additive Schwarz
}
        \label{OnedimensionForwardfig9}
    \end{subfigure}%
    \hspace{1mm}
    \begin{subfigure}[b]{0.23\textwidth}
        \centering
        {\includegraphics[width=\linewidth,page=17]{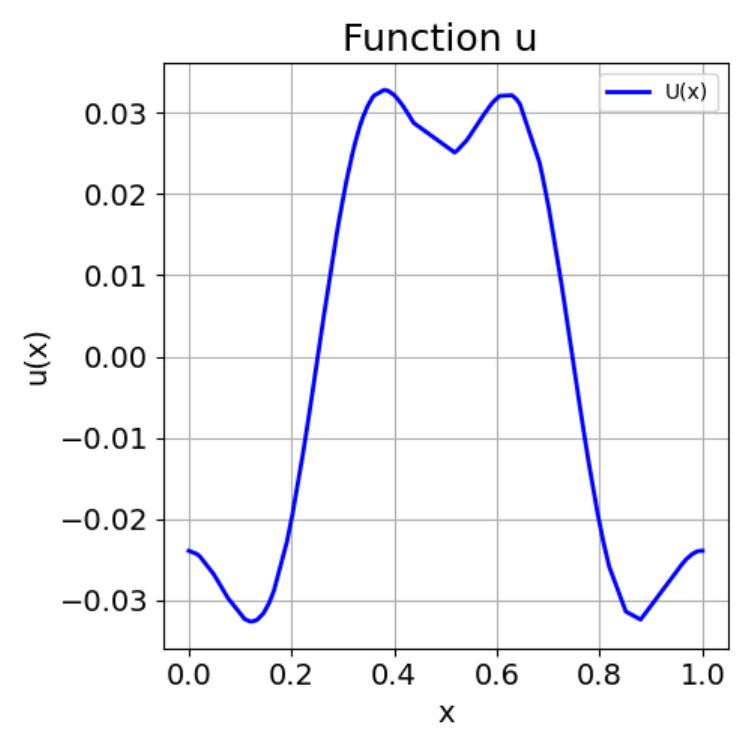}}
        \caption{Solution $u$ via additive Schwarz }
        \label{OnedimensionForwardfig10}
    \end{subfigure}
    \hspace{1mm}
    \begin{subfigure}[b]{0.23\textwidth}
        \centering
        {\includegraphics[width=\linewidth,page=18]{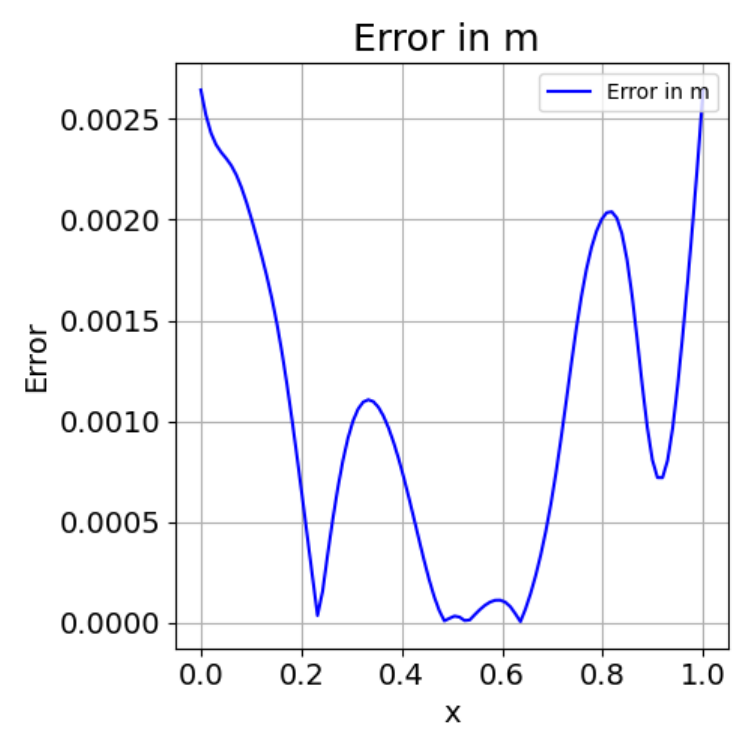}}
        \caption{Error of $m$ via additive Schwarz}
        \label{OnedimensionForwardfig11}
    \end{subfigure}%
    \hspace{1mm}
    \begin{subfigure}[b]{0.23\textwidth}
        \centering
        {\includegraphics[width=\linewidth,page=19]{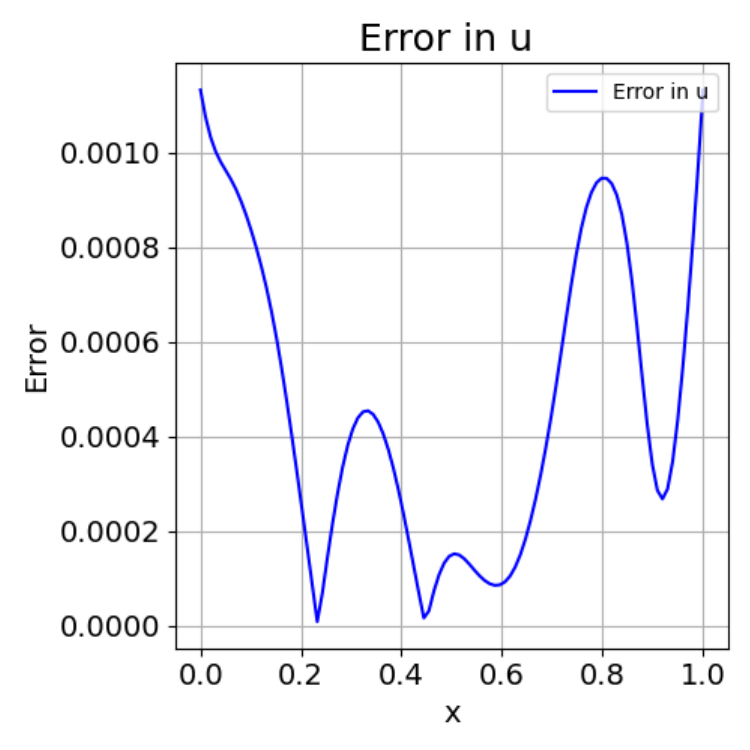}}
        \caption{Error of $u$ via additive Schwarz}
        \label{OnedimensionForwardfig12}
    \end{subfigure}
    
        \caption{Numerical results for the MFG in \eqref{eq:stationaryMFG}. Panels (a), (b), and (c) show the reference solutions for $m$, $u$, and $V$; panel (d) shows the log-log plot of the error of $m$ for GPPI and the additive Schwarz method across iterations; panels (e) and (f) show the solutions $m$ and $u$ via GPPI; panels (g) and (h) show the {absolute pointwise }errors of $m$ and $u$ via GPPI; panels (i) and (j) show the solutions $m$ and $u$ via the additive Schwarz method; and panels (k) and (l) show the {absolute pointwise }errors of $m$ and $u$ via the additive Schwarz method.}
    \label{fig:numerical_results_one_dimension}
\end{figure}

\textbf{Experimental Results.} 
Figure~\ref{fig:numerical_results_one_dimension} shows the discretized \(L^2\) errors \(\mathcal{E}_2(m^{(k)},m^*)\) from \eqref{eq:l2disc}, comparing the \(k\)-th iterate \(m^{(k)}\) with the reference \(m^*\) (Figure~\ref{OnedimensionForwardfig1}, computed by the PI method \cite{cacace2021policy}). It also displays the reference and recovered fields together with absolute pointwise error contours for \(m\) and \(u\). Table~\ref{tableonedimensionLambda} reports the estimates of \(\lambda\) across methods. {As shown in Figure~\ref{OnedimensionForwardfig4}, the GPPI-AS method reaches a comparable error level in fewer iterations than the GPPI method. Specifically, the error curve of GPPI becomes nearly flat after about 16 iterations, whereas GPPI-AS reaches a similar plateau after about 5 iterations. The running time of the GPPI method is \(3.749~\mathrm{s}\), while GPPI-AS reduces the running time to \(0.736~\mathrm{s}\).}

\begin{table}[H]
  \centering
  \caption{Numerical results for the variable \(\lambda\) in the MFG problem \eqref{eq:stationaryMFG}. }
\label{tableonedimensionLambda}
  \begin{tabular}{|c|c|c|c|}
    \hline
    Method& Finite Difference Method  &GPPI & GPPI-AS\\
    \hline
    $\lambda$&2.0033664  & 2.0029773&2.0029776\\
    \hline
  \end{tabular}
\end{table}

\subsection{Inverse Problems of Stationary MFGs}
\label{GPinverseproblem}
In this subsection, we address the inverse problems of the following stationary MFGs using the GPPI framework.  
\begin{equation}
\begin{cases}
-\nu\Delta u + H(x, \nabla u) + \lambda = F(m)+V(x), & \forall x \in \mathbb{T}^d, \\ 
-\nu \Delta m - \operatorname{div}\left( D_p H(x, \nabla u) \, m \right) = 0, & \forall x \in \mathbb{T}^d, \\
\int_{\mathbb{T}^d} u \, \dif x  = 0, \quad \int_{\mathbb{T}^d} m \, \dif x  = 1.
\end{cases}
\label{eqonedimensioninv}
\end{equation}
\mathfix{We begin with a one-dimensional stationary example and then consider two stationary models in two dimensions.}

\subsubsection{One-Dimensional Stationary MFG Inverse Problem}
\label{onedimensioninverse}

In this one-dimensional setting, the exact expressions for \(V\) and \(F\) are given by $
V(x) = \frac{1}{2}(\sin(2\pi x) + \cos(4\pi x))$ and $F(m) = m^3.$
The Hamiltonian is defined as $H(x, p) = \frac{1}{2}|p|^2.$ We set \(d = 1\), \(\nu = 0.3\), and identify \(\mathbb{T}\) with the interval \([0, 1)\). We focus on recovering the distribution \( m \), the value function \( u \), the spatial cost $V$, and the constant \( \lambda \) using the GPPI and \mathfix{GPPI-AS} frameworks proposed in Sections \ref{MFG Forward Problem Stationary} and \ref{Newton Acceleration}.

\textbf{Experimental Setup.}
{In this experiment, 100 sample points for \(m\), \(u\), and \(V\) are randomly generated in the spatial domain \([0,1\mathfix{)}\). For observations of \(m\), $N_m=5$ points are randomly selected from the sample points, and their observed values are obtained by interpolating from a finer reference solution (computed on a uniform grid of $500$ points by the finite difference method) using piecewise linear interpolation. Meanwhile, $N_v=13$ observation points for \(V\) are randomly generated in space, independent of the sample points.}
 {The observation regularization parameters are set to \( \alpha_{m^o} = \mathfix{10^5} \) and \( \alpha_{v^o} = \mathfix{10^5} \). The Gaussian regularization coefficients \( \alpha_m, \alpha_v,\alpha_u,\alpha_{\lambda} \) are all set to 0.5.}
 {Gaussian noise \( \mathcal{N}(0, \gamma^2 I) \) with a standard deviation \( \gamma = 10^{-3} \) is added to the observations.}
 To approximate \(m, u, V \), the GPs with the periodic kernel are employed in this experiment.
{The initial values are $V\equiv 0$,  $u\equiv 0$,  $\boldsymbol{Q}\equiv \boldsymbol{0}$,  $m\equiv 1$,  $\lambda= 0$.}
{The reconstruction error is evaluated on a uniform reference grid by evaluating the recovered GP representation at the grid points.}

\begin{figure}[H]
    \centering
    \begin{subfigure}[b]{0.23\textwidth}
        \centering
        {\includegraphics[width=\linewidth,page=20]{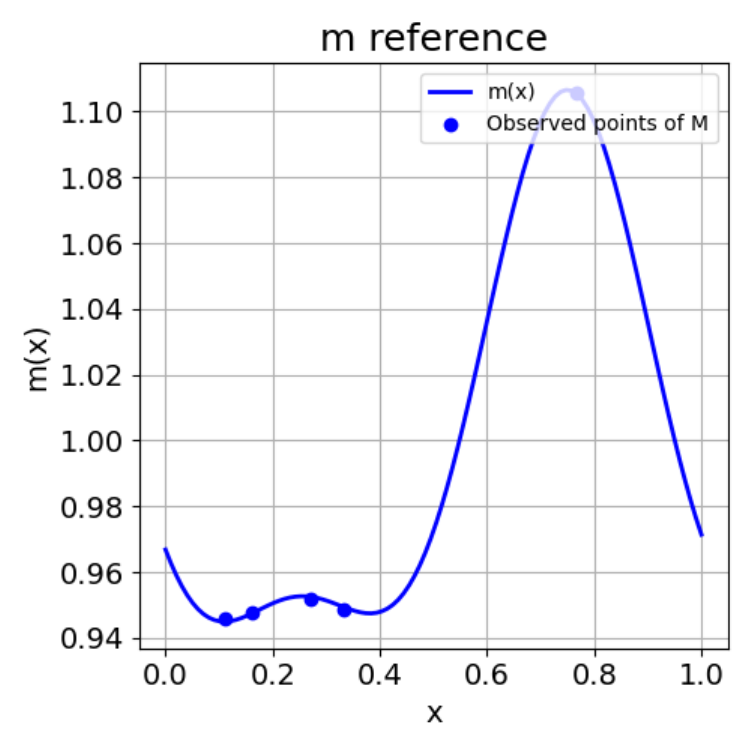}}
        \caption{$m$ reference}
        \label{Onedimensioninversefig1}
    \end{subfigure}%
    \hspace{1mm}
    \begin{subfigure}[b]{0.23\textwidth}
        \centering
        {\includegraphics[width=\linewidth,page=21]{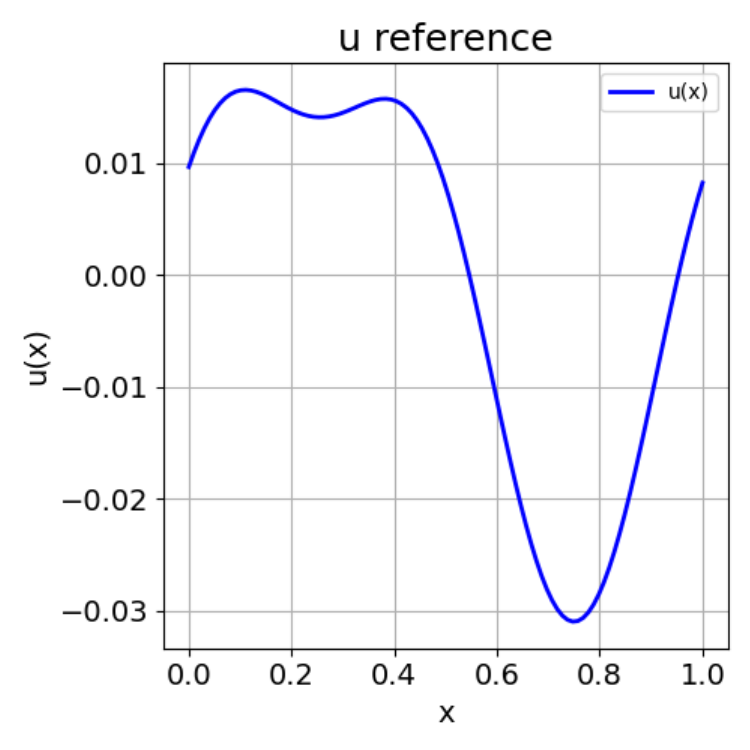}}
        \caption{$u$ reference}
        \label{Onedimensioninversefig2}
    \end{subfigure}%
    \hspace{1mm}
    \begin{subfigure}[b]{0.23\textwidth}
        \centering
        {\includegraphics[width=\linewidth,page=22]{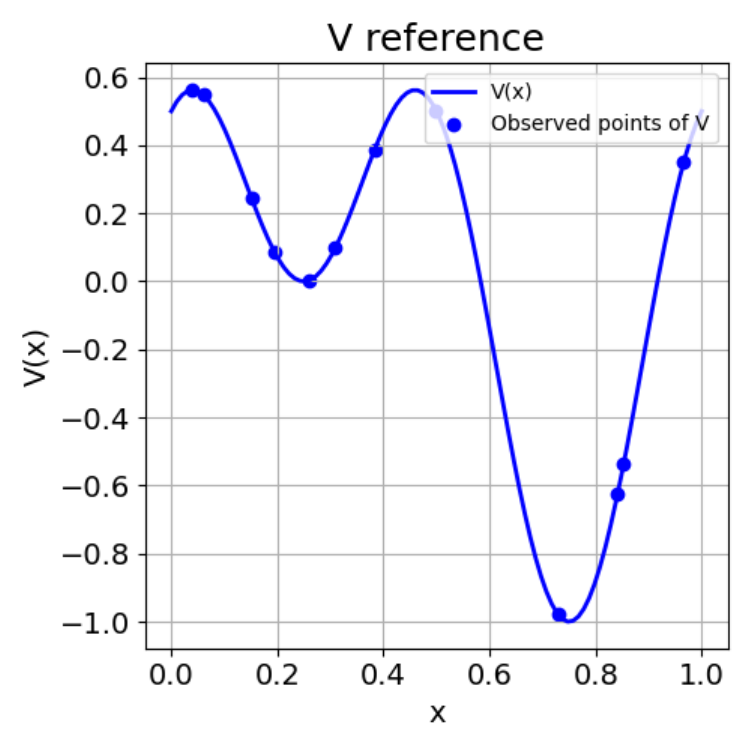}}
        \caption{$V$ reference}
        \label{Onedimensioninversefig3}
    \end{subfigure}%
    \hspace{1mm}
    \begin{subfigure}[b]{0.23\textwidth}
        \centering
        {\includegraphics[width=\linewidth,page=23]{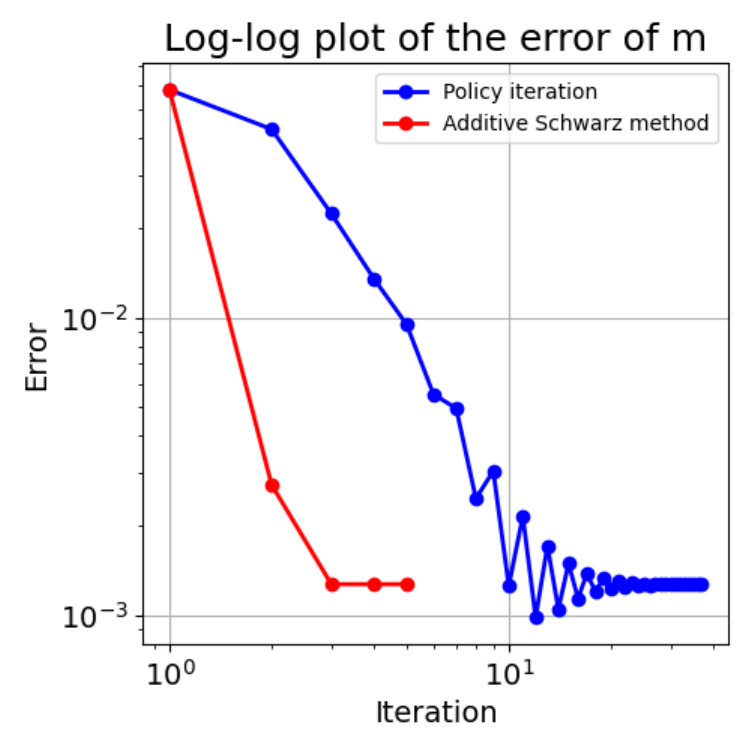}}
        \caption{Log-log error of $m$}
        \label{Onedimensioninversefig4}
    \end{subfigure}%

    \begin{subfigure}[b]{0.23\textwidth}
        \centering
        {\includegraphics[width=\linewidth,page=24]{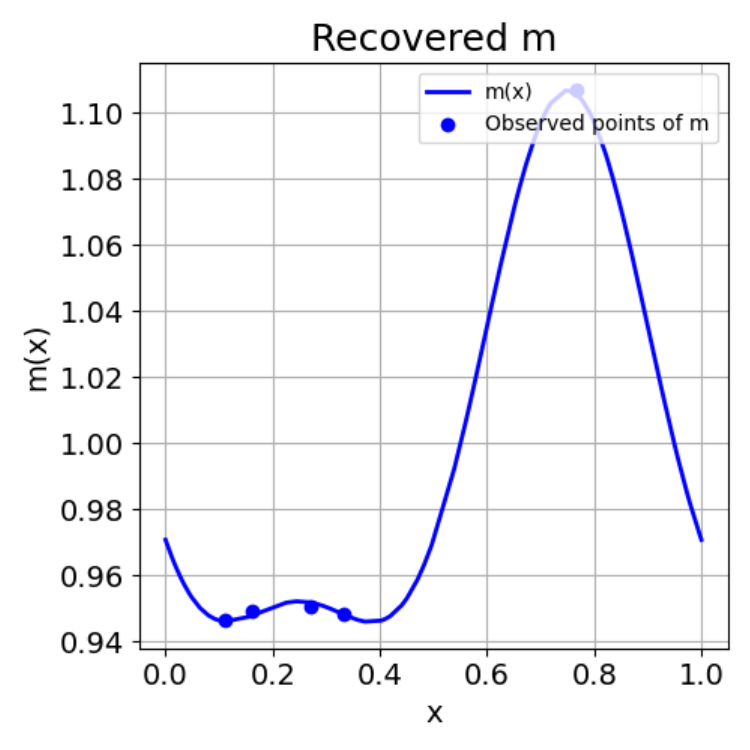}}
        \caption{Recovered $m$ via GPPI}
        \label{Onedimensioninversefig5}
    \end{subfigure}%
    \hspace{1mm}
    \begin{subfigure}[b]{0.23\textwidth}
        \centering
        {\includegraphics[width=\linewidth,page=25]{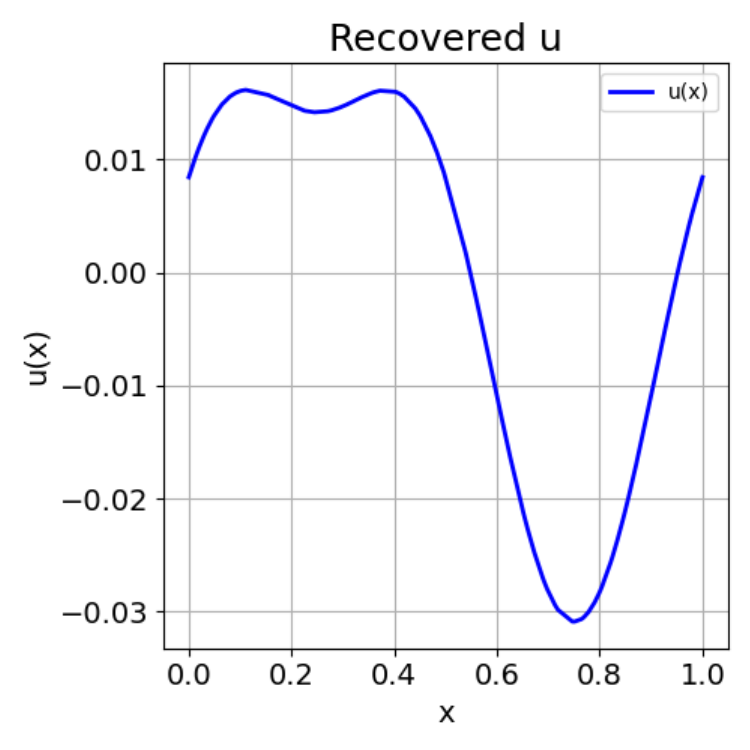}}
        \caption{Recovered $u$ via GPPI}
        \label{Onedimensioninversefig6}
    \end{subfigure}%
    \hspace{1mm}
    \begin{subfigure}[b]{0.23\textwidth}
        \centering
        {\includegraphics[width=\linewidth,page=26]{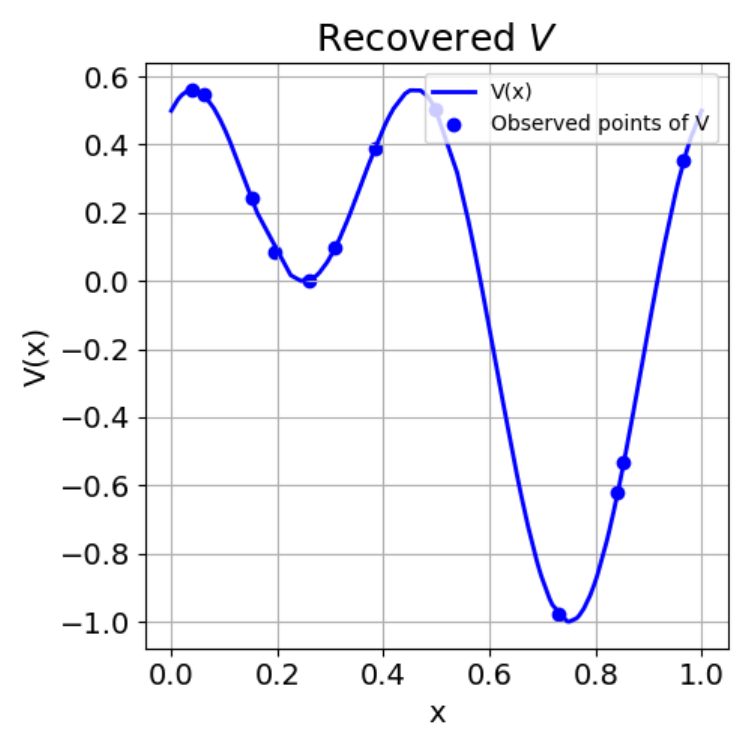}}
        \caption{Recovered $V$ via GPPI}
        \label{Onedimensioninversefig7}
    \end{subfigure}%
    \hspace{1mm}
    \begin{subfigure}[b]{0.23\textwidth}
        \centering
        {\includegraphics[width=\linewidth,page=27]{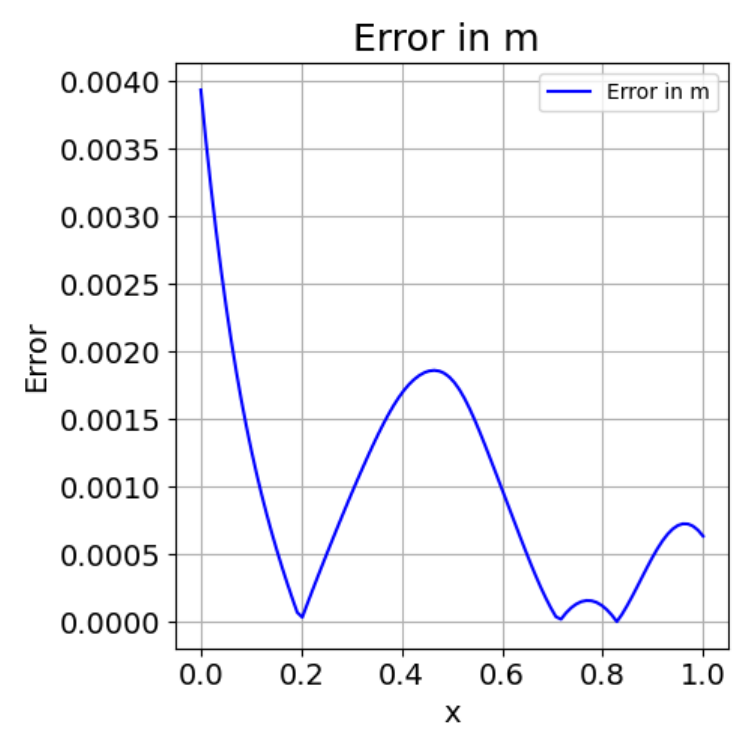}}
        \caption{Error of $m$ via GPPI}
        \label{Onedimensioninversefig8}
    \end{subfigure}%

    \begin{subfigure}[b]{0.23\textwidth}
        \centering
        {\includegraphics[width=\linewidth,page=28]{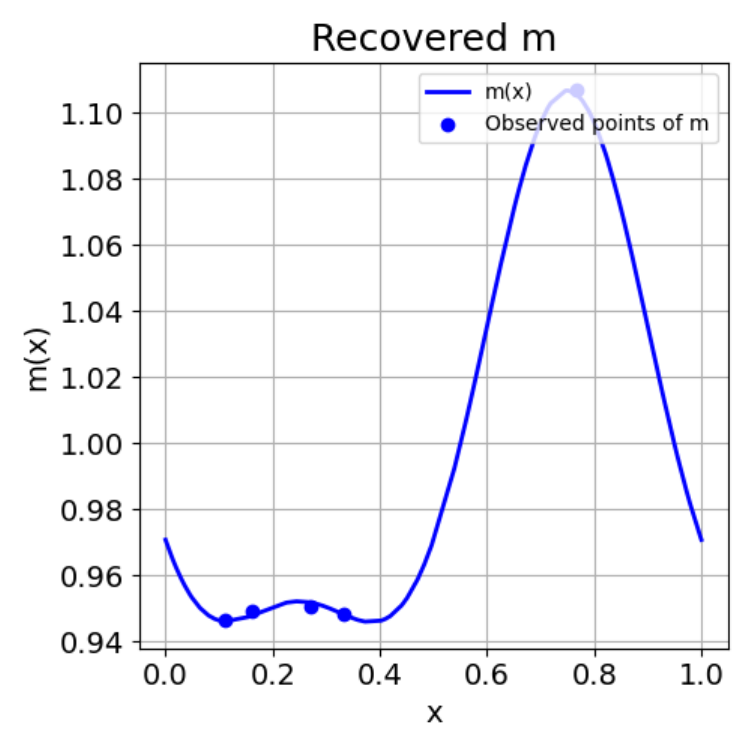}}
        \caption{Recovered $m$ via additive Schwarz}
        \label{Onedimensioninversefig9}
    \end{subfigure}%
    \hspace{1mm}
    \begin{subfigure}[b]{0.23\textwidth}
        \centering
        {\includegraphics[width=\linewidth,page=29]{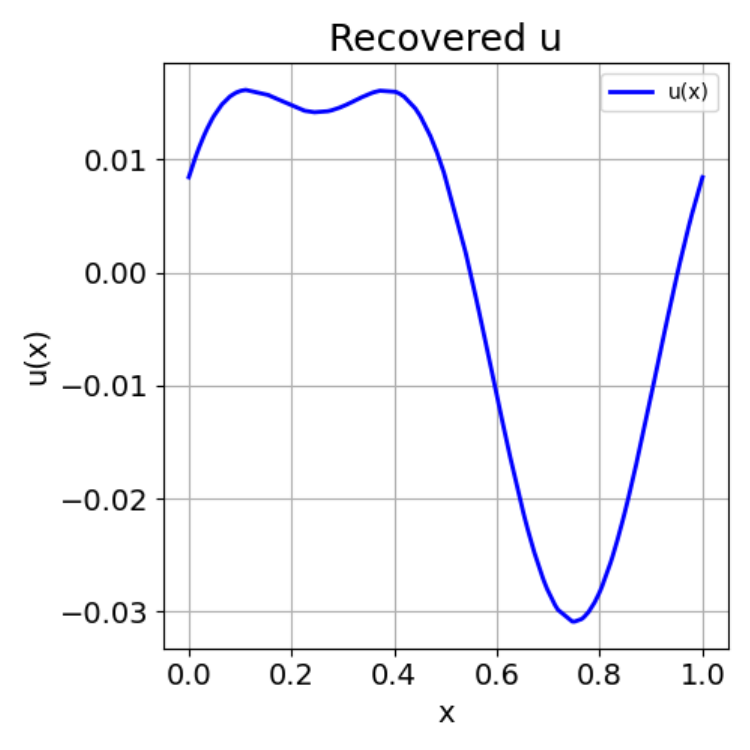}}
        \caption{Recovered $u$ via additive Schwarz}
        \label{Onedimensioninversefig10}
    \end{subfigure}%
    \hspace{1mm}
    \begin{subfigure}[b]{0.23\textwidth}
        \centering
        {\includegraphics[width=\linewidth,page=30]{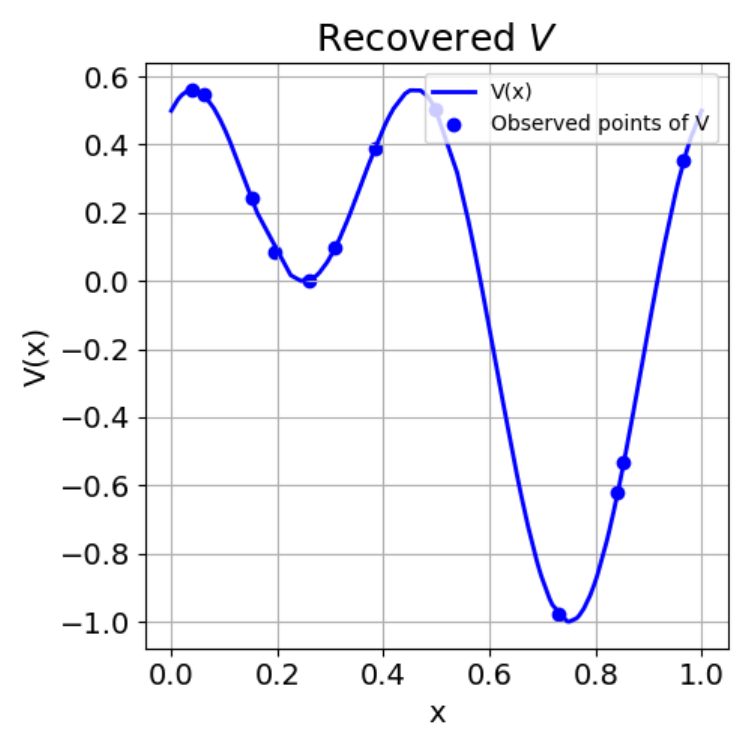}}
        \caption{Recovered $V$ via additive Schwarz}
        \label{Onedimensioninversefig11}
    \end{subfigure}%
    \hspace{1mm}
    \begin{subfigure}[b]{0.23\textwidth}
        \centering
       {\includegraphics[width=\linewidth,page=31]{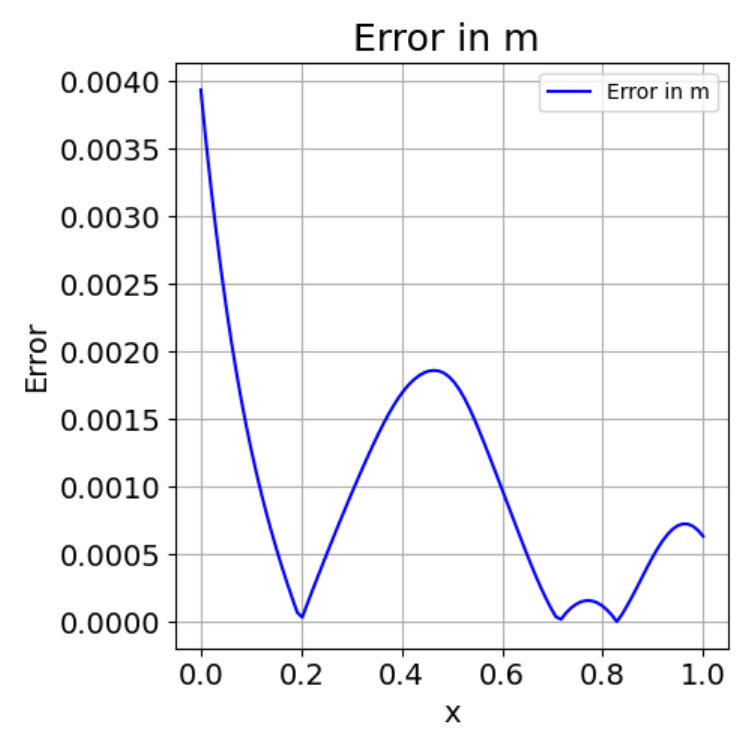}}
        \caption{Error of $m$ via additive Schwarz}
        \label{Onedimensioninversefig12}
    \end{subfigure}%

    \begin{subfigure}[b]{0.23\textwidth}
        \centering
       {\includegraphics[width=\linewidth,page=32]{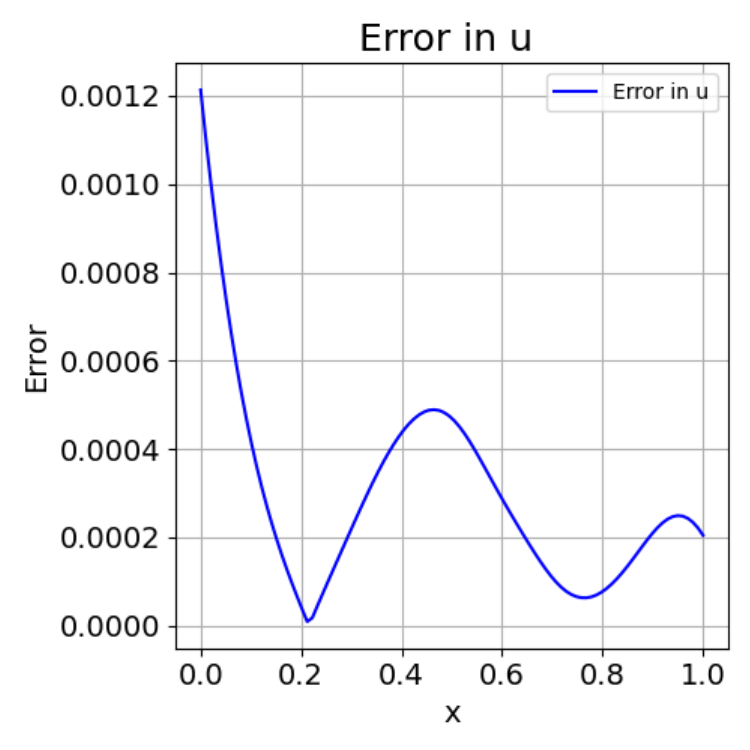}}
        \caption{Error of $u$ via GPPI}
        \label{Onedimensioninversefig13}
    \end{subfigure}%
    \hspace{1mm}
    \raisebox{-3.5mm}{%
\begin{subfigure}[b]{0.23\textwidth}
    \centering
    {\includegraphics[width=\linewidth,page=33]{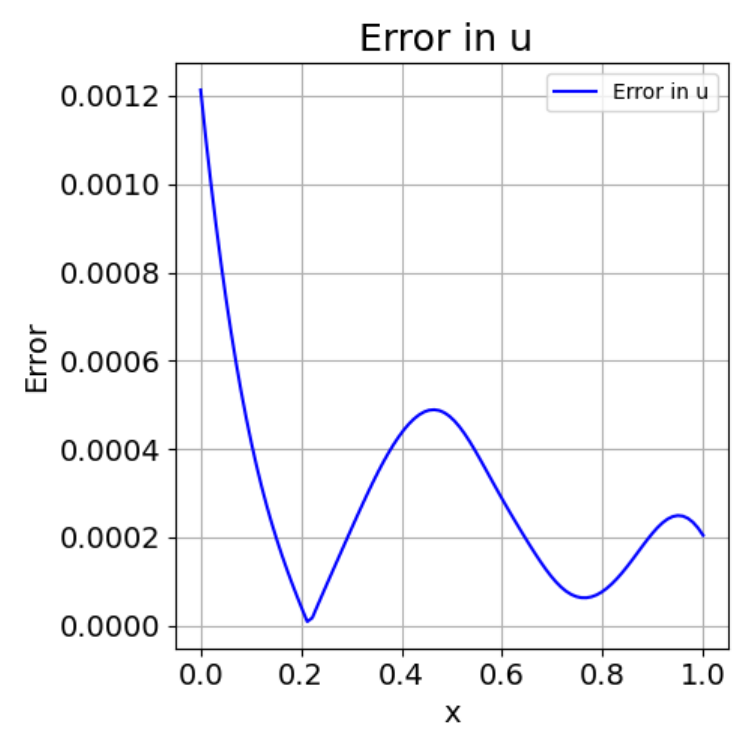}}
    \caption{Error of $u$ via additive Schwarz}
    \label{Onedimensioninversefig14}
\end{subfigure}}%
    \hspace{1mm}
    \begin{subfigure}[b]{0.23\textwidth}
        \centering
       {\includegraphics[width=\linewidth,page=34]{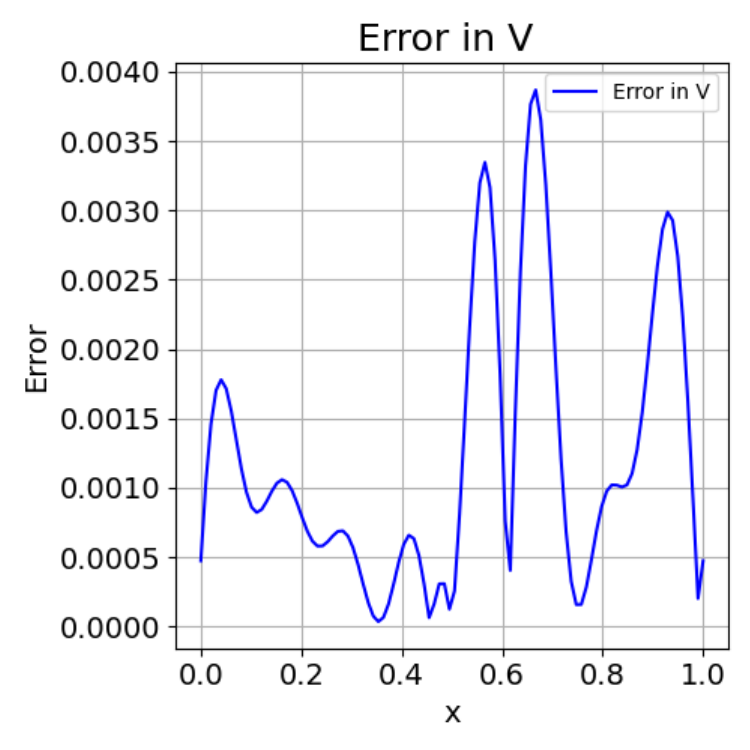}}
        \caption{Error of $V$ via GPPI}
        \label{Onedimensioninversefig15}
    \end{subfigure}%    
    \hspace{1mm}
    \raisebox{-3.5mm}{%
\begin{subfigure}[b]{0.23\textwidth}
    \centering
    {\includegraphics[width=\linewidth,page=35]{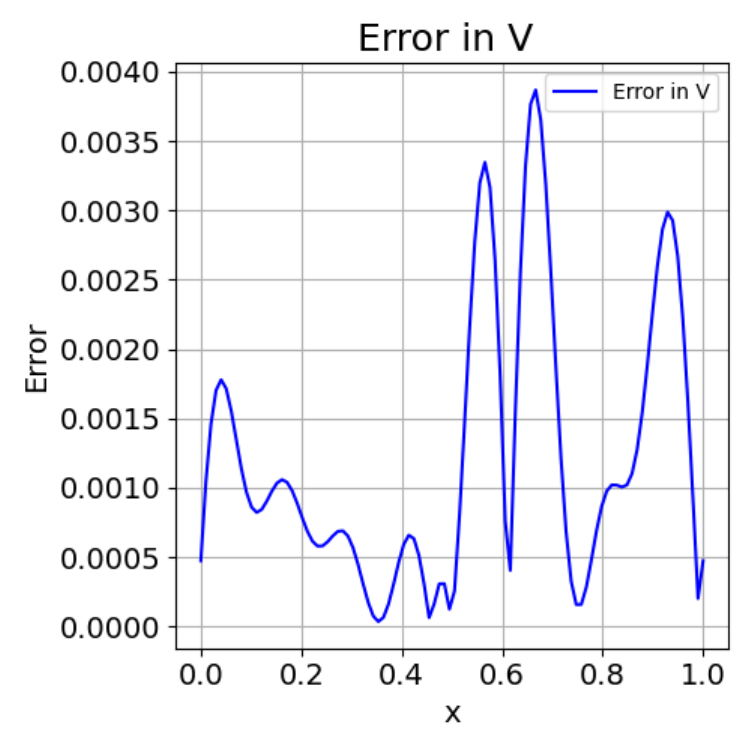}}
    \caption{Error of $V$ via additive Schwarz}
    \label{Onedimensioninversefig16}
\end{subfigure}}%   
    \caption{Numerical results for the one-dimensional inverse problem of MFG in \Cref{onedimensioninverse}. Panels (a)--(c) show the reference solutions for $m$, $u$, and $V$, and panel (d) shows the log-log plot of the error of $m$ for GPPI and the additive Schwarz method across iterations. Panels (e)--(g), (h), (m), (o) show the recovered $m$, $u$, $V$ and their {absolute pointwise }errors via GPPI; panels (i)--(k), (l), (n), (p) show the same via the additive Schwarz method.}
    \label{onedimensioninverse_result}
\end{figure}

\textbf{Experimental Results.}
Figure~\ref{onedimensioninverse_result} reports the discretized $L^2$ errors $\mathcal{E}_2(m^{(k)},m^*)$ from \eqref{eq:l2disc}, comparing each iterate $m^{(k)}$ with the reference $m^*$ in Figure~\ref{Onedimensioninversefig1}. The panels also show the ground truth and reconstructions for $m$, $u$, and $V$, together with {absolute} pointwise error curves. Table~\ref{tableonedimensioninv} summarizes the estimates of $\lambda$ across methods. 
{Finally, Figure~\ref{Onedimensioninversefig4} shows that \mathfix{GPPI-AS} reaches a stable error level faster than GPPI, requiring 4 iterations and \(0.898~\mathrm{s}\), compared with 36 iterations and \(5.236~\mathrm{s}\) for GPPI. Although the outer iteration count is reduced by roughly a factor of nine, the wall-clock time decreases by about a factor of six. This difference is mainly due to the additional per-iteration cost of the Newton step, including Jacobian assembly and the solution of the associated linear system. A promising direction for future work is to further exploit parallelism in the GPPI-AS updates of \(m\), \(u\), and \(\boldsymbol{Q}\) in parallel computing environments, and to reduce the Jacobian overhead using analytic derivatives that exploit the PDE and kernel operator structure.}

{
Tables~\ref{1Dinverseobserve} and \ref{1Dinversenoise} report the
relative \(L^1\), \(L^2\), and \(L^\infty\) reconstruction errors of \(m\)
for the one-dimensional inverse problem in \eqref{eqonedimensioninv}.
\mathfix{For the observation-number study in Table~\ref{1Dinverseobserve}, we use a separate setting in which $N_m$ and $N_v$ are kept equal and varied together.}
Table~\ref{1Dinverseobserve} shows that, for a fixed noise level as in the experimental setup, the
errors decrease as the number of observation points for \(m\) and \(V\)
increases jointly. Table~\ref{1Dinversenoise} shows that, for a fixed observation
set as in the experimental setup, adding relative Gaussian noise according to \eqref{eq:relnoise}
leads to a moderate increase in the errors as \(\tilde{\gamma}\) grows.
A similar qualitative behavior is observed
for the reconstruction errors of other variables. The subsequent numerical examples follow the same procedure for the \mathfix{observation-number and noise-level studies}.}

\begin{table}[H]
  \centering
  \caption{Numerical results for \(\lambda\) in the one-dimensional MFG problem in \Cref{onedimensioninverse}.}
  \label{tableonedimensioninv}
  \begin{tabular}{|c|c|c|c|}
    \hline
    Method& Finite Difference Method  &GPPI & GPPI-AS\\
    \hline
    $\lambda$& 1.0072959 & 1.0023512&1.0023513 \\
    \hline
  \end{tabular}
\end{table}
\begin{table}[H]
\centering
\caption{Relative recovery error of $m$ under different numbers of observations for the one-dimensional MFG problem in \Cref{onedimensioninverse}. }
\label{1Dinverseobserve}
\begin{tabular}{lccccc}
\toprule
Number of observations & 5 & 7 & 9 & 11 & 14 \\
\midrule
Relative $L^1$ error       & 2.477e-02 & 1.597e-02 & 8.356e-03 & 1.101e-03 & 1.074e-03 \\
Relative $L^2$ error       & 2.906e-02 & 1.835e-02 & 9.822e-03 & 1.430e-03 & 1.370e-03 \\
Relative $L^\infty$ error  & 4.819e-02 & 3.315e-02 & 1.803e-02 & 3.590e-03 & 3.421e-03 \\
\bottomrule
\end{tabular}
\end{table}
\begin{table}[H]
\centering
\caption{Relative recovery error of $m$ under different noise levels $\tilde{\gamma}$ for the one-dimensional MFG problem in \Cref{onedimensioninverse}.}
\label{1Dinversenoise}
\begin{tabular}{lccccc}
\toprule
Noise level $\tilde{\gamma}$ & $0e\!+\!00$ & $1e\!-\!03$ & $5e\!-\!03$ & $1e\!-\!02$ & $5e\!-\!02$ \\
\midrule
Relative $L^1$ error      & 9.662e-04 & 1.030e-03 & 1.424e-03 & 2.065e-03 & 7.962e-03 \\
Relative $L^2$ error      & 1.206e-03 & 1.245e-03 & 1.578e-03 & 2.220e-03 & 8.649e-03 \\
Relative $L^\infty$ error & 3.324e-03 & 3.360e-03 & 3.504e-03 & 3.684e-03 & 1.159e-02 \\
\bottomrule
\end{tabular}
\end{table}

\subsubsection{Two-Dimensional Stationary MFG Inverse Problem with Quadratic Coupling}
\label{Twodimensioninverseold}

In this case, we consider the MFG system in \eqref{eqonedimensioninv} when \(d=2\).
Here, the Hamiltonian is defined as \mathfix{\( H(x,y,p) = \frac{1}{2}|p|^2 \)}. In this experiment, we identify \(\mathbb{T}^2\) by \([-0.5, 0.5) \times [-0.5, 0.5)\), and the true spatial cost function is \( V(x, y) = -1.4(\sin(2 \pi x) + \cos(4 \pi y) + \sin(4 \pi y)) \). The function \( F(m) \) is defined as \( m^2 \), and the viscosity coefficient \( \nu \) is set to 0.3.

We concentrate on reconstructing the distribution \( m \), the value function \( u \), the spatial cost function \( V \), and the constant \( \lambda \) using the GPPI and \mathfix{GPPI-AS} frameworks outlined in Sections \ref{MFG Forward Problem Stationary} and \ref{Newton Acceleration}.

{\textbf{Experimental Setup.} A total of 361 sample points for \(m\), \(u\), and \(V\) are randomly generated in the spatial domain \(\mathbb{T}^2\). From these, 40 observation points \( \boldsymbol{m}^o \) are randomly selected, and their observed values are obtained by interpolating from a finer reference solution (computed on a uniform $30 \times 30$ grid) using cubic interpolation. The 90 observation points \( \boldsymbol{V}^o \) are randomly distributed in space, independent of the sample points. We set the regularization parameters as \( \alpha_{m^o} = \mathfix{10^5} \) and \( \alpha_{v^o} = \mathfix{10^5} \). The Gaussian regularization coefficients are set to \( \alpha_v = 0.5 \), with \( \alpha_{u} = 0.5, \alpha_{m} = 0.5 \), and \( \alpha_{\lambda} = 0.5\). Gaussian noise with a standard deviation \( \gamma = 10^{-3} \) is added to these observations, modeled as \( \mathcal{N}(0, \gamma^2 I) \).}

{The initial values are \( V\equiv 0 \), \( u \equiv 0 \), \( \boldsymbol{Q} \equiv \boldsymbol{0} \), \( m \equiv 1 \), and \( \lambda = 0 \). We employ periodic kernels for the GP priors. To evaluate the reconstruction error, the GP solution is interpolated back onto a uniform reference grid via the GP posterior mean and compared with the reference solution.}

\begin{figure}[H]
    \centering
    \begin{subfigure}[b]{0.23\textwidth}
        \centering
        {\includegraphics[width=\linewidth,page=36]{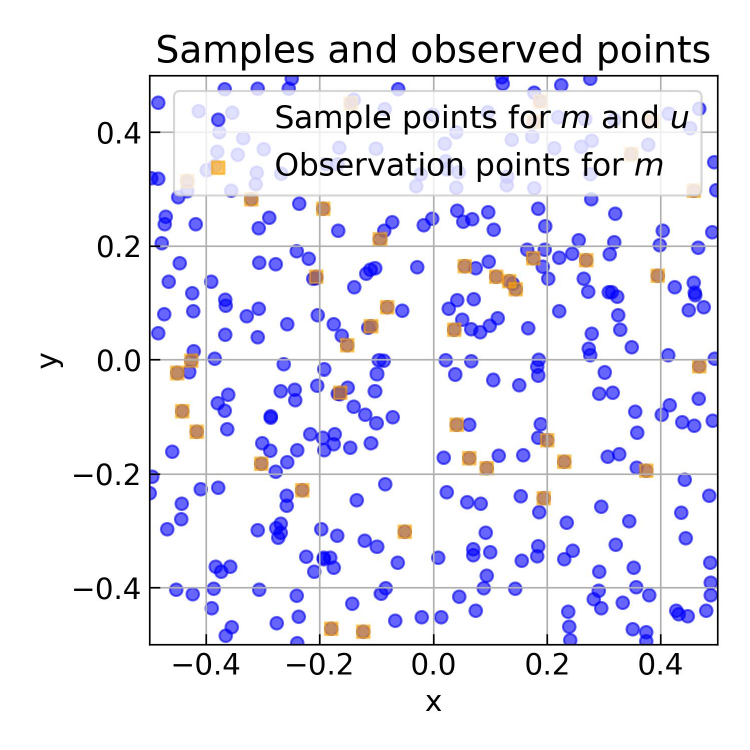}}
        \caption{Samples for $m$ and $u$ \& observations for $m$}
        \label{Multidimensioninversefig4}
    \end{subfigure}
    \hspace{5mm} % Adjust the spacing as needed between subfigures
    \begin{subfigure}[b]{0.23\textwidth}
        \centering
        {\includegraphics[width=\linewidth,page=37]{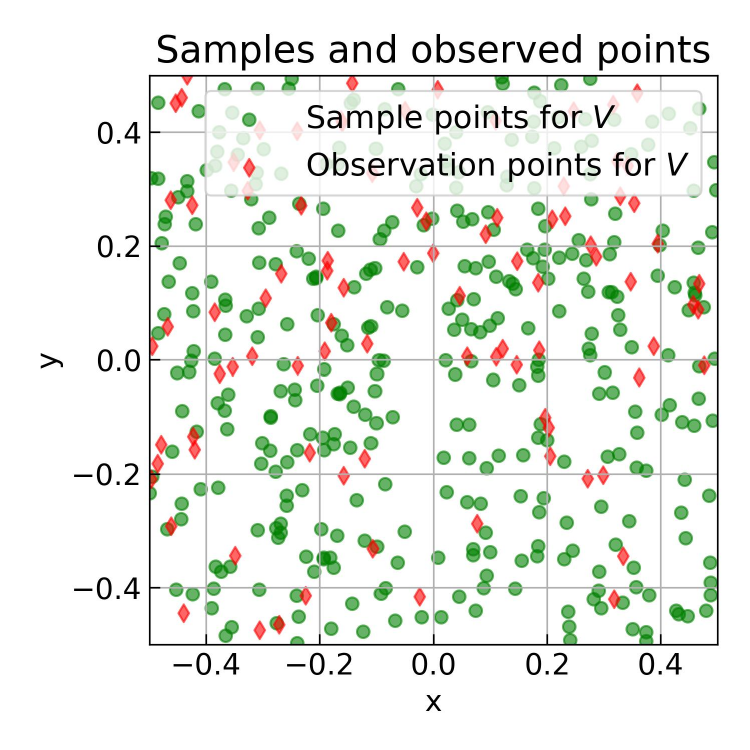}}
        \caption{Samples \& observation points for $V$}       \label{Multidimensioninversefig0}
    \end{subfigure}
    \caption{The inverse problem of the two-dimensional stationary MFG in  \Cref{Twodimensioninverseold}: samples for $m, u,$ and $V$ \& observations for $m$ and $V$.}
    \label{fig:samples_observationsmulti}
\end{figure}

\begin{figure}[H]
    \centering
    \begin{subfigure}[b]{0.23\textwidth}
        \centering
        {\includegraphics[width=\linewidth,page=38]{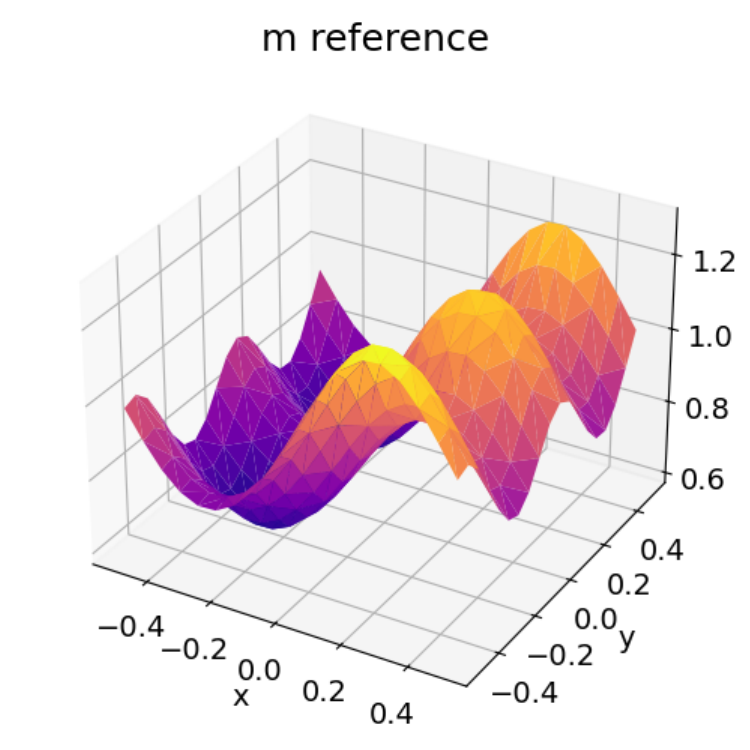}}
        \caption{$m$ reference}
        \label{Multidimensioninversefig1}
    \end{subfigure}%
    \hspace{1mm}
    \begin{subfigure}[b]{0.23\textwidth}
        \centering
        {\includegraphics[width=\linewidth,page=39]{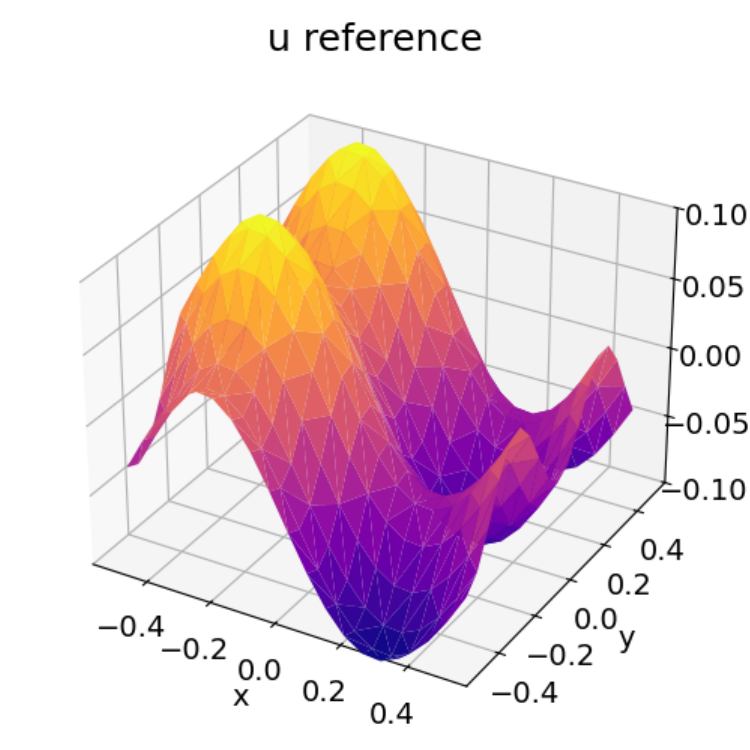}}
        \caption{$u$ reference}
        \label{Multidimensioninversefig2}
    \end{subfigure}%
    \hspace{1mm}
    \begin{subfigure}[b]{0.23\textwidth}
        \centering
        {\includegraphics[width=\linewidth,page=40]{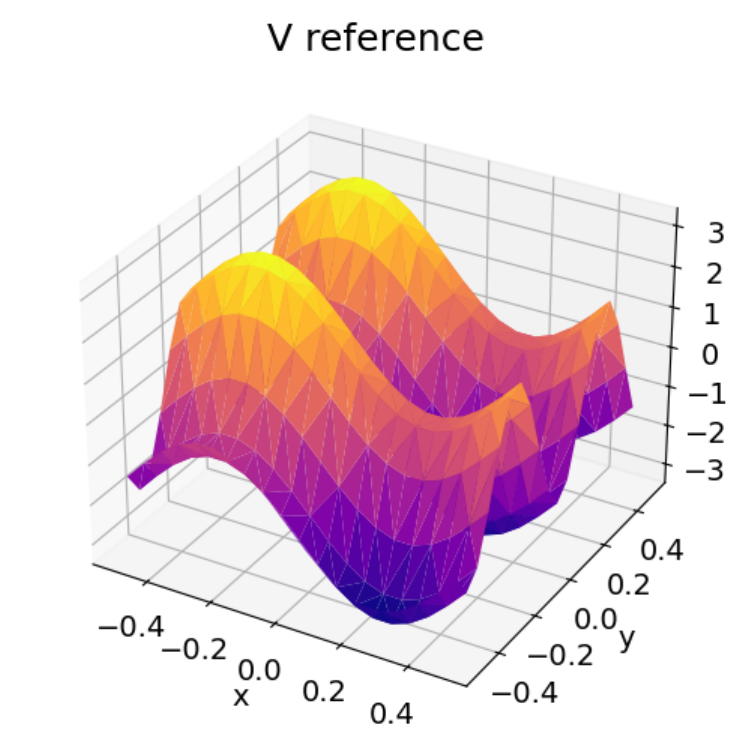}}
        \caption{$V$ reference}
        \label{Multidimensioninversefig3}
    \end{subfigure}%
    \hspace{1mm}
    \begin{subfigure}[b]{0.23\textwidth}
        \centering
        {\includegraphics[width=\linewidth,page=41]{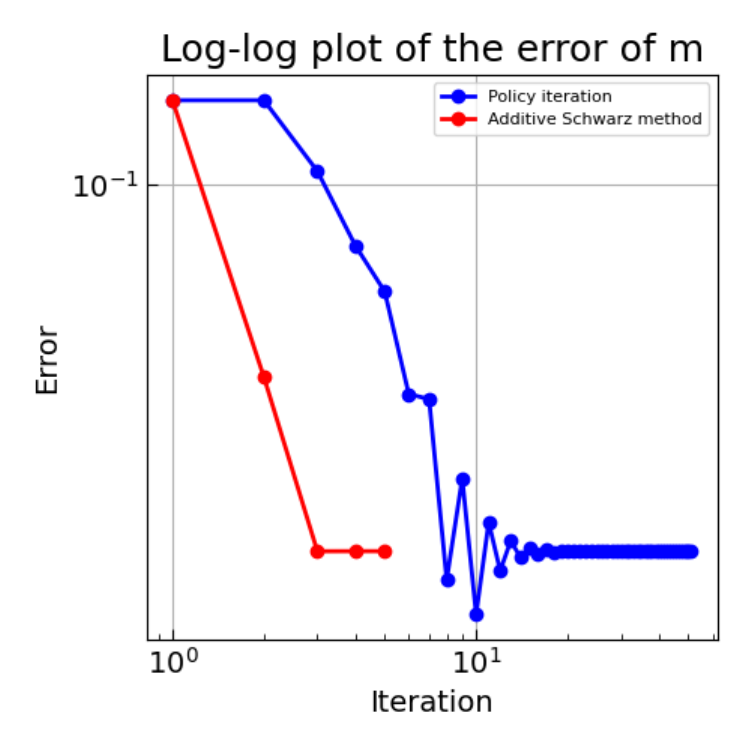}}
        \caption{Log-log error of $m$}
        \label{Multidimensioninversefig5}
    \end{subfigure}

    \begin{subfigure}[b]{0.23\textwidth}
        \centering
        {\includegraphics[width=\linewidth,page=42]{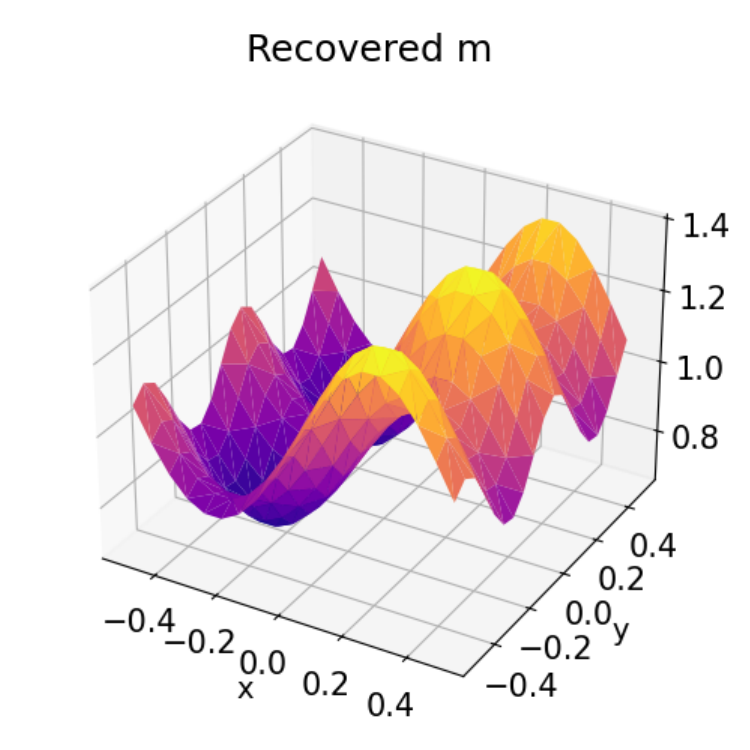}}
        \caption{Recovered $m$ via GPPI}
        \label{Multidimensioninversefig6}
    \end{subfigure}%
    \hspace{1mm}
    \begin{subfigure}[b]{0.23\textwidth}
        \centering
        {\includegraphics[width=\linewidth,page=43]{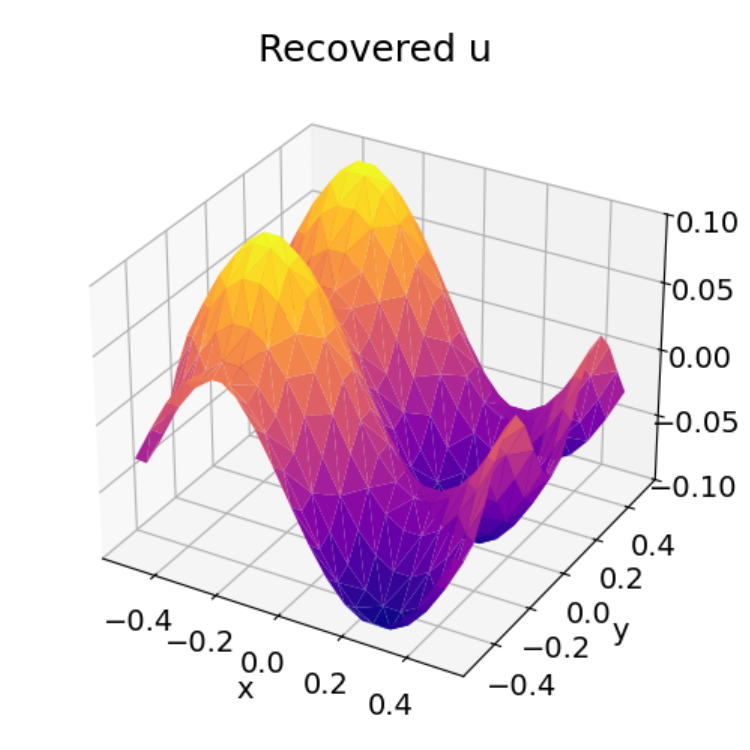}}
        \caption{Recovered $u$ via GPPI}
        \label{Multidimensioninversefig7}
    \end{subfigure}%
    \hspace{1mm}
    \begin{subfigure}[b]{0.23\textwidth}
        \centering
        {\includegraphics[width=\linewidth,page=44]{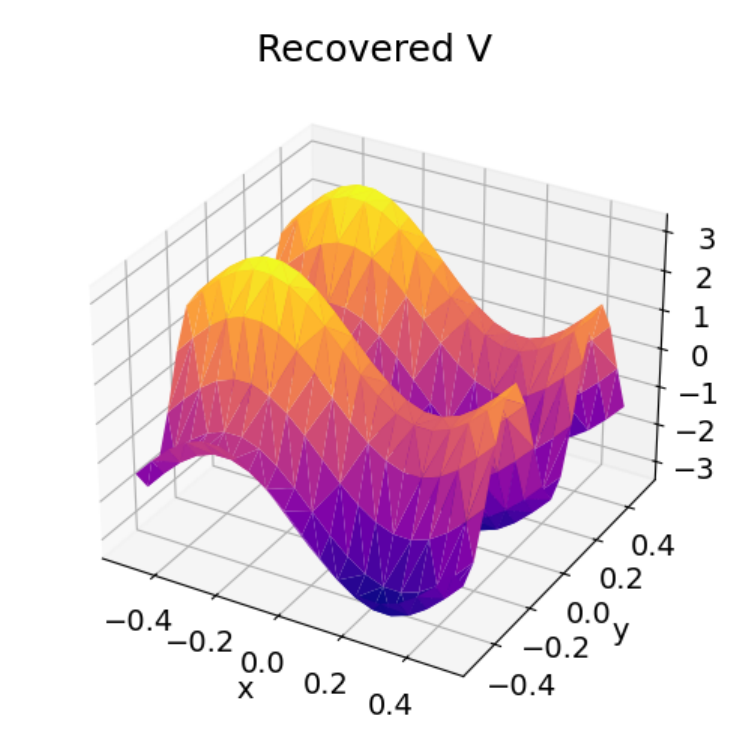}}
        \caption{Recovered $V$ via GPPI}
        \label{Multidimensioninversefig8}
    \end{subfigure}%
    \hspace{1mm}
    \raisebox{-3.8mm}{
    \begin{subfigure}[b]{0.23\textwidth}
        \centering
        {\includegraphics[width=\linewidth,page=45]{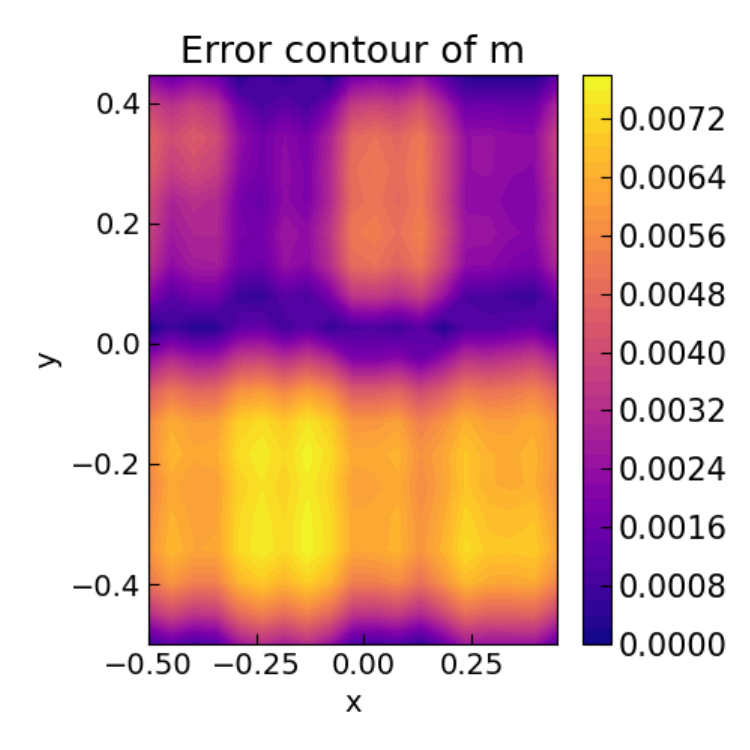}}
        \caption{Error contour of $m$ via GPPI}
        \label{Multidimensioninversefig9}
    \end{subfigure}}

    \begin{subfigure}[b]{0.23\textwidth}
        \centering
        {\includegraphics[width=\linewidth,page=46]{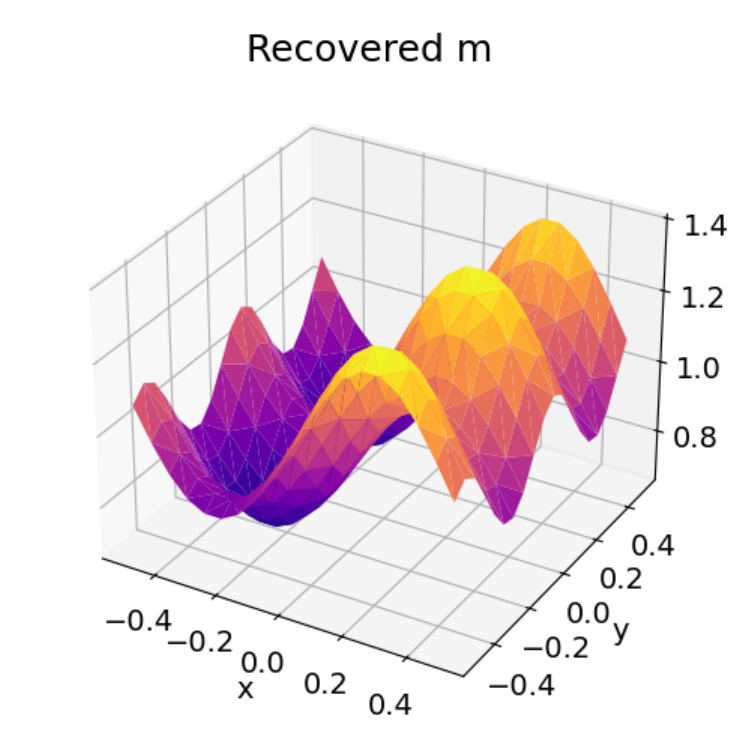}}
        \caption{Recovered $m$ via additive Schwarz method}
        \label{Multidimensioninversefig10}
    \end{subfigure}%
    \hspace{1mm}
    \begin{subfigure}[b]{0.23\textwidth}
        \centering
        {\includegraphics[width=\linewidth,page=47]{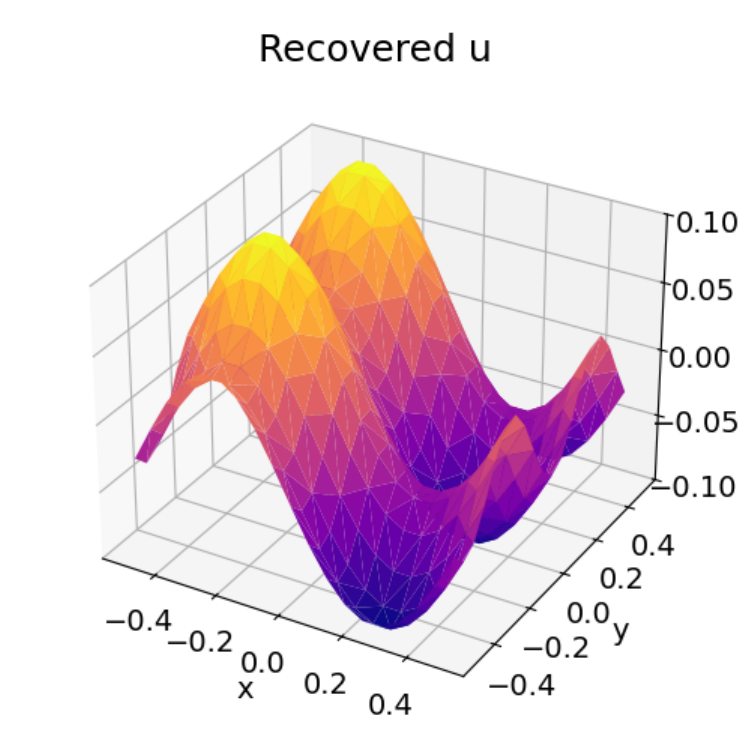}}
        \caption{Recovered $u$ via additive Schwarz method}
        \label{Multidimensioninversefig11}
    \end{subfigure}%
    \hspace{1mm}
    \begin{subfigure}[b]{0.23\textwidth}
        \centering
        {\includegraphics[width=\linewidth,page=48]{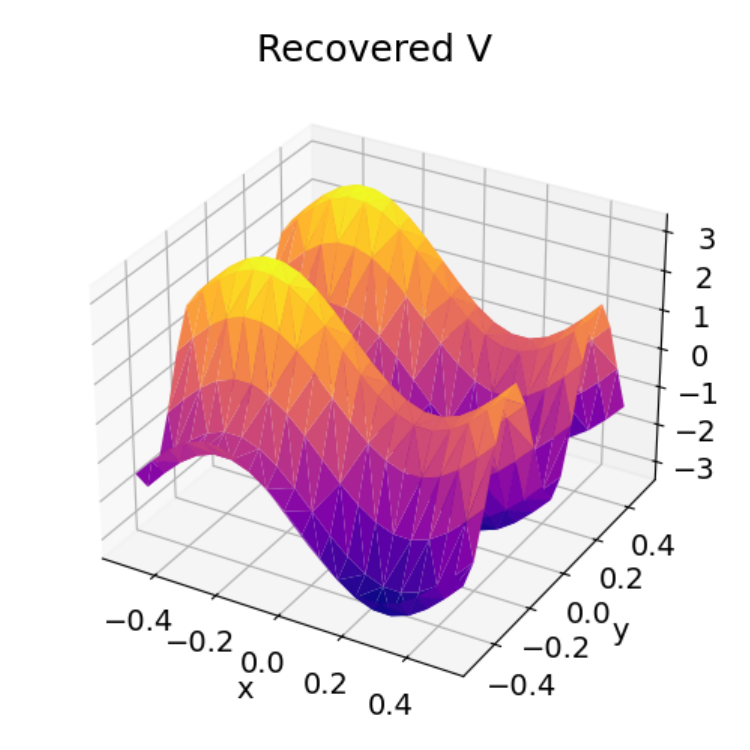}}
        \caption{Recovered $V$ via additive Schwarz method}
        \label{Multidimensioninversefig12}
    \end{subfigure}%
    \hspace{1mm}
    \begin{subfigure}[b]{0.23\textwidth}
        \centering
        {\includegraphics[width=\linewidth,page=49]{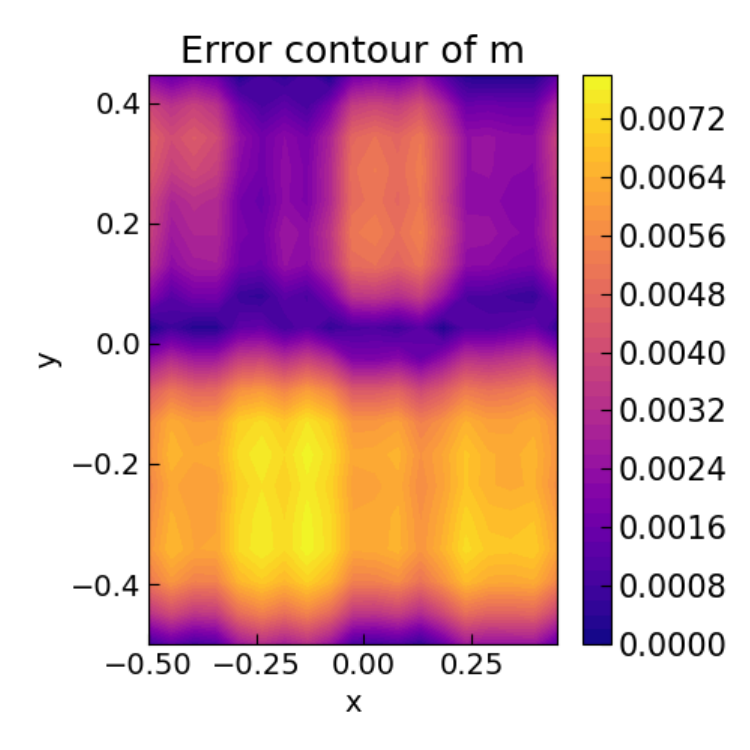}}
        \caption{Error contour of $m$ via additive Schwarz method}
        \label{Multidimensioninversefig13}
    \end{subfigure}

    \begin{subfigure}[b]{0.23\textwidth}
        \centering
        {\includegraphics[width=\linewidth,page=50]{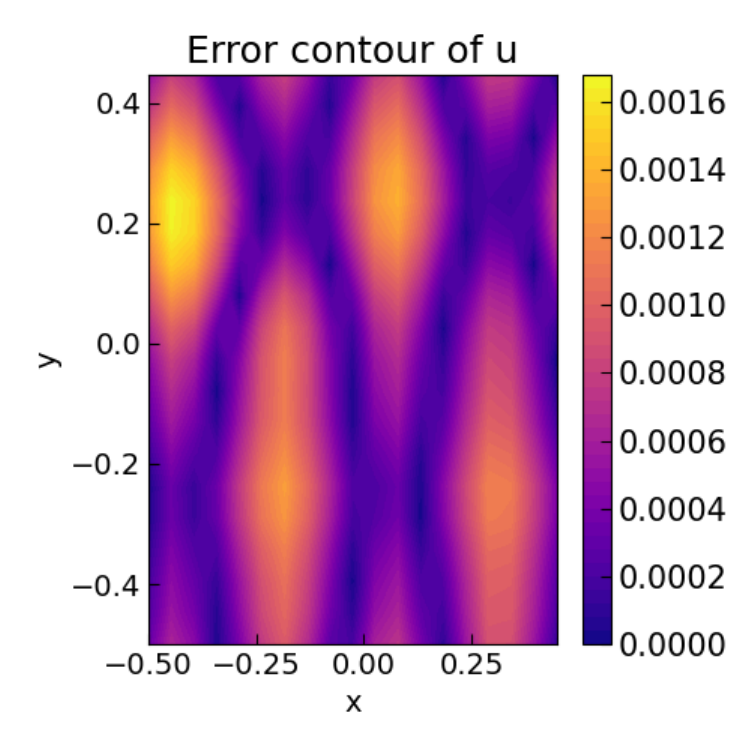}}
        \caption{Error contour of $u$ via GPPI}
        \label{Multidimensioninversefig14}
    \end{subfigure}
    \hspace{1mm}
    \begin{subfigure}[b]{0.23\textwidth}
        \centering
        {\includegraphics[width=\linewidth,page=51]{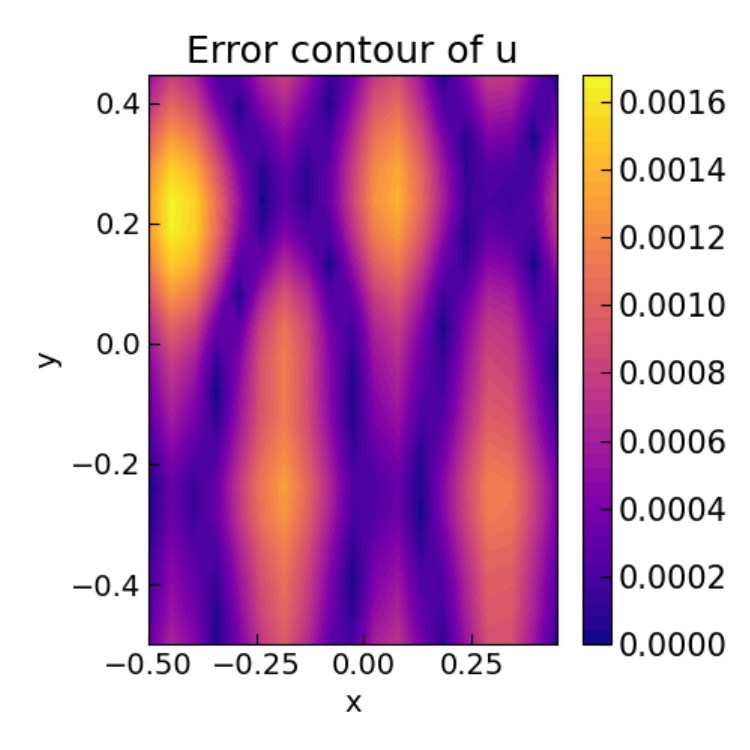}}
        \caption{Error contour of $u$ via additive Schwarz method}
        \label{Multidimensioninversefig15}
    \end{subfigure}
    \hspace{1mm}
    \begin{subfigure}[b]{0.23\textwidth}
        \centering
        {\includegraphics[width=\linewidth,page=52]{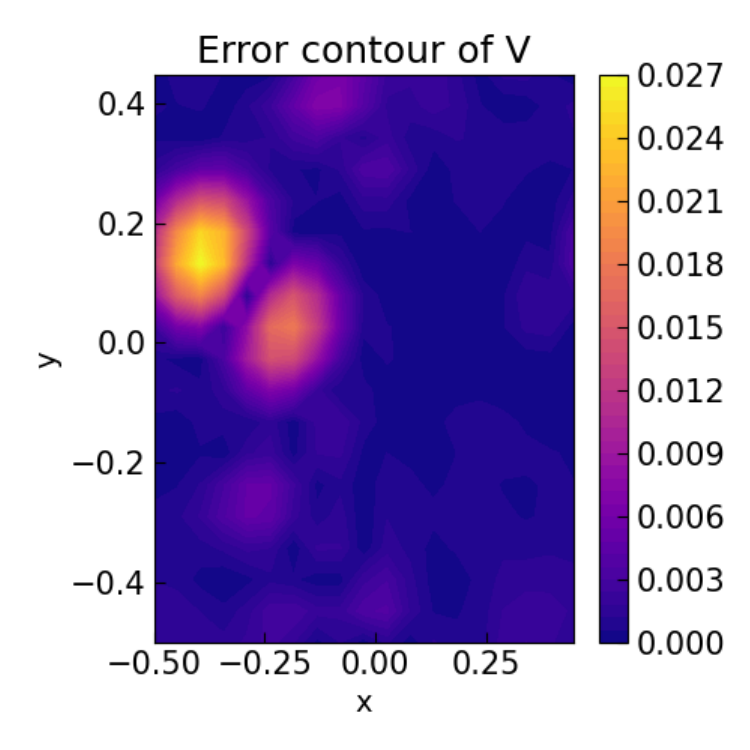}}
        \caption{Error contour of $V$ via GPPI}
        \label{Multidimensioninversefig16}
    \end{subfigure}%
    \hspace{1mm}
    \begin{subfigure}[b]{0.23\textwidth}
        \centering
        {\includegraphics[width=\linewidth,page=53]{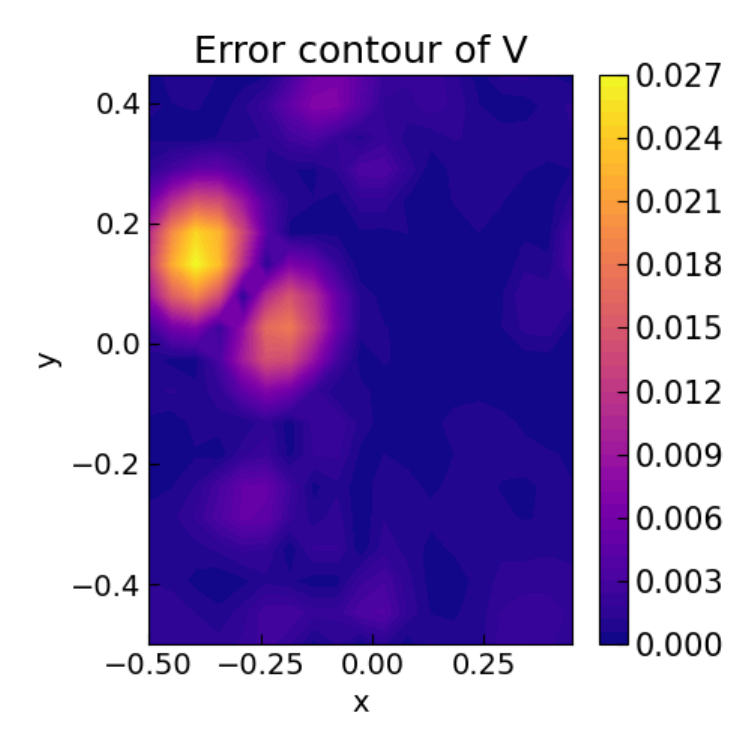}}
        \caption{Error contour of $V$ via additive Schwarz method}
        \label{Multidimensioninversefig17}
    \end{subfigure}%

    \caption{Numerical results for the two-dimensional inverse problem of MFG in \Cref{Twodimensioninverseold}. Panels (a)--(c) show the reference solutions for $m$, $u$, and $V$, and panel (d) shows the log-log plot of the error of $m$ for GPPI and the additive Schwarz method across iterations. Panels (e)--(g), (h), (m), (o) show the recovered $m$, $u$, $V$ and their {absolute pointwise }error contours via GPPI; panels (i)--(k), (l), (n), (p) show the same via the additive Schwarz method.}
    \label{multidimensionV}
\end{figure}

\textbf{Experimental Results.} 
{Figure \ref{fig:samples_observationsmulti} displays the collocation sample points for \(m,u, V\) and the observation points for both \(m\) and \(V\). Figure \ref{multidimensionV} illustrates the discretized \(L^2\) errors \(\mathcal{E}_2(m^{(k)}, m^*)\), as defined in \eqref{eq:l2disc}.}
\mathfix{These errors measure the discrepancy between the approximation \(m^{(k)}\) and the \mathfix{reference} solution \(m^*\) across the policy iterations and the  additive Schwarz  steps. Figure \ref{Multidimensioninversefig1} displays the reference solution \(m^*\), and the associated panels also include the reference solutions, the recovered results, and the {absolute }pointwise error contours for the approximated functions of \(m\), \(u\), and \(V\).} Table \ref{tablemulti} displays the numerical results for the variable \(\lambda\) utilizing various computational methods.
{Moreover, Figure~\ref{Multidimensioninversefig5} shows that \mathfix{GPPI-AS} requires fewer outer iterations than GPPI to reach a comparable error level. The GPPI method requires 43 iterations and \(8.689~\mathrm{s}\), whereas the GPPI-AS method reaches a comparable level in 4 iterations and \(4.477~\mathrm{s}\).} {Table~\ref{2Dinverseoldnoise} summarizes the relative $L^1$, $L^2$, and $L^\infty$ reconstruction errors for the two-dimensional stationary test in \Cref{Twodimensioninverseold}. In this table, relative Gaussian noise following the model \eqref{eq:relnoise} is added to the observations of $m$ and $V$. These results exhibit the same qualitative behavior as in the one-dimensional case: larger noise levels $\tilde{\gamma}$ lead to larger errors.}

To further compare the proposed method with an inverse-problem solver based on a different discretization strategy, Tables~\ref{2Dinverseoldnoise} and~\ref{tab:GN_comparison} report the relative recovery errors of $m$ under different noise levels.  Table~\ref{2Dinverseoldnoise} shows the results of the proposed GPPI-based method, while Table~\ref{tab:GN_comparison} reports the results obtained by applying the monotone flow method of~\cite{yan2026globally} with a uniform finite-difference discretization under the same setting as the experimental setup.  The two methods produce errors of the same order of magnitude.  The monotone flow method is slightly more accurate in $L^1$ and $L^2$ at low noise levels, whereas the proposed GPPI-based method gives smaller $L^\infty$ errors and is more robust at higher noise levels.

\begin{table}[H]
  \centering
  \caption{Numerical results for variable $\lambda$ in the two-dimensional MFG in \Cref{Twodimensioninverseold} under different methods. GPPI denotes the Gaussian Process Policy Iteration method, and GPPI-AS refers to the GPPI method accelerated with additive Schwarz correction.}
  \label{tablemulti}
  \begin{tabular}{|c|c|c|c|}
    \hline
    Method& Finite Difference Method  &GPPI & GPPI-AS\\
    \hline
    $\lambda$&0.9272171  & 0.9257646&0.9257647 \\
    \hline
  \end{tabular}
\end{table}

\begin{table}[H]
\centering
\caption{Relative recovery error of $m$ under different noise levels $\tilde{\gamma}$ for the two-dimensional stationary MFG in \Cref{Twodimensioninverseold} using the GPPI method.}
\begin{tabular}{lccccc}
\toprule
Noise level $\tilde{\gamma}$ & 0e+00 & 1e-03 & 5e-03 & 1e-02 & 5e-02 \\
\midrule
Relative $L^1$ error     & 4.333e-03 & 4.437e-03 & 4.922e-03 & 5.643e-03 & 1.303e-02 \\
Relative $L^2$ error      & 4.790e-03 & 4.909e-03 & 5.522e-03 & 6.511e-03 & 1.718e-02 \\
Relative $L^\infty$ error  & 5.663e-03 & 5.929e-03 & 7.782e-03 & 1.194e-02 & 4.577e-02    \\
\bottomrule
\end{tabular}
\label{2Dinverseoldnoise}
\end{table}
\begin{table}[H]
\centering
\caption{Relative recovery error of $m$ under different noise levels $\tilde{\gamma}$ for the two-dimensional stationary MFG in \Cref{Twodimensioninverseold}, using the monotone flow method of \cite{yan2026globally} with finite-difference discretization.}
\begin{tabular}{lccccc}
\toprule
Noise level $\tilde{\gamma}$ & 0e+00 & 1e-03 & 5e-03 & 1e-02 & 5e-02 \\
\midrule
Relative $L^1$ error      & 3.198e-03 & 3.105e-03 & 4.028e-03 & 6.387e-03 & 3.282e-02 \\
Relative $L^2$ error      & 4.277e-03 & 4.180e-03 & 4.969e-03 & 7.670e-03 & 3.864e-02 \\
Relative $L^\infty$ error & 1.142e-02 & 1.109e-02 & 1.036e-02 & 1.413e-02 & 6.707e-02 \\
\bottomrule
\end{tabular}
\label{tab:GN_comparison}
\end{table}

\subsubsection{Two-Dimensional Stationary MFG Inverse Problem with Non-Quadratic Coupling}
\label{Twodimensioninversenew}

In this case, we consider the MFG system in \eqref{eqonedimensioninv} for $d=2$, employing viscosity coefficient, spatial cost, and coupling functions distinct from those in \Cref{Twodimensioninverseold}. This setup serves to demonstrate the consistent acceleration of the GPPI-AS method.

\textbf{Experimental Setup.} We identify \(\mathbb{T}^2\) by \([-0.5, 0.5) \times [-0.5, 0.5)\). The viscosity coefficient is set to $\nu=0.35$. The Hamiltonian is chosen as \mathfix{$H(x,y,p)= \frac{1}{2}|p|^2$} and the spatial cost is $V(x,y)=\frac{1}{2}(\sin(2\pi x) + \cos(4\pi x)+\sin (4\pi y)).$
 The coupling function is \( F(m) = m^{3.5}\). 
 
 {A total of 625 sample points for \(m\), \(u\), and \(V\) are randomly generated in the spatial domain \(\mathbb{T}^2\). From these, 69 observation points \( \boldsymbol{m}^o \) are randomly selected, and their observed values are obtained by interpolating from a finer reference solution (computed on a uniform $30 \times 30$ grid) using cubic interpolation. The 156 observation points \( \boldsymbol{V}^o \) are randomly distributed in space, independent of the sample points. We set the regularization parameters as \( \alpha_{m^o} = \mathfix{10^5} \) and \( \alpha_{v^o} = \mathfix{10^5} \). The regularization coefficients are set to $\alpha_v = \alpha_u = \alpha_m = \alpha_\lambda = 0.5$. Gaussian noise with standard deviation $\gamma = 10^{-3}$ is added to the observations, modeled as $\mathcal{N}(0, \gamma^2 I)$. The system is initialized with $V \equiv 0$, $u \equiv 0$, $\boldsymbol{Q} \equiv \boldsymbol{0}$, $m \equiv 1$, and $\lambda = 0$. 
 For the GP priors, we employ periodic kernels. The reconstruction error is evaluated by interpolating the GP solution back onto a uniform reference grid via the GP posterior mean and comparing it with the reference.
 }

\begin{figure}[H]
    \centering
    \begin{subfigure}[b]{0.23\textwidth}
        \centering
        {\includegraphics[width=\linewidth,page=54]{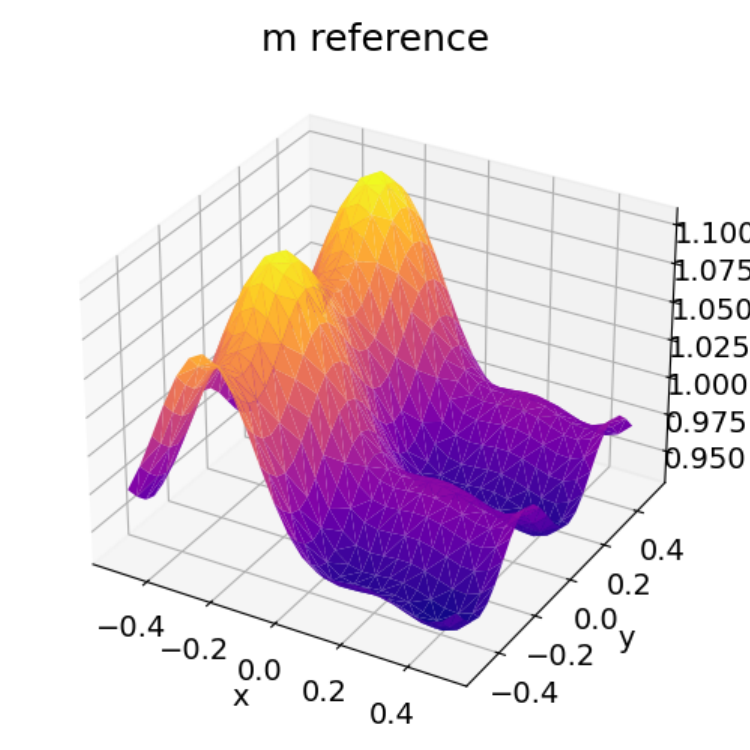}}
        \caption{$m$ reference}
        \label{Multidimensioninversefig1_2}
    \end{subfigure}%
    \hspace{1mm}
    \begin{subfigure}[b]{0.23\textwidth}
        \centering
        {\includegraphics[width=\linewidth,page=55]{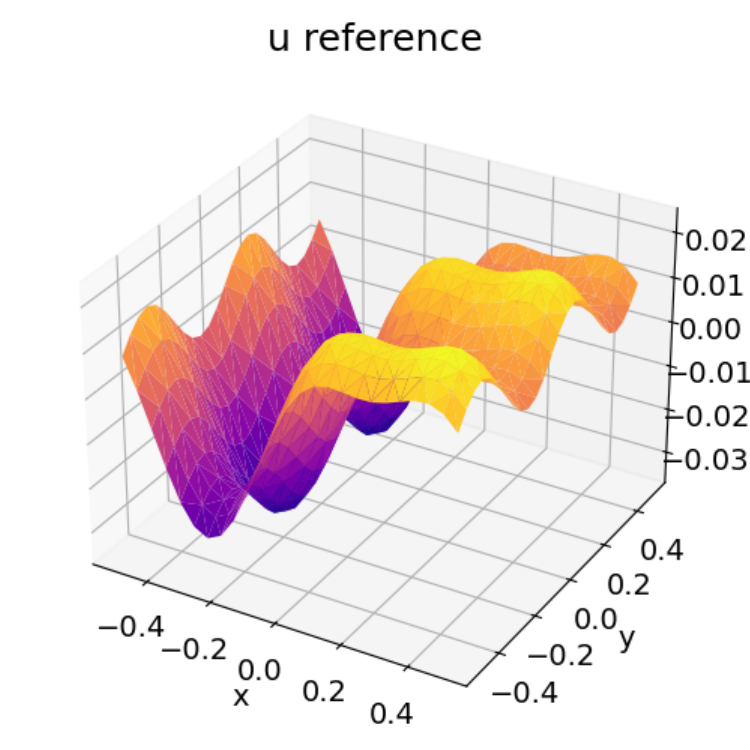}}
        \caption{$u$ reference}
        \label{Multidimensioninversefig2_2}
    \end{subfigure}%
    \hspace{1mm}
    \begin{subfigure}[b]{0.23\textwidth}
        \centering
        {\includegraphics[width=\linewidth,page=56]{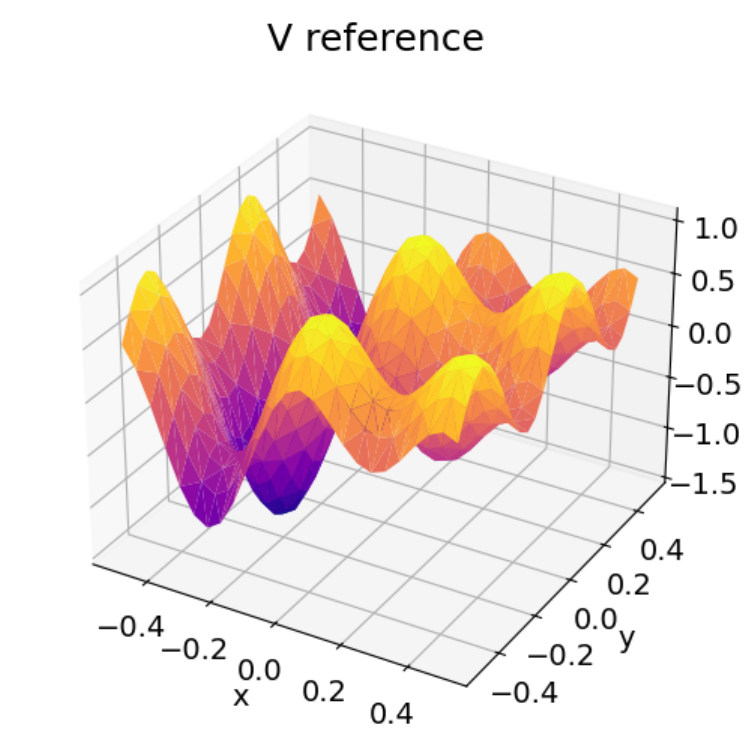}}
        \caption{$V$ reference}
        \label{Multidimensioninversefig3_2}
    \end{subfigure}%
    \hspace{1mm}
    \begin{subfigure}[b]{0.23\textwidth}
        \centering
        {\includegraphics[width=\linewidth,page=57]{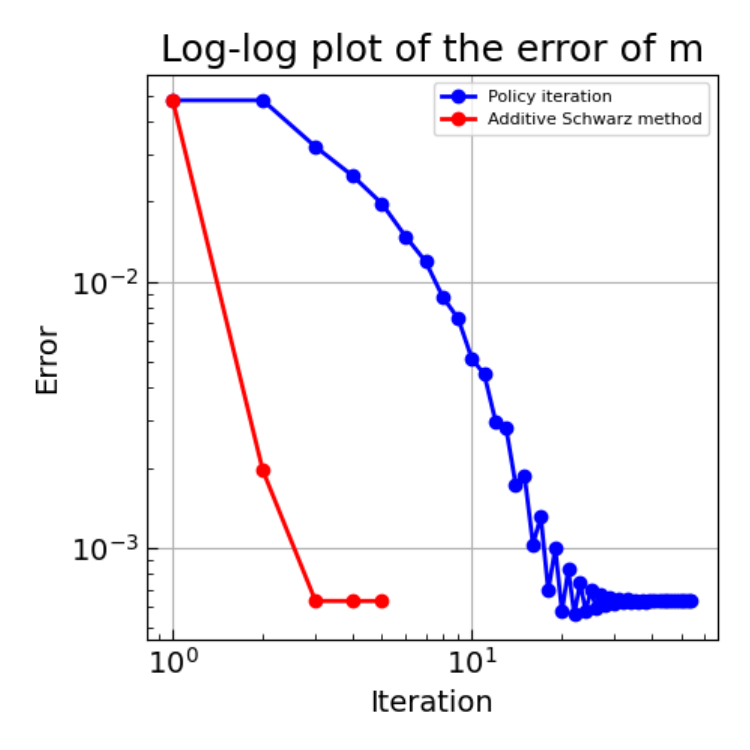}}
        \caption{Log-log error of $m$}
        \label{Multidimensioninversefig5_2}
    \end{subfigure}

    \begin{subfigure}[b]{0.23\textwidth}
        \centering
        {\includegraphics[width=\linewidth,page=58]{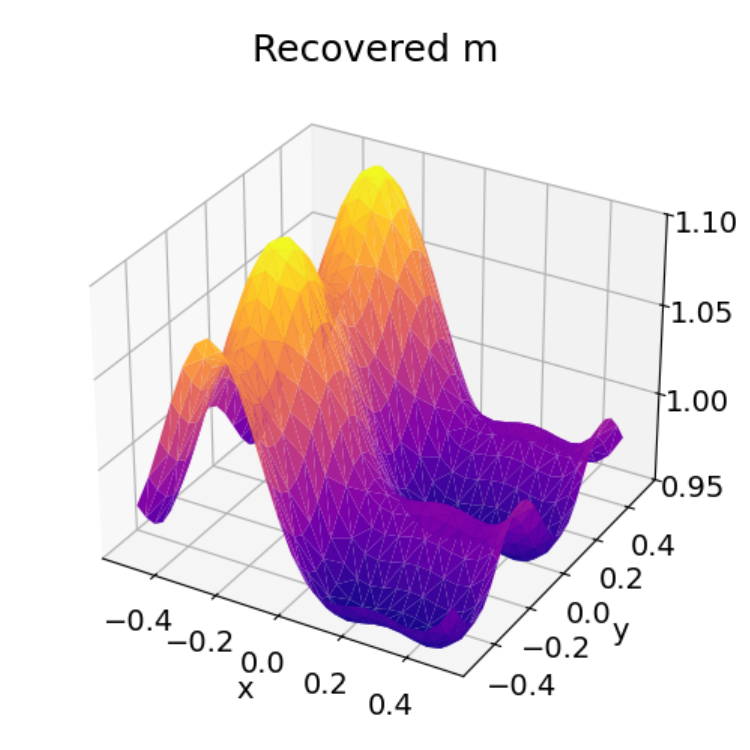}}
        \caption{Recovered $m$ via GPPI}
        \label{Multidimensioninversefig6_2}
    \end{subfigure}%
    \hspace{1mm}
    \begin{subfigure}[b]{0.23\textwidth}
        \centering
        {\includegraphics[width=\linewidth,page=59]{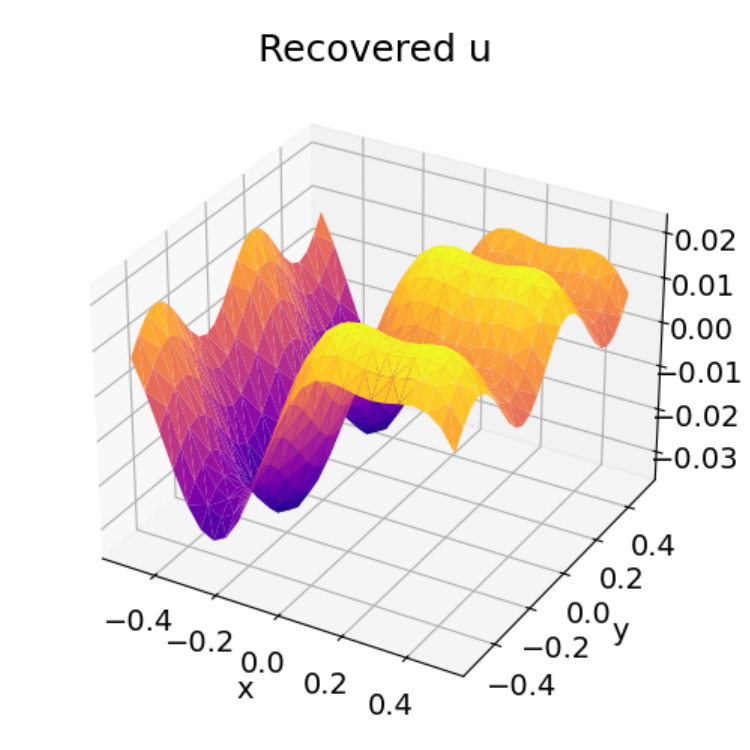}}
        \caption{Recovered $u$ via GPPI}
        \label{Multidimensioninversefig7_2}
    \end{subfigure}%
    \hspace{1mm}
    \begin{subfigure}[b]{0.23\textwidth}
        \centering
        {\includegraphics[width=\linewidth,page=60]{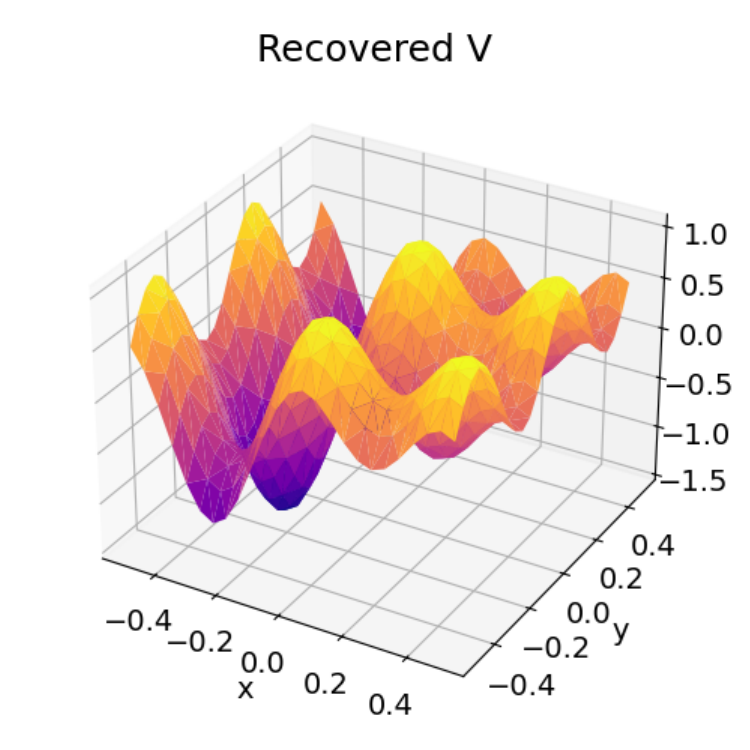}}
        \caption{Recovered $V$ via GPPI}
        \label{Multidimensioninversefig8_2}
    \end{subfigure}%
    \hspace{1mm}
    \raisebox{-3.8mm}{
    \begin{subfigure}[b]{0.23\textwidth}
        \centering
        {\includegraphics[width=\linewidth,page=61]{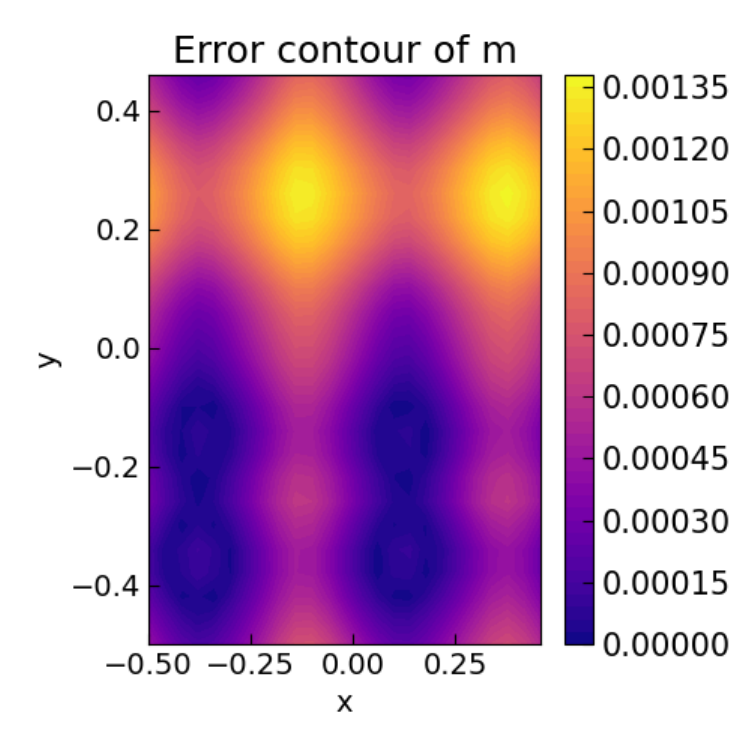}}
        \caption{Error contour of $m$ via GPPI}
        \label{Multidimensioninversefig9_2}
    \end{subfigure}}

    \begin{subfigure}[b]{0.23\textwidth}
        \centering
        {\includegraphics[width=\linewidth,page=62]{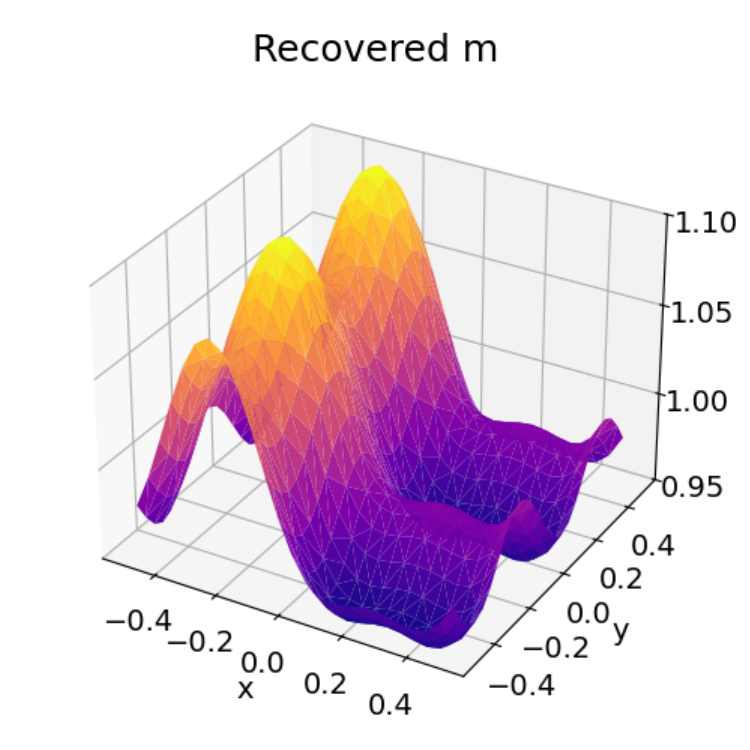}}
        \caption{Recovered $m$ via additive Schwarz method}
        \label{Multidimensioninversefig10_2}
    \end{subfigure}%
    \hspace{1mm}
    \begin{subfigure}[b]{0.23\textwidth}
        \centering
        {\includegraphics[width=\linewidth,page=63]{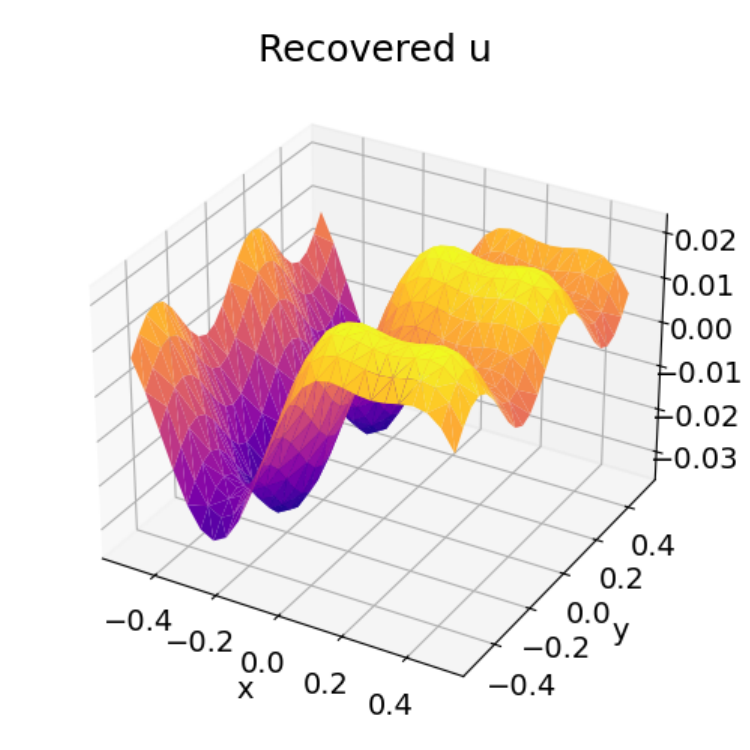}}
        \caption{Recovered $u$ via additive Schwarz method}
        \label{Multidimensioninversefig11_2}
    \end{subfigure}%
    \hspace{1mm}
    \begin{subfigure}[b]{0.23\textwidth}
        \centering
        {\includegraphics[width=\linewidth,page=64]{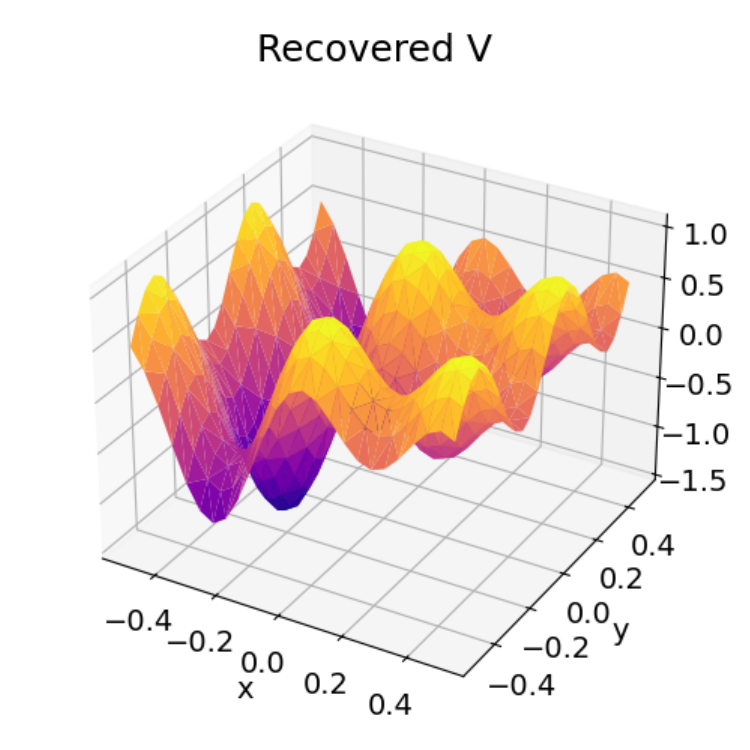}}
        \caption{Recovered $V$ via additive Schwarz method}
        \label{Multidimensioninversefig12_2}
    \end{subfigure}%
    \hspace{1mm}
    \begin{subfigure}[b]{0.23\textwidth}
        \centering
        {\includegraphics[width=\linewidth,page=65]{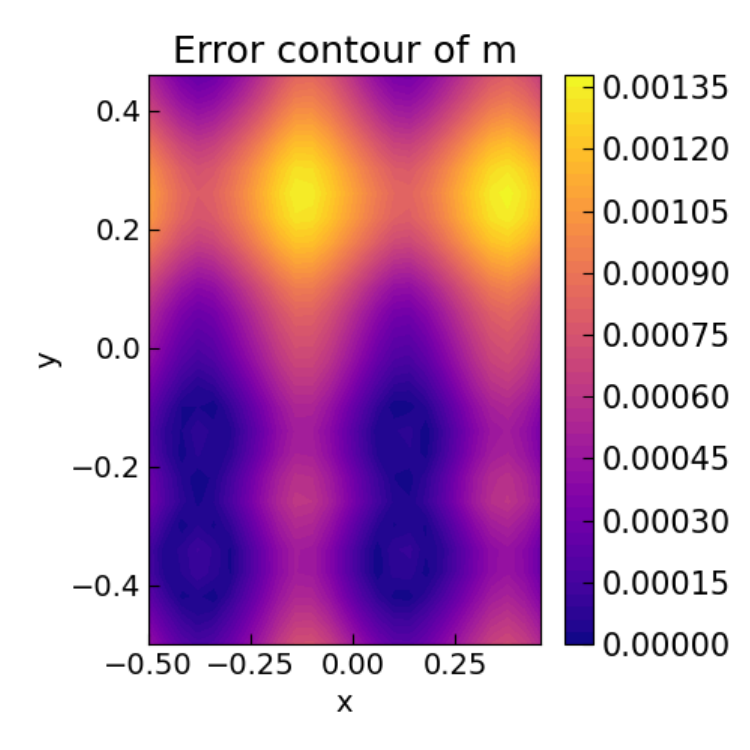}}
        \caption{Error contour of $m$ via additive Schwarz method}
        \label{Multidimensioninversefig13_2}
    \end{subfigure}

    \begin{subfigure}[b]{0.23\textwidth}
        \centering
        {\includegraphics[width=\linewidth,page=66]{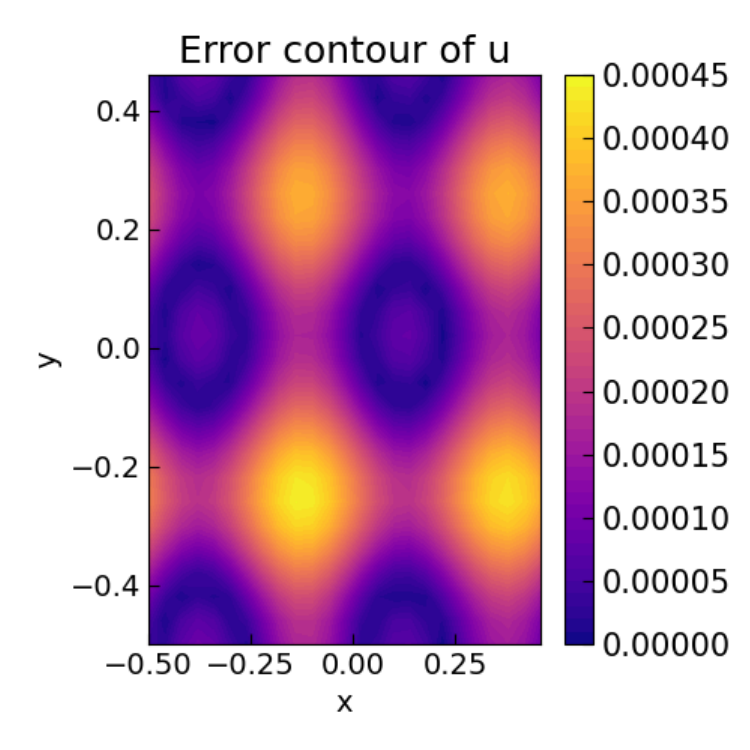}}
        \caption{Error contour of $u$ via GPPI}
        \label{Multidimensioninversefig14_2}
    \end{subfigure}
    \hspace{1mm}
    \begin{subfigure}[b]{0.23\textwidth}
        \centering
        {\includegraphics[width=\linewidth,page=67]{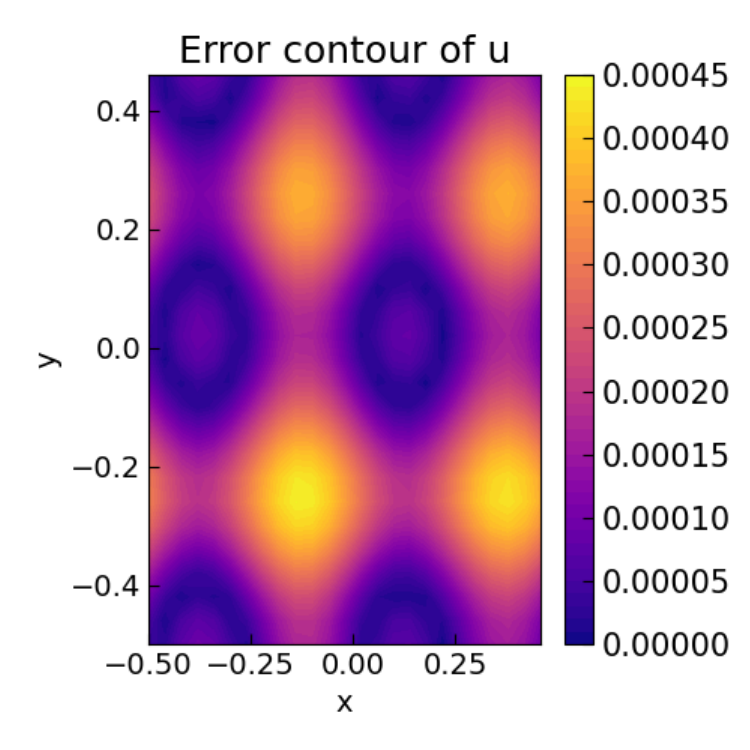}}
        \caption{Error contour of $u$ via additive Schwarz method}
        \label{Multidimensioninversefig15_2}
    \end{subfigure}
    \hspace{1mm}
    \begin{subfigure}[b]{0.23\textwidth}
        \centering
        {\includegraphics[width=\linewidth,page=68]{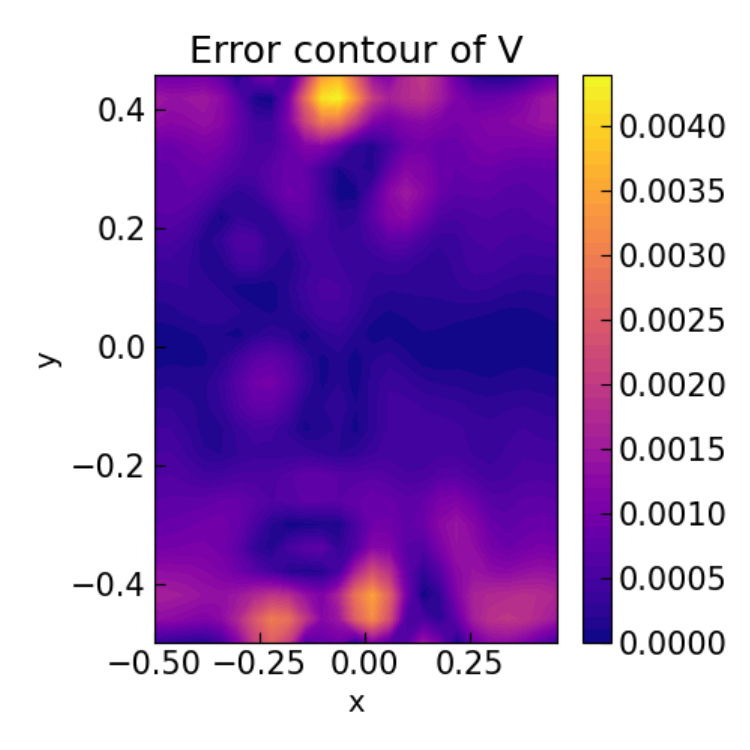}}
        \caption{Error contour of $V$ via GPPI}
        \label{Multidimensioninversefig16_2}
    \end{subfigure}%
    \hspace{1mm}
    \begin{subfigure}[b]{0.23\textwidth}
        \centering
        {\includegraphics[width=\linewidth,page=69]{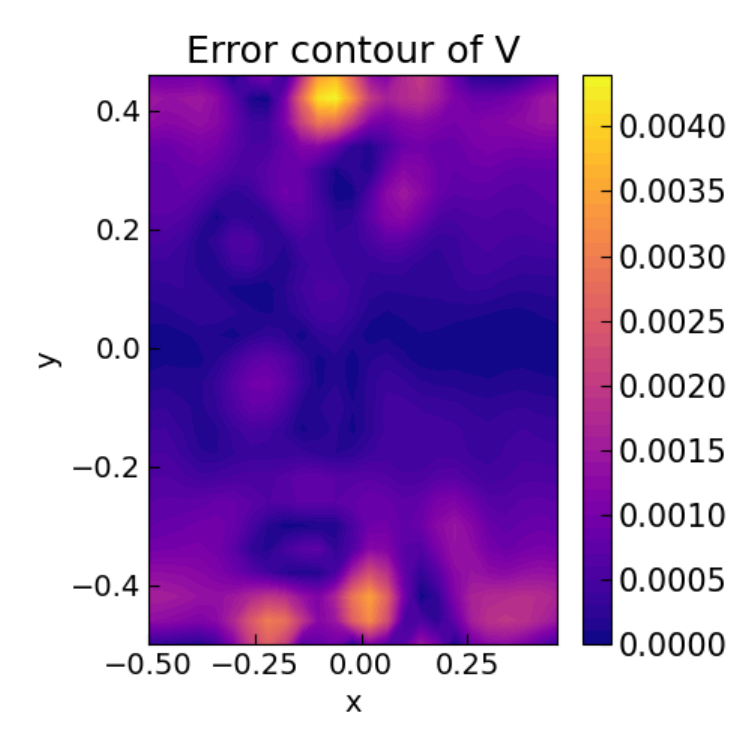}}
        \caption{Error contour of $V$ via additive Schwarz method}
        \label{Multidimensioninversefig17_2}
    \end{subfigure}
    \caption{Numerical results for the two-dimensional inverse problem of MFG in \Cref{Twodimensioninversenew}. Panels (a)--(c) show the reference solutions for $m$, $u$, and $V$, and panel (d) shows the log-log plot of the error of $m$ for GPPI and the additive Schwarz method across iterations. Panels (e)--(g), (h), (m), (o) show the recovered $m$, $u$, $V$ and their {absolute pointwise }error contours via GPPI; panels (i)--(k), (l), (n), (p) show the same via the additive Schwarz method.}
    \label{multidimensionV2}
\end{figure}

\begin{figure}[h]
    \centering
    \begin{subfigure}[b]{0.23\textwidth}
        \centering
        {\includegraphics[width=\linewidth,page=70]{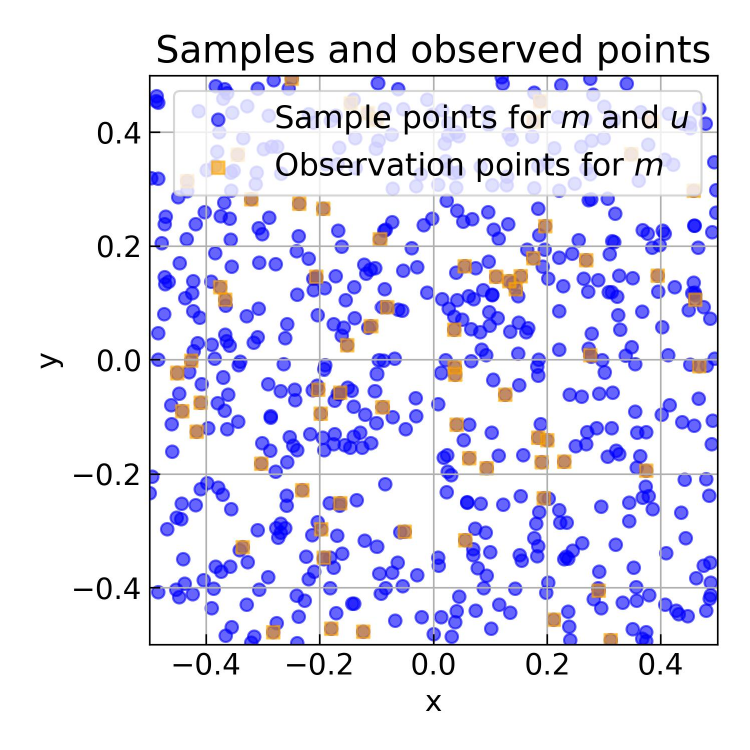}}
        \caption{Samples for $m$ and $u$ \& observations for $m$}
        \label{Multidimensioninversefig4_2}
    \end{subfigure}
    \hspace{5mm}
    \begin{subfigure}[b]{0.23\textwidth}
        \centering
        {\includegraphics[width=\linewidth,page=71]{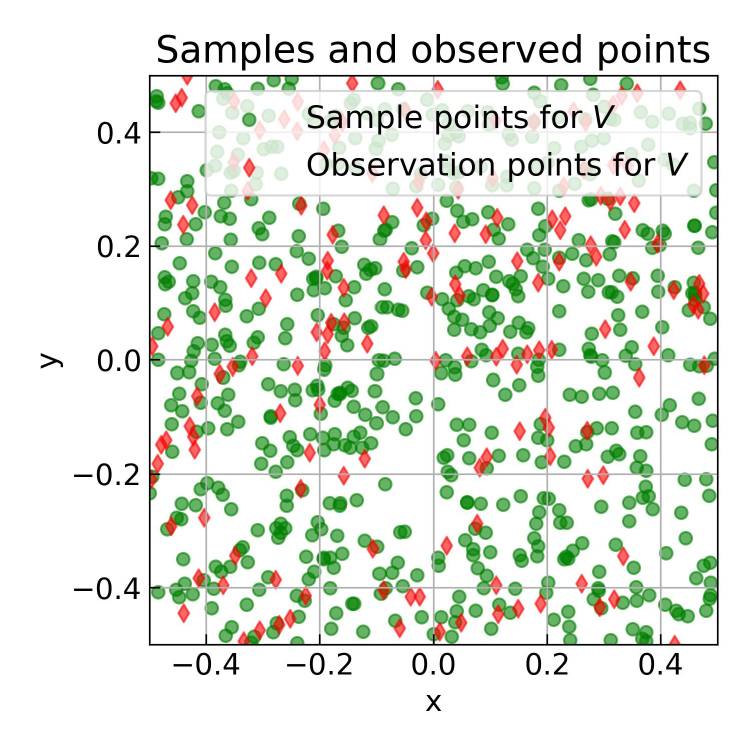}}
        \caption{Samples \& observation points for $V$}
        \label{Multidimensioninversefig0_2}
    \end{subfigure}
    \caption{The inverse problem of the two-dimensional stationary MFG in \Cref{Twodimensioninversenew}: samples for $m, u,$ and $V$ \& observations for $m$ and $V$.}
    \label{fig:samples_observationsmulti2}
\end{figure}

\textbf{Experimental Results.}
Figure~\ref{fig:samples_observationsmulti2} shows the collocation points for
$m,u,V$ together with the observation locations for $m$ and $V$.  Figure
\ref{multidimensionV2} displays the discretized $L^2$ errors
$\mathcal{E}_2(m^{(k)},m^*)$ from \eqref{eq:l2disc} over the iterations, along with
the reference solutions, the reconstructed fields, and the pointwise error
contours for $m$, $u$, and $V$.  Table~\ref{tablemulti2} reports the
corresponding values of~$\lambda$ for the finite-difference, GPPI, and
GPPI-AS solvers. {Figure~\ref{Multidimensioninversefig5_2} shows that additive Schwarz acceleration substantially reduces the computational cost. GPPI takes $25.010~\mathrm{s}$ and converges in $51$ outer iterations, whereas GPPI-AS takes $9.184~\mathrm{s}$ and converges in $4$ outer iterations.} For the two-dimensional stationary
inverse problem in \Cref{Twodimensioninversenew}, 
Tables~\ref{2Dinversenewobservation2} and~\ref{2Dinversenewnoise2} summarize
{the relative $L^1$, $L^2$, and $L^\infty$ errors of the reconstruction. In Table~\ref{2Dinversenewnoise2}, relative Gaussian noise following the model \eqref{eq:relnoise} is added to the observations of $m$ and $V$. The}
errors decrease as the common number of observation points for $m$ and $V$ increases. \mathfix{For the observation-number study in Table~\ref{2Dinversenewobservation2}, we use a separate setting in which $N_m$ and $N_v$ are kept equal and varied together.} 
For a fixed set of observation points, as in the experimental setup, Table~\ref{2Dinversenewnoise2} shows that the relative error increases as the noise level \(\tilde{\gamma}\) increases. The reconstructions of \(m\), \(u\), and \(\lambda\) exhibit the same qualitative behavior.

\begin{table}[H]
\centering
\caption{Relative recovery error of $m$ under different numbers of observations for the two-dimensional stationary MFG in \Cref{Twodimensioninversenew}.}
\begin{tabular}{lccccc}
\toprule
Number of observations & 12 & 17 & 25 & 39 & 69 \\
\midrule
Relative $L^1$ error      & 2.794e-02 & 2.150e-02 & 8.417e-03 & 2.490e-03 & 6.246e-04 \\
Relative $L^2$ error      & 3.586e-02 & 2.782e-02 & 1.096e-02 & 3.233e-03 & 7.936e-04 \\
Relative $L^\infty$ error & 8.137e-02 & 7.472e-02 & 3.235e-02 & 1.037e-02 & 1.659e-03 \\
\bottomrule
\end{tabular}
\label{2Dinversenewobservation2}
\end{table}

\begin{table}[H]
\centering
\caption{Relative recovery error of $m$ under different noise levels $\tilde{\gamma}$ for the two-dimensional stationary MFG in \Cref{Twodimensioninversenew}.}
\begin{tabular}{lccccc}
\toprule
Noise level $\tilde{\gamma}$ & 0e+00 & 1e-03 & 5e-03 & 1e-02 & 5e-02 \\
\midrule
Relative $L^1$ error      & 5.258e-04 & 5.267e-04 & 5.463e-04 & 5.913e-04 & 1.248e-03 \\
Relative $L^2$ error      & 6.026e-04 & 6.064e-04 & 6.328e-04 & 6.887e-04 & 1.547e-03 \\
Relative $L^\infty$ error & 1.145e-03 & 1.166e-03 & 1.276e-03 & 1.478e-03 & 4.530e-03 \\
\bottomrule
\end{tabular}
\label{2Dinversenewnoise2}
\end{table}

\begin{table}[H]
  \centering
  \caption{Numerical results for variable $\lambda$ in the two-dimensional MFG in \Cref{Twodimensioninversenew} under different methods. GPPI denotes the Gaussian Process Policy Iteration method, and GPPI-AS refers to the GPPI method accelerated with additive Schwarz correction.}
  \label{tablemulti2}
  \begin{tabular}{|c|c|c|c|}
    \hline
    Method& Finite Difference Method  &GPPI & GPPI-AS\\
    \hline
    $\lambda$&1.0013002  & 0.9995050&0.9995050 \\
    \hline
  \end{tabular}
\end{table}

\subsection{The Time-Dependent MFG Inverse Problem}
\label{Timedependentexperiment}
In this subsection, we study the inverse problem for the following time-dependent MFG  
\begin{equation}
\begin{cases}
-\partial_t u - \nu \Delta u + \mathfix{H(x,t,\nabla u)}  = F(m) + V(x), & \forall (x, t) \in \mathbb{T}^d \times (0, T), \\ 
\partial_t m - \nu\Delta m - \operatorname{div}\left( \mathfix{D_p H(x,t,\nabla u)} \, m \right) = 0, & \forall (x, t) \in \mathbb{T}^d \times (0, T), \\
m(x, 0)=m_0(x), \quad u(x, T)=U_T(x), & \forall x\in \mathbb{T}^d.
\end{cases}
\label{eqtimedepend}
\end{equation}

\mathfix{The Hamiltonian is set as  \(H(x,t,p) = \frac{1}{2} |p|^2\).} Let  \( \mathbb{T}\) be identified with \([-0.5, 0.5) \), $d=1$, $T=1$,  \( m_0(\cdot) = 1 \), and \( U_T(\cdot) = 0 \). The coupling is \(F(m) = m^{4}\), and the spatial cost function is \(V(x) = 0.5\bigl(\sin(2\pi x) + 3\cos(2\pi x)\bigr)\). 
The viscosity coefficient is \(\nu = \frac{1}{3}\). Given \(\nu\), \(F\), and \(V\), we solve  \eqref{eqtimedepend} using the PI algorithm \cite{cacace2021policy} to obtain the reference solutions \((u^*, m^*)\). 

The inverse problem then seeks to recover \(u\), \(m\), \(V\) from noisy, partial observations of \(m\) and \(V\) via the GPPI algorithm and \mathfix{GPPI-AS} detailed in Sections \ref{MFG Forward Problem Time Dependent} and \ref{Newton Acceleration}. We compare the errors from these approaches with the reference solutions. 

\begin{figure}[H]
    {
    \centering    \includegraphics[width=0.23\textwidth,page=72]{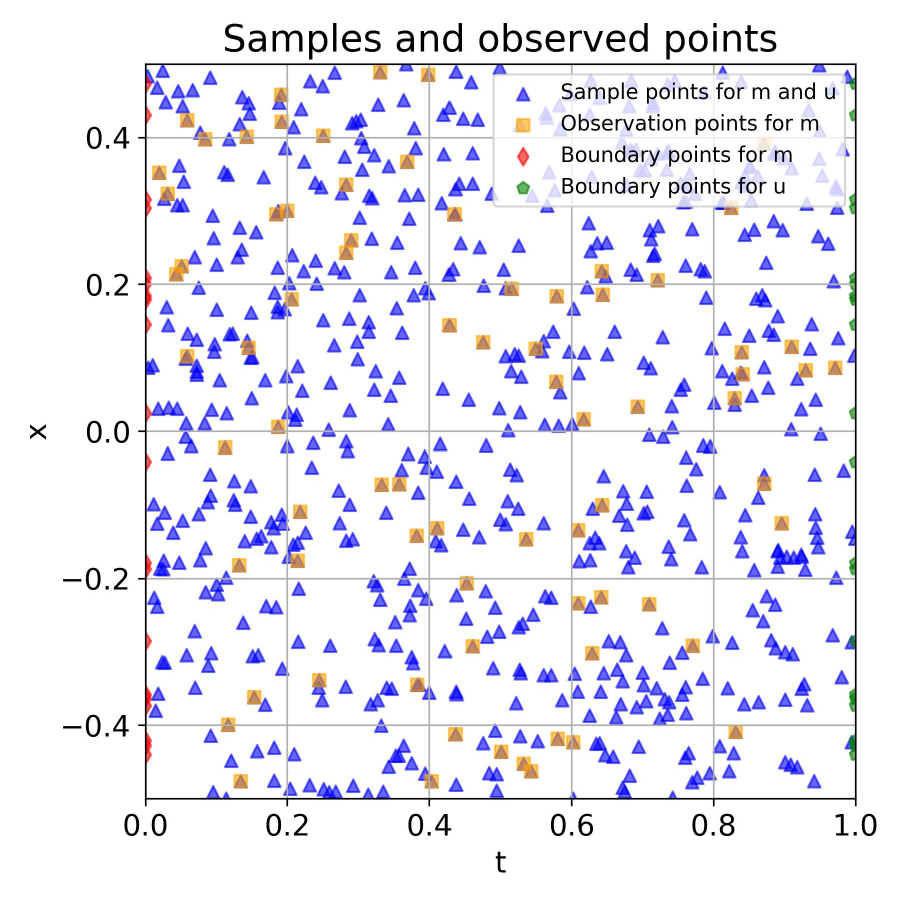}
    }
    \caption{The inverse problem of the time-dependent MFG in \eqref{eqtimedepend}: samples for $m$ and $u$ \& observations for $m$.}
    \label{Timedependentinversefig4}
\end{figure}

\begin{figure}[H]
    \centering
    \begin{subfigure}[b]{0.23\textwidth}
        \centering
        {\includegraphics[width=\linewidth,page=73]{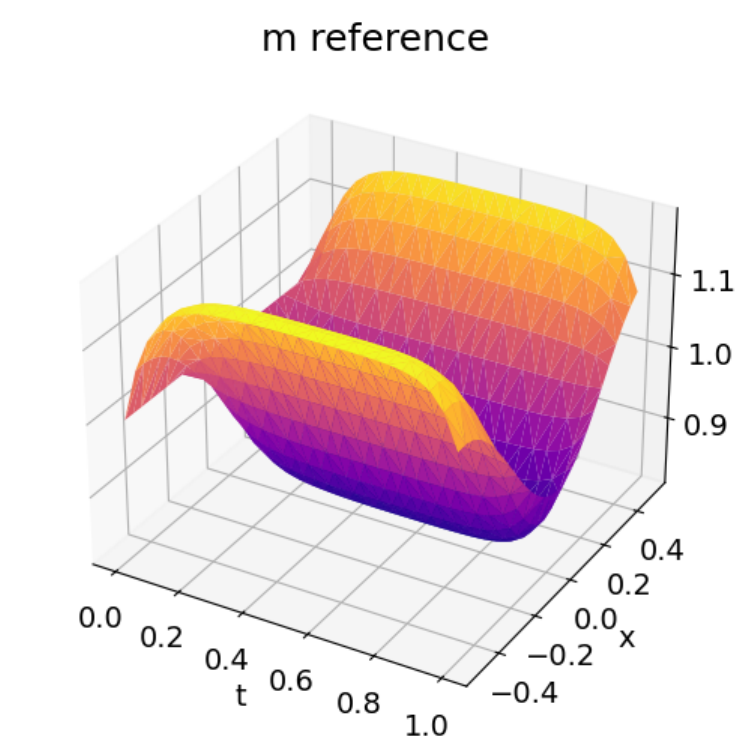}}
        \caption{$m$ reference}
        \label{Timedependentinversefig1}
    \end{subfigure}%
    \hspace{1mm}
    \begin{subfigure}[b]{0.23\textwidth}
        \centering
        {\includegraphics[width=\linewidth,page=74]{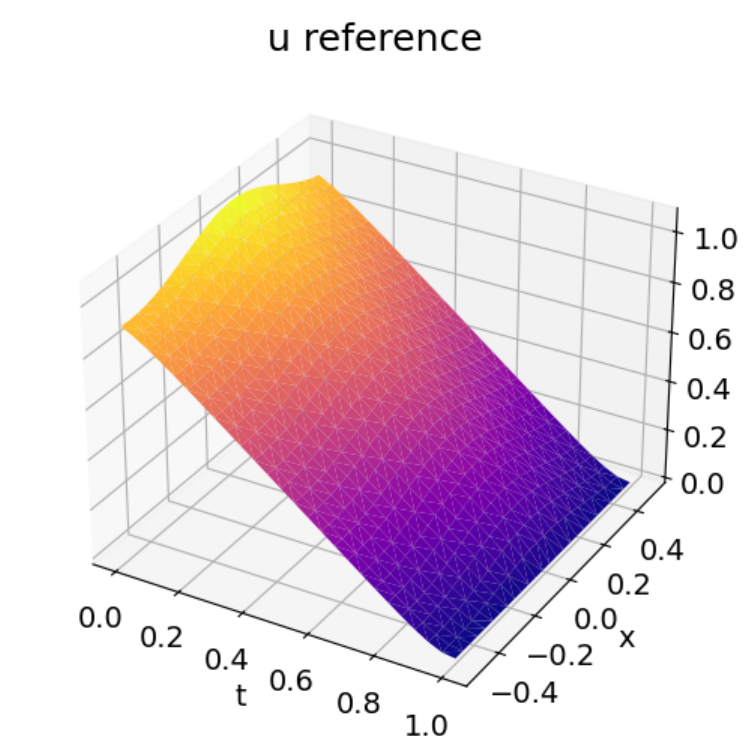}}
        \caption{$u$ reference}
        \label{Timedependentinversefig2}
    \end{subfigure}%
    \hspace{1mm}
    \begin{subfigure}[b]{0.23\textwidth}
        \centering
        {\includegraphics[width=\linewidth,page=75]{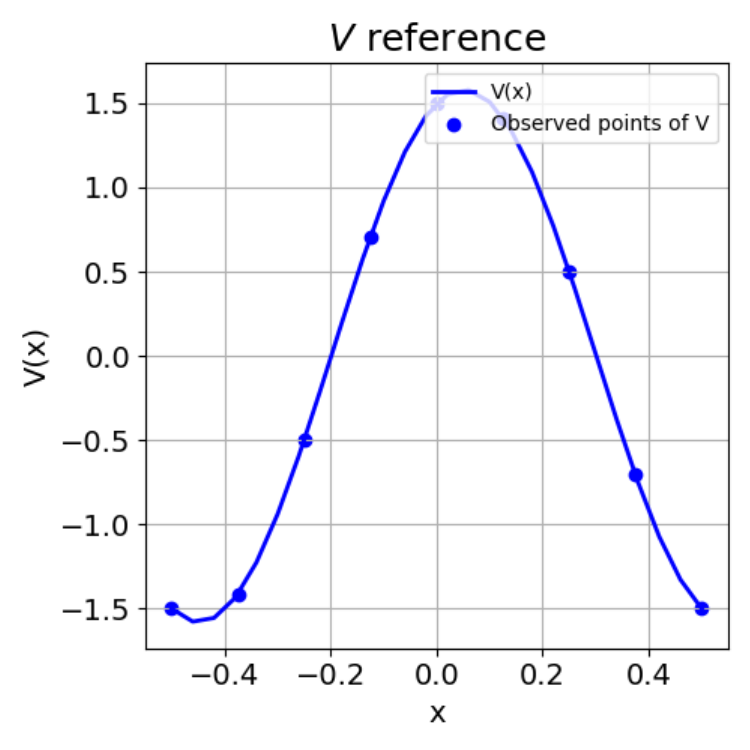}}
        \caption{$V$ reference}
        \label{Timedependentinversefig3}
    \end{subfigure}%
    \hspace{1mm}
    \begin{subfigure}[b]{0.23\textwidth}
        \centering
        {\includegraphics[width=\linewidth,page=76]{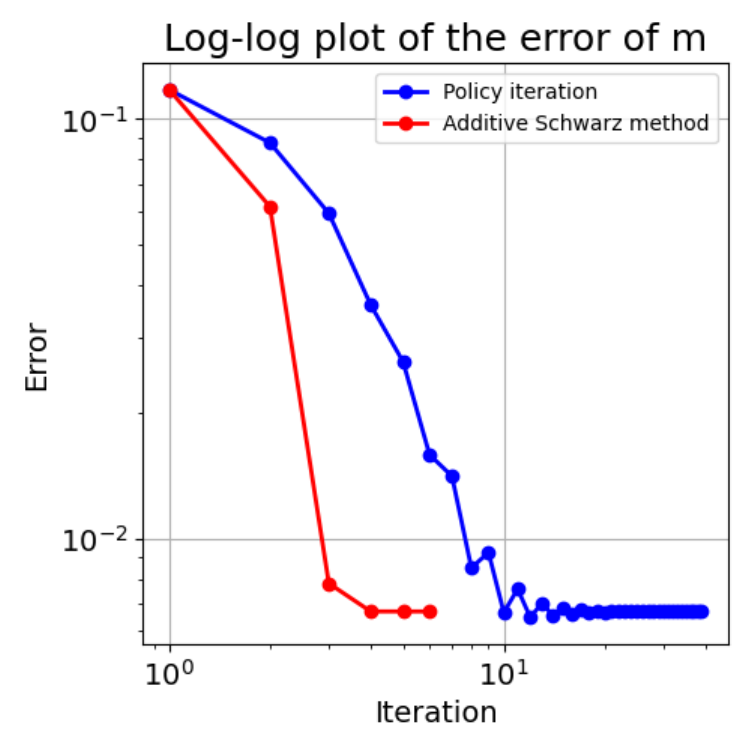}}
        \caption{Log-log error of $m$}
        \label{Timedependentinversefig5}
    \end{subfigure}
    
    \begin{subfigure}[b]{0.23\textwidth}
        \centering
        {\includegraphics[width=\linewidth,page=77]{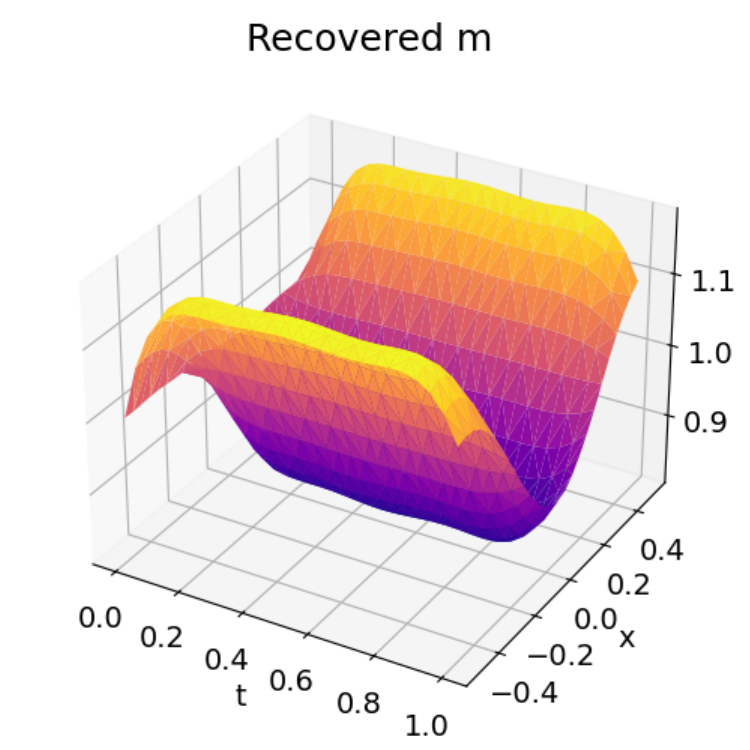}}
        \caption{Recovered $m$ via GPPI}
        \label{Timedependentinversefig6}
    \end{subfigure}%
    \hspace{1mm}
    \begin{subfigure}[b]{0.23\textwidth}
        \centering
        {\includegraphics[width=\linewidth,page=78]{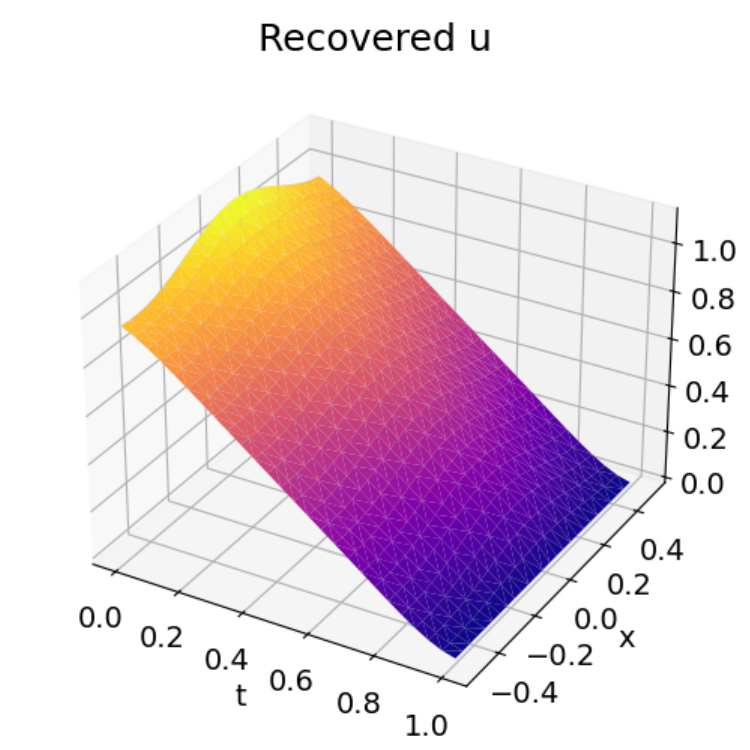}}
        \caption{Recovered $u$ via GPPI}
        \label{Timedependentinversefig7}
    \end{subfigure}%
    \hspace{1mm}
    \begin{subfigure}[b]{0.23\textwidth}
        \centering
        {\includegraphics[width=\linewidth,page=79]{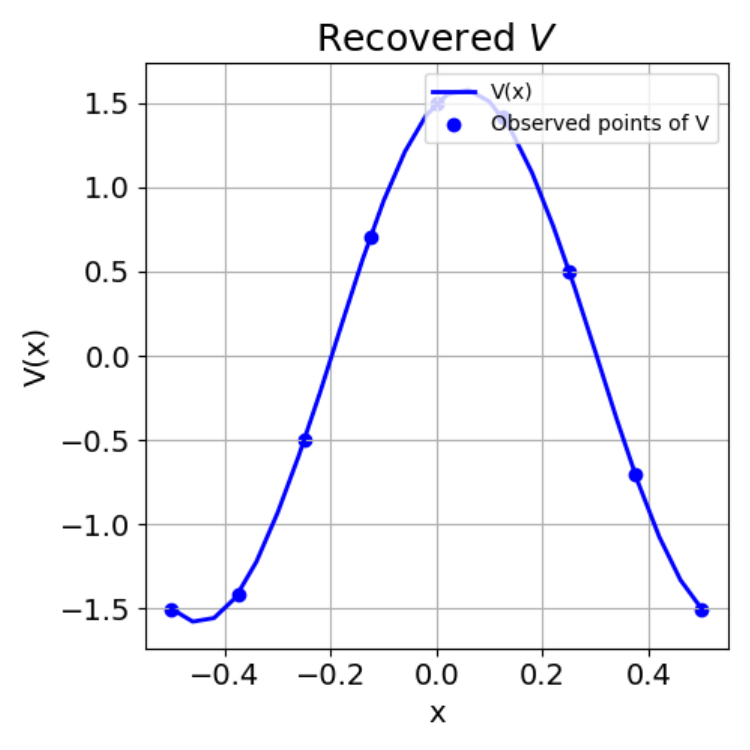}}
        \caption{Recovered $V$ via GPPI}
        \label{Timedependentinversefig8}
    \end{subfigure}%
    \hspace{1mm}
    \begin{subfigure}[b]{0.23\textwidth}
        \centering
        {\includegraphics[width=\linewidth,page=80]{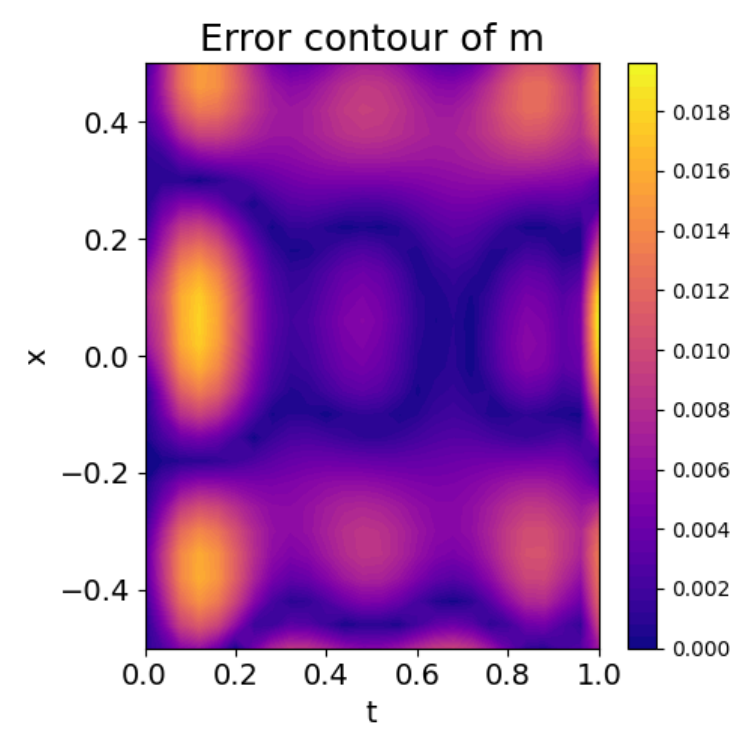}}
        \caption{Error of $m$ via GPPI}
        \label{Timedependentinversefig9}
    \end{subfigure}
    
    \begin{subfigure}[b]{0.23\textwidth}
        \centering
        {\includegraphics[width=\linewidth,page=81]{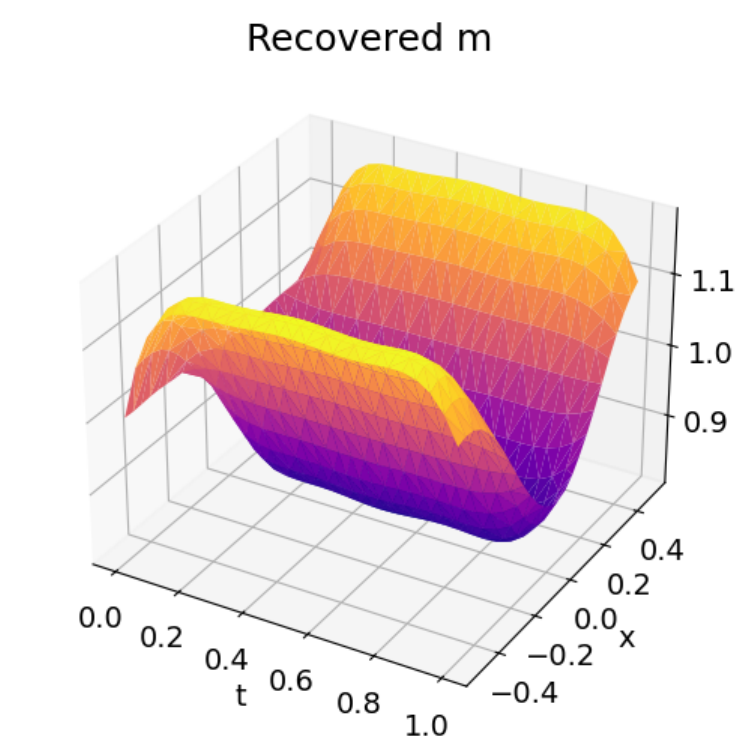}}
        \caption{Recovered $m$ via additive Schwarz}
        \label{Timedependentinversefig10}
    \end{subfigure}%
    \hspace{1mm}
    \begin{subfigure}[b]{0.23\textwidth}
        \centering
        {\includegraphics[width=\linewidth,page=82]{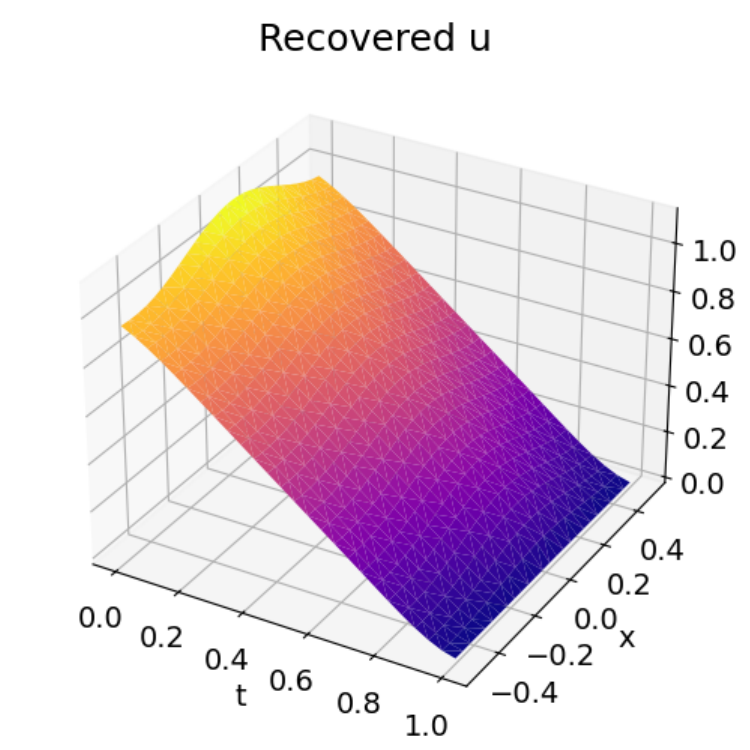}}
        \caption{Recovered $u$ via additive Schwarz}
        \label{Timedependentinversefig11}
    \end{subfigure}%
    \hspace{1mm}
    \begin{subfigure}[b]{0.23\textwidth}
        \centering
        {\includegraphics[width=\linewidth,page=83]{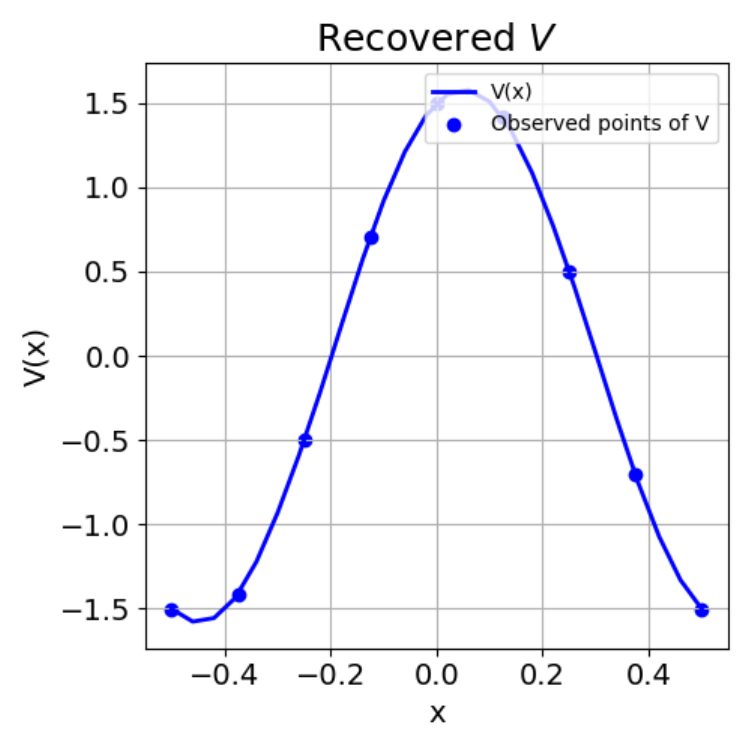}}
        \caption{Recovered $V$ via additive Schwarz}
        \label{Timedependentinversefig12}
    \end{subfigure}%
    \hspace{1mm}
    \begin{subfigure}[b]{0.23\textwidth}
        \centering
        {\includegraphics[width=\linewidth,page=84]{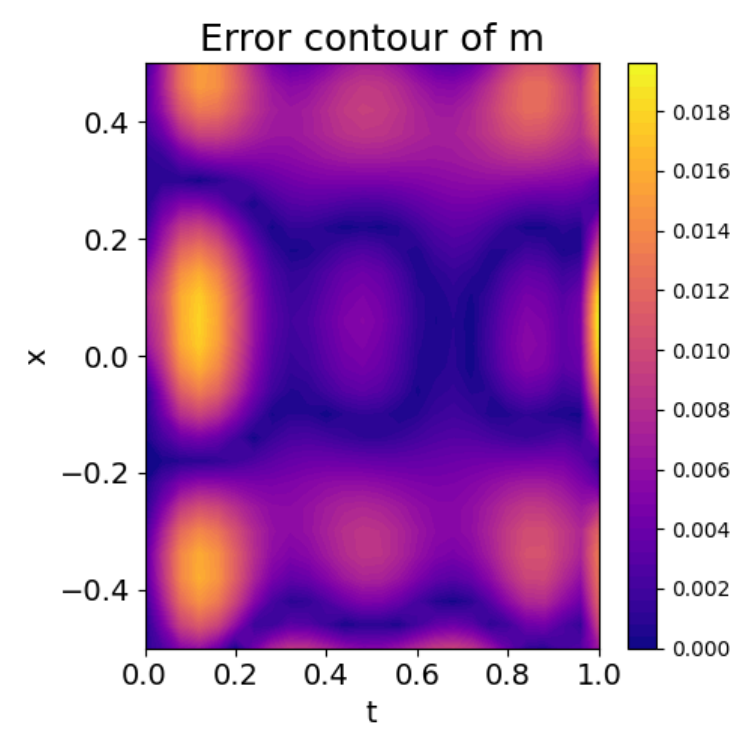}}
        \caption{Error of $m$ via additive Schwarz}
        \label{Timedependentinversefig13}
    \end{subfigure}
    
    \begin{subfigure}[b]{0.23\textwidth}
        \centering
        {\includegraphics[width=\linewidth,page=85]{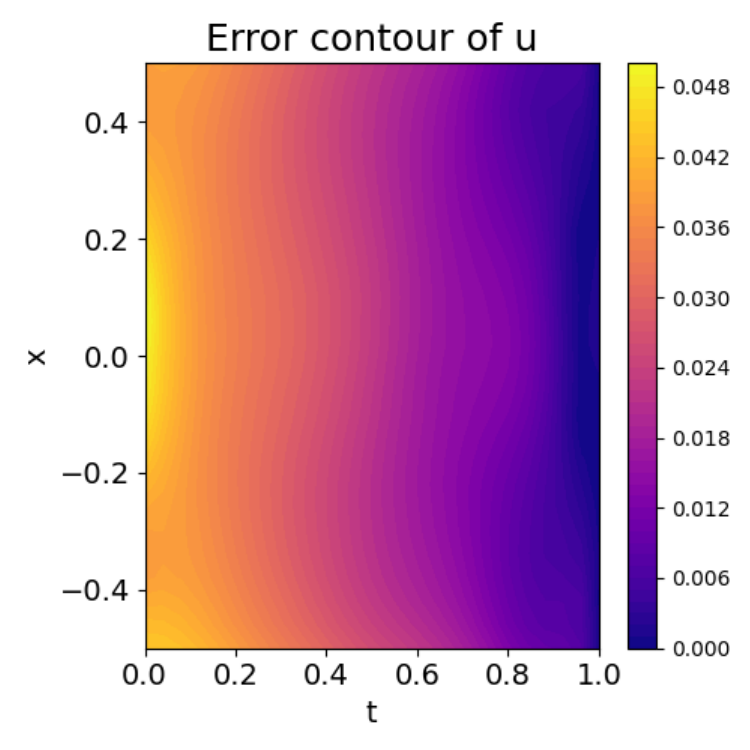}}
        \caption{Error of $u$ via GPPI}
        \label{Timedependentinversefig14}
    \end{subfigure}
    \hspace{1mm}
    \raisebox{-3.8mm}{%
    \begin{subfigure}[b]{0.23\textwidth}
        \centering
        {\includegraphics[width=\linewidth,page=86]{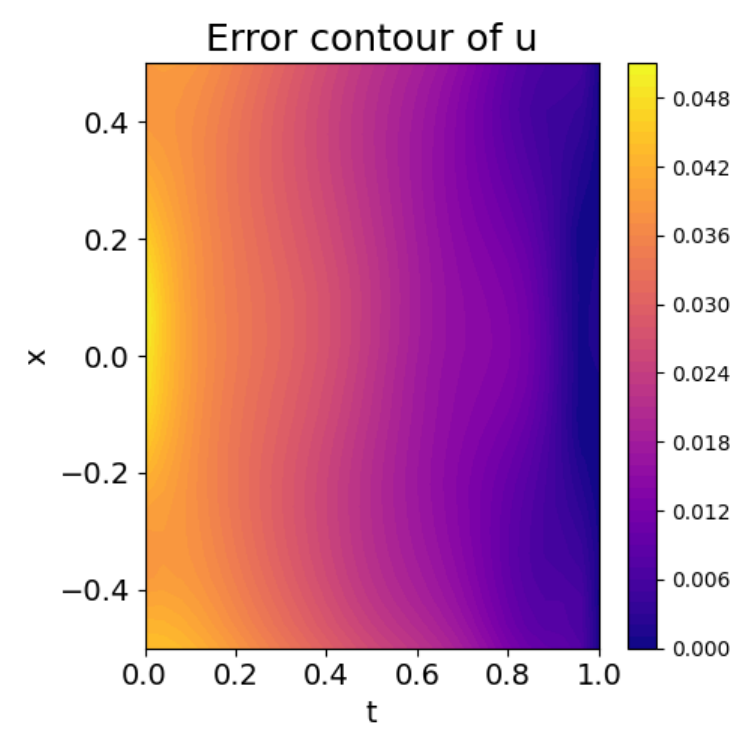}}
        \caption{Error of $u$ via additive Schwarz}
        \label{Timedependentinversefig15}
    \end{subfigure}}
    \hspace{1mm}
    \begin{subfigure}[b]{0.23\textwidth}
        \centering
        {\includegraphics[width=\linewidth,page=87]{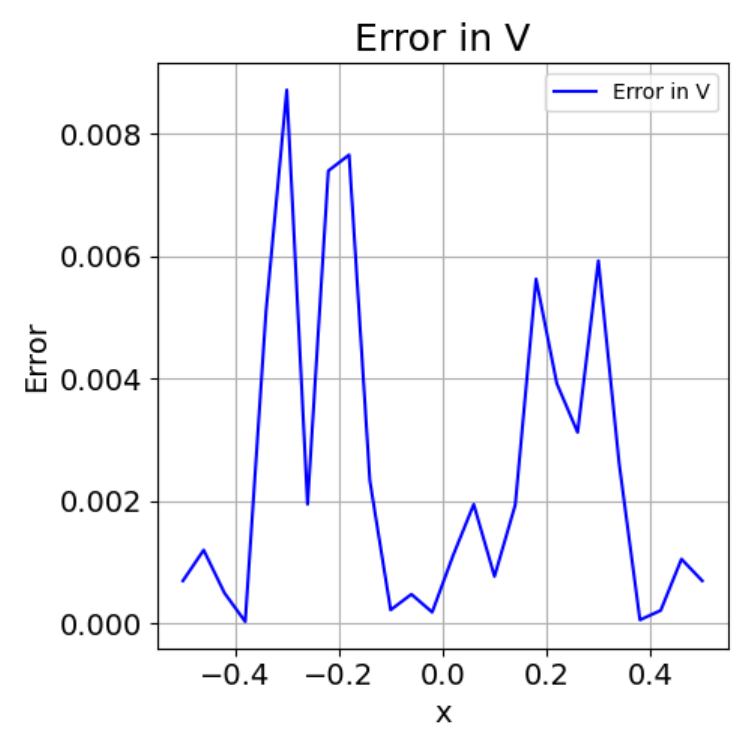}}
        \caption{Error of $V$ via GPPI}
        \label{Timedependentinversefig16}
    \end{subfigure}%
    \hspace{1mm}
    \raisebox{-3.8mm}{%
    \begin{subfigure}[b]{0.23\textwidth}
        \centering
        {\includegraphics[width=\linewidth,page=88]{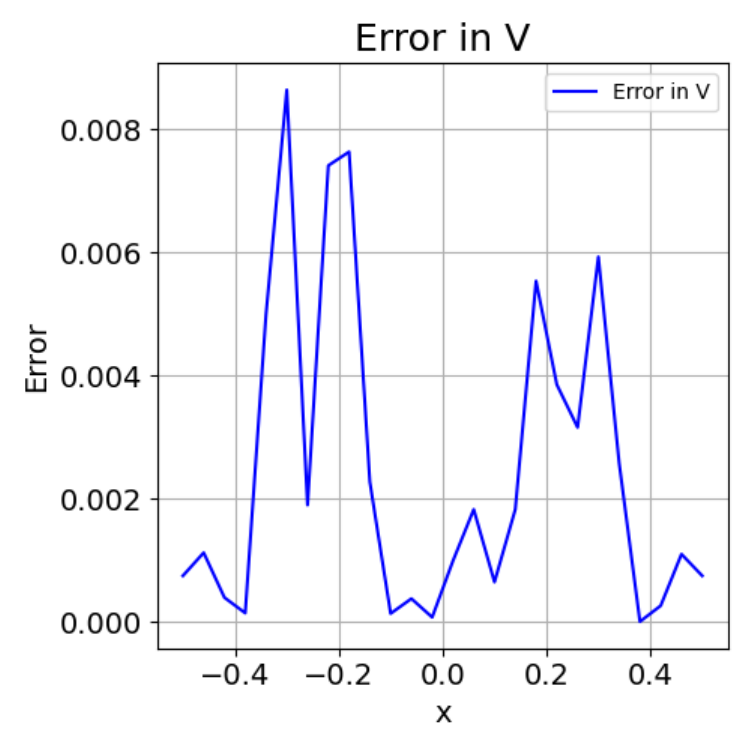}}
        \caption{Error of $V$ via additive Schwarz}
        \label{Timedependentinversefig17}
    \end{subfigure}}%

    \caption{Numerical results for the inverse problem of the time-dependent MFG in \eqref{eqtimedepend}. Panels (a)--(c) show the reference solutions for $m$, $u$, and $V$, and panel (d) shows the log-log plot of the error of $m$ for GPPI and the additive Schwarz method across iterations. Panels (e)--(g), (h), (m), (o) show the recovered $m$, $u$, $V$ and their {absolute pointwise }errors via GPPI; panels (i)--(k), (l), (n), (p) show the same via the additive Schwarz method.}
    \label{timedependV}
\end{figure}

\textbf{Experimental Setup.} 
{We randomly generate 676 interior sample points for \mathfix{\(m\) and \(u\)} in \(\mathbb{T}\times(0,T)\), together with \mathfix{20 initial-time collocation points for \(m(x,0)=m_0(x)\) and 20 terminal-time collocation points for \(u(x,T)=U_T(x)\)}. From these sample points, 75 observation points for \(m\) are randomly selected, and their observed values are obtained by interpolating from a finer reference solution (computed on a uniform $40 \times 40$ grid) using cubic interpolation. Meanwhile, 9 observation points for \(V\) are generated in the spatial dimension. The Gaussian regularization coefficients are set to \( \alpha_v = 13 \),  \( \alpha_{u} = 0.5\), \(\alpha_{m} = 0.5\), \( \alpha_{m^o} = \mathfix{10^5} \), and \( \alpha_{v^o} = \mathfix{10^5} \).  Gaussian noise with a standard deviation \(\gamma = 10^{-3}\), modeled as \(\mathcal{N}(0,\gamma^2 I)\), is added to the observations. \mathfix{The initial guesses for $m,u,\boldsymbol{Q},V$ are all set to 1.} We choose the kernel  }
\(K\left((x, t),(x^{\prime}, t^{\prime}) ; \boldsymbol{\sigma}\right)=\exp\left(\sigma_1^{-2}\left(\cos(2\pi(x-x^{\prime}))-1\right)\right)\exp \left(-\sigma_2^{-2}(t-t^{\prime})^2\right)\) for GPs of $m$, $u$, and $\boldsymbol{Q}$, while choosing the kernel \(K\left(\mathfix{x, x^{\prime}} ; \bar\sigma\right)=\exp\left(\bar{\sigma}^{-2}\left(\cos(2\pi(x-x^{\prime}))-1\right)\right)\) for the GP of $V$.

\textbf{Experimental Results.} 
{Figure \ref{Timedependentinversefig4} presents the collocation  points and boundary points for both \(m\) and \(u\), together with the observation points for \(m\).} Figure \ref{timedependV} illustrates the discretized \(L^2\) errors \(\mathcal{E}_2(m^{(k)}, m^*)\), as specified in \eqref{eq:l2disc}. 
{These errors quantify the discrepancies between the approximated solution \(m^{(k)}\) at the \(k\)-th iteration and the reference solution \(m^*\). The figure includes the reference solutions, recovered results, and the {absolute }pointwise error contours for the approximated functions of \(m\), \(u\), and \(V\). The running times are \(16.038\) seconds for the GPPI method and \(8.296\) seconds for the GPPI-AS method. The GPPI error stabilizes after 38 iterations, whereas the GPPI-AS error stabilizes after 5 iterations. }

Tables~\ref{timedependentobserve} and \ref{timedependentnoise} collect the
{relative $L^1$, $L^2$, and $L^\infty$ recovery errors of $V$ for the time-dependent MFG
\eqref{eqtimedepend}. 
In Table~\ref{timedependentobserve}, we fix the observation number of $m$ and the Gaussian noise as in the experimental setup, and gradually increase the observation number of $V$. The reconstruction error decreases as more observations of $V$ are added.
In \mathfix{Table}~\ref{timedependentnoise}, relative Gaussian noise following the model \eqref{eq:relnoise} is added to the observations of $m$ and $V$. Using more observation points improves the reconstruction, while higher noise levels $\tilde{\gamma}$ worsen the accuracy.}

\begin{table}[h]
\centering
\caption{Relative recovery error of $V$ under different numbers of observed $V$ points for the time-dependent MFG in \eqref{eqtimedepend}.}

\begin{tabular}{lccccc}
\toprule
Number of observed $V$ points & 5 & 6 & 7 & 9 & 11 \\
\midrule
Relative $L^1$ error      & 2.110e-01 & 9.041e-02 & 2.831e-02 & 3.372e-03 & 2.266e-03 \\
Relative $L^2$ error      & 2.392e-01 & 1.063e-01 & 3.222e-02 & 4.074e-03 & 2.403e-03 \\
Relative $L^\infty$ error & 3.300e-01 & 1.597e-01 & 4.589e-02 & 6.508e-03 & 3.492e-03  \\
\bottomrule
\end{tabular}
\label{timedependentobserve}
\end{table}
\begin{table}[H]
\centering
\caption{Relative recovery error of $V$ under different noise levels $\tilde{\gamma}$ for the time-dependent MFG in \eqref{eqtimedepend}.}

\begin{tabular}{lccccc}
\toprule
Noise level $\tilde{\gamma}$ & 0e+00 & 1e-03 & 5e-03 & 1e-02 & 5e-02 \\
\midrule
Relative $L^1$ error      & 3.210e-03 & 3.478e-03 & 5.324e-03 & 8.412e-03 & 3.570e-02 \\
Relative $L^2$ error      & 4.175e-03 & 4.400e-03 & 6.194e-03 & 9.356e-03 & 3.869e-02 \\
Relative $L^\infty$ error & 6.702e-03 & 7.355e-03 & 1.036e-02 & 1.411e-02 & 5.657e-02 \\
\bottomrule
\end{tabular}
\label{timedependentnoise}
\end{table}

\section{Conclusion and Future Work}
\label{sec:Conclusion}

In this paper, we present mesh-free GPPI frameworks to address both forward and inverse problems associated with HJB and MFG equations. Additionally, we integrated the additive Schwarz Newton method into our GPPI frameworks to further accelerate computational performance. The numerical experiments validate the effectiveness and efficiency of our proposed methods. \mathfix{The results highlight several features of the proposed methods.  Across the experiments, the additive Schwarz correction consistently reduces the number of outer policy-iteration steps while maintaining comparable accuracy.  In the inverse examples, the GP formulation also allows the collocation and observation points to be chosen nonuniformly, which is useful when the data are not tied to a prescribed computational grid.  As expected, the reconstruction errors increase as the observation noise grows and decrease as the number of observations increases.  }

Several other directions may further broaden the scope of the proposed framework.
One practical avenue is to incorporate scalable GP computational techniques, such as random Fourier features \cite{mou2022numerical}, sparse GPs \cite{meng2023sparse}, and mini-batch optimization methods \cite{yang2023mini}, to better accommodate large-scale datasets.
Another avenue is to apply GPPI to real-world problems in economics, finance, and biology, especially in settings where standard MFG models are not yet well established.

It is also natural to integrate UQ directly into our framework.
While classical PI alternates policy updates with solving HJB and FP equations via linear solvers, our GPPI imposes Gaussian priors on each unknown and yields posterior means conditioned on linear PDE constraints at collocation points.
The resulting Gaussian posteriors can support resampling strategies, experimental design, and error estimation, and motivate an adaptive-sampling GPPI variant with principled UQ.

Another extension concerns inverse MFGs under trajectory-based observations, where the density is not directly observed and only agent samples $\{X_i(t_\ell)\}_{i=1}^N$ are available at discrete times.
This setting has been considered in \cite{huang2025joint}, and one can incorporate it into our GPPI context by forming the empirical measure $\widehat{m}_N(t_\ell)=\frac{1}{N}\sum_{i=1}^N\delta_{X_i(t_\ell)}$ (or a smoothed density estimator) from the observed trajectories.
Accordingly, the data fidelity term is no longer defined via a linear observation operator acting on $m$, but instead is formulated through a distributional discrepancy, such as the kernel Maximum Mean Discrepancy (MMD) or the Wasserstein distance, to measure the mismatch between $m(t_\ell)$ and the empirical measure $\widehat{m}_N(t_\ell)$.
We leave a full implementation and a rigorous error analysis of this trajectory-based extension for future work.

\section*{Acknowledgments}
XY acknowledges support from the Air Force Office of Scientific Research under MURI awards FA9550-20-1-0358 (Machine Learning and Physics-Based Modeling and Simulation) and FOA-AFRL-AFOSR-2023-0004 (Mathematics of Digital Twins); the Department of Energy under award DE-SC0023163 (SEA-CROGS: Scalable, Efficient, and Accelerated Causal Reasoning Operators, Graphs and Spikes for Earth and Embedded Systems); the National Science Foundation under award 2425909 (Discovering the Law of Stress Transfer and Earthquake Dynamics in a Fault Network using a Computational Graph Discovery Approach); and the Vannevar Bush Faculty Fellowship under ONR award N000142512035. JZ acknowledges support from the National University of Singapore Risk Management Institute research scholarship and the IoTeX Foundation Industry Grant A-8001180-00-00.
\appendix

\section{Derivation Details for the GPPI Method}
\label{AppendixA}
\mathfix{In this section, we describe how the subproblems arising at each step of the GPPI framework are formulated and solved for HJB and MFG problems.} \Cref{Appendix5} provides the details for the HJB equation, \Cref{Appendix3} focuses on the stationary MFG case, and \Cref{Appendix4} addresses the time-dependent MFG problem.

\subsection{The HJB Equation Problem}
\label{Appendix5}
In this subsection, we derive the details for solving the inverse problem associated with HJB equations, as described in Section~\ref{HJB Equation Inverse}.

Assume that the value function \( U \) of a stochastic optimal control problem satisfies the following HJB equation:
\begin{equation}
\begin{cases}
\begin{aligned}
-\partial_t U(x, t)- \frac{1}{2} \sigma(t)^2 \Delta U(x, t) + \sup_{\boldsymbol{q} \in \mathcal{Q}} \Big\{-\nabla U(x, t)^T f(x, t, \boldsymbol{q}) -\ell(x, t, \boldsymbol{q})\Big\} = 0,
\end{aligned} & \forall  (x, t)\in \mathbb{T}^d \times(0, T), \\
U(x, T) = U_T(x), & \forall x\in \mathbb{T}^d,
\end{cases}
\label{HJBtimedependappendix}
\end{equation}
where \( f \) is the drift term, \( \ell \) is the running cost, and \( \sigma \) is the diffusion coefficient. We seek to solve Problem~\ref{st_probHJB}.
Based on this formulation, the GPPI algorithm can be structured into the following steps. 

\textbf{Step 1}. We first deal with the HJB equation. We use GPs to approximate the unknown value function $U$ and the unknown {cost field} $V$ with partial observations $\boldsymbol{U}^o$ and $\boldsymbol{V}^o$.  We select \(M\) collocation points \(\{(x_i,t_i)\}_{i=1}^M \subset \mathbb{T}^d \times (0,T]\), where the first \(M_\Omega\) points lie in the interior \(\mathbb{T}^d \times (0,T)\) and the remaining \(M - M_\Omega\) lie on the terminal slice \(\mathbb{T}^d \times \{T\}\).

Let \(\mathcal{U}\) and \(\mathcal{V}\) denote the RKHSs associated with the kernels \(K_u\) and \(K_v\), respectively. We assume that \(U \in \mathcal{U}\) and \(V \in \mathcal{V}\).
\mathfix{Given the current policy \(\boldsymbol{q}^{(k)}\), we write the inverse HJB subproblem as the following joint minimization over \((U,V)\):
\begin{align}
\begin{cases}
\inf_{(U,V) \in \mathcal{U} \times \mathcal{V}} & \alpha_u\|U\|_{\mathcal{U}}^2 +\alpha_v \|V\|_{\mathcal{V}}^2+ \alpha_{v^o}| [\Psi, V] - \boldsymbol{V}^o|^2+\alpha_{u^o}|\left[\boldsymbol{\phi}^o, U\right]-\boldsymbol{U}^{o}|^2 \\
\text{s.t.} & \mathfix{-\partial_t U(x_i,t_i)- \frac{1}{2} \sigma(t_i)^2 \Delta U(x_i, t_i)- \nabla U(x_i, t_i)\cdot f(x_i, t_i, \boldsymbol{q}^{(k)}(x_i, t_i))} \\
&\quad\quad\quad\quad\quad\quad\mathfix{-  V(x_i, t_i)-G(t_i, \boldsymbol{q}^{(k)}(x_i, t_i))  = 0}, \quad \forall i = 1,\dots, M_\Omega, \\
& U(x_i, T) = U_T(x_i), \quad \forall i=M_\Omega+1, \dots, M.  
\end{cases}
\label{OptHJBt1}
\end{align}
In the forward case, the same formulation is read with \(V\) frozen at the prescribed cost, so the minimization is only over \(U\).}
{To solve \eqref{OptHJBt1}, we leverage the idea in \cite{chen2021solving} and introduce latent variables, $\boldsymbol{z}$ and $\boldsymbol{v}$. For \((x,t)\in \mathbb{T}^d\times[0,T]\), let \(\delta_{(x,t)}\) denote the point-evaluation functional, i.e., \([\delta_{(x,t)},f]=f(x,t)\). Let \(\boldsymbol{\delta}^\Omega = (\delta_{(x_1,t_1)}, \dots, \delta_{(x_{M_\Omega},t_{M_\Omega})})\) and  \(\boldsymbol{\delta}^{\partial\Omega} = (\delta_{(x_{M_\Omega+1},t_{M_\Omega+1})}, \dots, \delta_{(x_M,t_M)})\). For brevity, we denote \(\boldsymbol{q}_i^{(k)} := \boldsymbol{q}^{(k)}(x_i, t_i)\) for \(i = 1, \dots, M_\Omega\). Define \(\boldsymbol{\delta} := (\boldsymbol{\delta}^{\Omega}, \boldsymbol{\delta}^{\partial\Omega})\).}
Thus, we rewrite \eqref{OptHJBt1} as
\begin{equation}
    \begin{aligned}
        & \begin{cases} 
            \inf _{\boldsymbol{z},\boldsymbol{v}} \begin{cases}
                \inf _{(U,V) \in \mathcal{U} \times \mathcal{V}}\alpha_u\|U\|_{\mathcal{U}}^2+ \alpha_{v}\|V\|_{\mathcal{V}}^2\\
                \text{s.t.} \quad \left[\boldsymbol{\delta}, U\right]=\boldsymbol{z}^{(1)}, 
                            [\boldsymbol{\delta}^{\Omega} \circ \partial_{t}, U]=\boldsymbol{z}^{(2)}, 
                            [\boldsymbol{\delta}^{\Omega} \circ \nabla , U]=\boldsymbol{z}^{(3)},\\  \quad\quad[\boldsymbol{\delta}^{\Omega} \circ \Delta, U]=\boldsymbol{z}^{(4)},\left[\boldsymbol{\phi}^o, U\right]=\boldsymbol{z}^{(5)},
     [\boldsymbol{\delta}^{\Omega}, V]=\boldsymbol{v}^{(1)},[\Psi, V]=\boldsymbol{v}^{(2)},
            \end{cases} \\
   \quad\quad\quad+\alpha_{u^o}|\boldsymbol{z}^{(5)}-\boldsymbol{U}^{o}|^2+\alpha_{v^o}|\boldsymbol{v}^{(2)}-\boldsymbol{V}^{o}|^2 \\
            \text{s.t.} \quad \mathfix{-z_i^{(2)}- \frac{1}{2} \sigma(t_i )^2 z_i^{(4)} - \boldsymbol{z}_i^{(3)}\cdot f( x_i,t_i, \boldsymbol{q}_i^{(k)}) - v_i^{(1)} - G(t_i, \boldsymbol{q}_i^{(k)})  = 0}
, \quad \forall i=1, \ldots, M_{\Omega}, \\
            \quad\quad z^{(1)}_i = U_T(x_i), \quad \forall i=M_{\Omega}+1, \ldots, M. 
          \end{cases}
    \end{aligned}
    \label{HJBoptimize}
\end{equation}
In this context, $\boldsymbol{z} = (z^{(1)}_1, \dots, z^{(1)}_M, z^{(2)}_1, \dots, z^{(2)}_{M_\Omega}, \boldsymbol{z}^{(3)}_1, \dots, \boldsymbol{z}^{(3)}_{M_\Omega}, z^{(4)}_1, \dots, z^{(4)}_{M_\Omega}, z^{(5)}_1, \dots, z^{(5)}_{N_u})$, and $\boldsymbol{v} = (v^{(1)}_1,$\\$ \dots, v^{(1)}_{M_\Omega}, v^{(2)}_1, \dots, v^{(2)}_{N_v})$, where $N_u$ and $N_v$ denote the numbers of observations on $U$ and $V$, respectively, as defined in Problem \ref{st_probHJB}.   {Let $\boldsymbol{\delta}^{V,\Omega}:=\boldsymbol{\delta}^{\Omega}$ denote the collocation evaluations acting on $V$.} Denote $\boldsymbol{\phi}^u := \left(\boldsymbol{\delta}, \boldsymbol{\delta}^{\Omega} \circ \partial_t, \boldsymbol{\delta}^{\Omega} \circ \nabla,\boldsymbol{\delta}^{\Omega} \circ \Delta, \boldsymbol{\phi}^o\right)$ and $\boldsymbol{\phi}^{V} :=(\boldsymbol{\delta}^{V,\Omega}, \Psi)$. {In this appendix, \(V\) is treated in the general time-dependent form \(V(x,t)\), so \(\boldsymbol{\delta}^{V,\Omega}\) consists of space-time evaluation functionals and \(K_v\) is a kernel on \(\mathbb{T}^d\times[0,T]\). When one restricts to the special case \(V(x,t)\equiv V(x)\), the \(V\)-block should instead be assembled from the corresponding distinct spatial evaluations.} \mathfix{The first-level optimization problem admits an explicit representer solution \((U^\dagger, V^\dagger)\),} such that 
\begin{align*}
		U^\dagger(x,t)= K_u((x,t),\boldsymbol{\phi}^u) K_u(\boldsymbol{\phi}^u, \boldsymbol{\phi}^u)^{-1}\boldsymbol{z} \quad \text { and }\quad
  V^\dagger(x,t)= K_v((x,t),\boldsymbol{\phi}^V) K_v(\boldsymbol{\phi}^V, \boldsymbol{\phi}^V)^{-1}\boldsymbol{v}.
\end{align*}
Thus, 
\begin{align*}
\|U^\dagger\|_{\mathcal{U}}^2=\boldsymbol{z}^TK_u(\boldsymbol{\phi}^u, \boldsymbol{\phi}^u)^{-1}\boldsymbol{z}\quad \text{ and }\quad  \|V^\dagger\|_{\mathcal{V}}^2=\boldsymbol{v}^TK_v(\boldsymbol{\phi}^V, \boldsymbol{\phi}^V)^{-1}\boldsymbol{v}. 
\end{align*}
Hence, we can formulate \eqref{HJBoptimize} as a finite-dimensional optimization problem
\begin{align}
\label{HJBoptimize finite}
\begin{cases}
\inf _{\boldsymbol{z},\boldsymbol{v}} \alpha_u\boldsymbol{z}^T K_u\left(\boldsymbol{\phi}^u, \boldsymbol{\phi}^u\right)^{-1} \boldsymbol{z}+\alpha_v\boldsymbol{v}^T K_v(\boldsymbol{\phi}^{V}, \boldsymbol{\phi}^{V})^{-1} \boldsymbol{v}+\alpha_{v^o}|\boldsymbol{v}^{(2)}-\boldsymbol{V}^{o}|^2+\alpha_{u^o}|\boldsymbol{z}^{(5)}-\boldsymbol{U}^{o}|^2\\
 \text { s.t. } \quad \mathfix{-z^{(2)}_i- \frac{1}{2} \sigma(t_i )^2 z_i^{(4)}-\boldsymbol{z}^{(3)}_i \cdot f( x_i,t_i, \boldsymbol{q}_i^{(k)})-v_i^{(1)}-G(t_i,\boldsymbol{q}_i^{(k)})=0}, \quad \forall i=1, \ldots, M_{\Omega},\\
 \quad\quad\quad z^{(1)}_i = U_T(x_i), \quad \forall i=M_{\Omega}+1, \ldots, M.
\end{cases}
\end{align}
\mathfix{The problem in \eqref{HJBoptimize finite} is a convex quadratic optimization problem with linear equality constraints. If the feasible set is nonempty and the objective is strictly convex on it, the minimizer exists and 
  is unique. The Lagrange multiplier method reduces the computation to a linear system, so no iterative optimization algorithm is needed. We omit the algebraic details.}

\textbf{Step 2.} Let $\chi = \{(x_1, t_1), \dots, (x_{M_\Omega}, t_{M_\Omega})\}$ be the collection of collocation points on $\mathbb{T}^d \times (0, T)$. For the second step, we update the policy at the collocation points as follows
\begin{align*}
\boldsymbol{\mathfrak{q}}^{k+1, i} :=\underset{\boldsymbol{q} \in \mathcal{Q}}{\arg \max }\{-\boldsymbol{z}^{(3)}_i \cdot f(x_i, t_i,  \boldsymbol{q})-\ell(x_i,t_i, \boldsymbol{q})\}, \quad \forall i = 1,\dots, M_\Omega.
\end{align*}
We concatenate the vectors \(\boldsymbol{\mathfrak{q}}^{k+1, i}\) into a single vector \(\boldsymbol{\mathfrak{q}}\), as defined in \eqref{eq:defqvec}. The resulting optimal policy is then approximated using a GP regression model:
\begin{align*}
\boldsymbol{q}^{(k+1)}(x, t) = \boldsymbol{K}_{\boldsymbol{q}}((x, t), \chi) \boldsymbol{K}_{\boldsymbol{q}}(\chi, \chi)^{-1} \boldsymbol{\mathfrak{q}}^{k+1}, \quad \forall (x, t)\in \mathbb{T}^d \times(0, T).
\end{align*}
{
Admissibility of the policy is imposed only at the collocation points, so the GP reconstruction is not automatically guaranteed to satisfy \(\boldsymbol q(x,t)\in\mathcal Q\) throughout the domain. In this work, we restrict attention to cases where the reconstructed policy remains admissible in the numerical experiments. A possible alternative is to project the GP mean pointwise onto \(\mathcal Q\), which we do not pursue here.
}

The above procedure is repeated iteratively until a stopping criterion is met.

\subsection{The Stationary MFG Problem}
\label{Appendix3}
In this subsection, we present the details of the GPPI method for solving the inverse problem corresponding to the following stationary  MFG introduced in \Cref{MFG Forward Problem Stationary}:
\begin{equation}
\begin{cases}
-\nu\Delta u + H(x, \nabla u) + \lambda = F(m)+V(x), & \quad \forall x\in \mathbb{T}^d, \\ 
-\nu \Delta m - \operatorname{div}\left( D_p H(x, \nabla u) \, m \right) = 0, & \quad \forall x\in \mathbb{T}^d, \\
\int_{\mathbb{T}^d} u \, \dif x = 0, \quad \int_{\mathbb{T}^d} m \, \dif x = 1,
\end{cases}
\label{eqonedimension1forwardappendix}
\end{equation}
{
where \( u \) denotes the value function, \( m \) the population distribution, \( V \) the spatial cost function, and \( \lambda \) is a constant. Let \(\{x_i\}_{i=1}^M\) denote the collocation points on \(\mathbb{T}^d\). The GPPI algorithm is detailed below.
}

\textbf{Step 1.} We begin by solving the FP equation within the GP framework, assuming that the solution \(m\) belongs to the RKHS \(\mathcal{M}\) associated with the kernel \(K_m\). We solve 
\begin{align}
\begin{cases}
\displaystyle \inf_{m \in \mathcal{M}} & \alpha_m\|m\|_{\mathcal{M}}^2+ \alpha_{m^o} \left|[\boldsymbol{\phi}^o, m] - \boldsymbol{m}^o\right|^2, \\
\text{s.t.} & - \nu \Delta m(x_i) - \operatorname{div}(m \boldsymbol{Q}^{(k)})(x_i) = 0, \quad\forall i = 1, \dots, M, \\
& \int_{\mathbb{T}^d} m(x)\,\dif x = 1.
\end{cases}
\label{OptGPtdProb_CptVo}
\end{align}
{
Let \(\boldsymbol{\delta} = (\delta_{x_i})_{i=1}^M\) denote the vector of Dirac measures at the collocation points, and let \(\boldsymbol{\phi}^o\) denote the observation operator vector as defined in Problem~\ref{st_prob}. We also introduce the integral functional \(\eta_m \in \mathcal{M}^*\) defined by
\[
[\eta_m,m] := \int_{\mathbb{T}^d} m(x)\,\dif x.
\]
To solve \eqref{OptGPtdProb_CptVo}, we introduce latent variables \( \boldsymbol{\rho} \) and reformulate \eqref{OptGPtdProb_CptVo} into 
}
\begin{equation}
    \begin{aligned}
& \begin{cases} \inf _{\boldsymbol{\rho}} \begin{cases}\inf _{m \in\mathcal{M}}\alpha_m\|m\|_{\mathcal{M}}^2 \\
\text { s.t. } {\left[\boldsymbol{\delta}, m\right]=\boldsymbol{\rho}^{(1)},[\boldsymbol{\delta} \circ \nabla, m]=\boldsymbol{\rho}^{(2)},[\boldsymbol{\delta} \circ \Delta, m]=\boldsymbol{\rho}^{(3)},\left[\boldsymbol{\phi}^o, m\right]=\boldsymbol{\rho}^{(4)},[\eta_m,m]=\rho^{(5)}},\end{cases} \\
\quad\quad+\alpha_{m^o}|\boldsymbol{\rho}^{(4)}-\boldsymbol{m}^o|^2\\
 \text { s.t. } 
 \rho^{(3)}_i=-\frac{1}{\nu }(\boldsymbol{\rho}^{(2)}_i \cdot \boldsymbol{Q}^{(k)}(x_i)+\rho^{(1)}_i\div(\boldsymbol{Q}^{(k)})(x_i)) , \quad \forall i=1, \ldots, M,\\
\quad\quad \rho^{(5)}=1.\end{cases}
\end{aligned}
\label{OptGPProb_CptV}
\end{equation}
{
Here, \( \boldsymbol{\rho} = (\boldsymbol{\rho}^{(1)}, \boldsymbol{\rho}^{(2)}, \boldsymbol{\rho}^{(3)}, \boldsymbol{\rho}^{(4)}, \rho^{(5)}) \) is the collection of latent variables. Denote \( \boldsymbol{\phi}^m = (\boldsymbol{\delta}, \boldsymbol{\delta} \circ \nabla, \boldsymbol{\delta} \circ \Delta, \boldsymbol{\phi}^o,\eta_m) \).} Let $m^\dagger$ be the solution to the first level minimization problem for $m$ in \eqref{OptGPProb_CptV}. Given $\boldsymbol{\rho}$, we get
\[
m^{\dagger}(x)=K_m\left(x, \boldsymbol{\phi}^m\right) K_m\left(\boldsymbol{\phi}^m, \boldsymbol{\phi}^m\right)^{-1} \boldsymbol{\rho}\quad\text{ and }\quad
\|m^{\dagger}\|_{\mathcal{M}}^2=\boldsymbol{\rho}^T K_m\left(\boldsymbol{\phi}^m, \boldsymbol{\phi}^m\right)^{-1} \boldsymbol{\rho}.
\]
Hence, we can formulate \eqref{OptGPProb_CptV} as a finite-dimensional optimization problem
\begin{align}
\label{OptGPProb_rhoV}
\begin{cases}
\inf_{\boldsymbol{\rho}}  \alpha_m\boldsymbol{\rho}^T K_m\left(\boldsymbol{\phi}^m, \boldsymbol{\phi}^m\right)^{-1} \boldsymbol{\rho} +\alpha_{m^o}|\boldsymbol{\rho}^{(4)}- \boldsymbol{m}^o|^2
\\
 \text { s.t. } \rho^{(3)}_i=-\frac{1}{\nu }(\boldsymbol{\rho}^{(2)}_i \cdot \boldsymbol{Q}^{(k)}(x_i)+\rho^{(1)}_i\div(\boldsymbol{Q}^{(k)})(x_i)) , \quad \forall i=1, \ldots, M,
 \\
{
\quad\quad \rho^{(5)}=1.
}
\end{cases}
\end{align}
{The problem in \eqref{OptGPProb_rhoV} is a convex quadratic optimization problem with linear equality constraints. If the feasible set is nonempty and the objective is strictly convex on it, the minimizer exists and is unique. The Lagrange multiplier method then reduces the computation to solving a linear system.}

\textbf{Step 2}. In this step, we deal with the HJB equation.
Within the GP framework, we approximate the value function \(u\), the spatial cost \(V\), and the constant \(\lambda\), leading to the following optimization problem:
\begin{align}
\begin{cases}
\inf _{(u,\lambda, V) \in\mathcal{U}\times \mathbb{R}\times \mathcal{V}}&\alpha_{u}\|u\|_{\mathcal{U}}^2+\alpha_{\lambda}|\lambda|^2+\alpha_v\|V\|_{\mathcal{V}}^2+ \alpha_{v^o}| [\Psi, V] - \boldsymbol{V}^o|^2 \\
\text{s.t.} &  -\nu \Delta u(x_i) + \boldsymbol{Q}^{(k)}(x_i) \cdot \nabla u(x_i) + \lambda\\
& \quad\quad\quad\quad= L(x_i, \boldsymbol{Q}^{(k)}(x_i)) + V(x_i) + F(m^{(k)}(x_i)), \quad \forall i = 1, \ldots, M,  \\
& 
{
\int_{\mathbb{T}^d} u(x)\,\dif x = 0.
}
\end{cases}
\label{eq:mfg_systeU}
\end{align}
{
We introduce the integral functional \(\eta_u \in \mathcal{U}^*\) defined by
\[
[\eta_u,u] := \int_{\mathbb{T}^d} u(x)\,\dif x,
\]
and the latent variables $\boldsymbol{z}$, $\lambda$, and $\boldsymbol{v}$. Then \eqref{eq:mfg_systeU} becomes
}
\begin{equation}
    \begin{aligned}
& \begin{cases} \inf _{\boldsymbol{z},\lambda,\boldsymbol{v}} \begin{cases}\inf _{(u,V) \in\mathcal{U}\times \mathcal{V}}\alpha_u\|u\|_{\mathcal{U}}^2+\alpha_v\|V\|_{\mathcal{V}}^2 \\
\begin{aligned}\text { s.t. }& 
    
    {\left[\boldsymbol{\delta}, u\right]=\boldsymbol{z}^{(1)},[\boldsymbol{\delta} \circ \nabla, u]=\boldsymbol{z}^{(2)},[\boldsymbol{\delta} \circ \Delta, u]=\boldsymbol{z}^{(3)},[\eta_u,u]=z^{(4)},}\\&
    \left[\boldsymbol{\delta}, V\right]=\boldsymbol{v}^{(1)},[\Psi, V]=\boldsymbol{v}^{(2)},
    \end{aligned}
\end{cases} \\ \quad\quad\quad\quad+\alpha_{v^o}|\boldsymbol{v}^{(2)}-\boldsymbol{V}^o|^2+\alpha_{\lambda}|\lambda|^2\\
 \text { s.t. } -\nu z^{(3)}_i + \boldsymbol{Q}^{(k)}(x_i) \cdot\boldsymbol{z}^{(2)}_i + \lambda = L(x_i, \boldsymbol{Q}^{(k)}(x_i)) + v^{(1)}_i + F(m^{(k)}(x_i))
, \quad \forall i=1, \ldots, M, \\
\quad\quad z^{(4)}=0, \end{cases}
\end{aligned}
\label{StationaryVU}
\end{equation}
{
where $\boldsymbol{z}=(\boldsymbol{z}^{(1)},\boldsymbol{z}^{(2)},\boldsymbol{z}^{(3)},z^{(4)})$ and $\boldsymbol{v}=(\boldsymbol{v}^{(1)},\boldsymbol{v}^{(2)})$. Denote $
\boldsymbol{\phi}^u=(\boldsymbol{\delta},\boldsymbol{\delta} \circ \nabla, \boldsymbol{\delta} \circ \Delta,\eta_u)$ and $
 \boldsymbol{\phi}^{V} :=(\boldsymbol{\delta},\Psi)$.
}
Let $K_u$ and $K_v$ be kernels associated with the RKHSs $\mathcal{U}$ and $\mathcal{V}$, respectively. Moreover, let $(u^\dagger, V^\dagger)$ be the solution to the first level minimization problem for $u$ and $V$ in \eqref{StationaryVU} given $\boldsymbol{z}$ and $\boldsymbol{v}$. Then,
\[
u^\dagger(x) = K_u(x, \boldsymbol{\phi}^u) K_u(\boldsymbol{\phi}^u, \boldsymbol{\phi}^u)^{-1} \boldsymbol{z}\quad \text{ and } \quad
V^\dagger(x) = K_v(x, \boldsymbol{\phi}^V) K_v(\boldsymbol{\phi}^V, \boldsymbol{\phi}^V)^{-1} \boldsymbol{v}.
\]
Consequently, 
\[
\|u^\dagger\|_{\mathcal{U}}^2 = \boldsymbol{z}^T K_u(\boldsymbol{\phi}^u, \boldsymbol{\phi}^u)^{-1} \boldsymbol{z} \quad\text{ and }\quad
\|V^\dagger\|_{\mathcal{V}}^2 = \boldsymbol{v}^T K_v(\boldsymbol{\phi}^V, \boldsymbol{\phi}^V)^{-1} \boldsymbol{v}.
\]
Hence, \eqref{StationaryVU} can be formulated as a finite-dimensional, linearly constrained quadratic optimization problem: 
\begin{align}
\begin{cases}
\inf _{\boldsymbol{z},\lambda,\boldsymbol{v}} \alpha_u\boldsymbol{z}^T K_u\left(\boldsymbol{\phi}^u, \boldsymbol{\phi}^u\right)^{-1} \boldsymbol{z}+\alpha_{\lambda}|\lambda|^2+\alpha_v\boldsymbol{v}^T K_v(\boldsymbol{\phi}^{V}, \boldsymbol{\phi}^{V})^{-1} \boldsymbol{v}+\alpha_{v^o}|\boldsymbol{v}^{(2)}-\boldsymbol{V}^{o}|^2 \\
 \text { s.t. } z_i^{(3)}=\frac{1}{\nu}(\lambda+\boldsymbol{Q}^{(k)}(x_i)\cdot \boldsymbol{z}^{(2)}_i-(L(x_i, \boldsymbol{Q}^{(k)}(x_i))+v_i^{(1)}+F(m^{(k)}(x_i)))),\quad \forall i=1,\ldots, M,
 \\
{
\quad\quad z^{(4)}=0.
}
\end{cases}
\label{StationaryVappendix}
\end{align}
{
Thus, we solve \eqref{StationaryVappendix} using the method of Lagrange multipliers. The zero-mean normalization of \(u\) is handled by the continuous linear functional \(\eta_u\), in exact analogy with the mass constraint in Step 1.
}

\textbf{Step 3.} In the third step, we update the policy at the collocation points by solving
\[
\mathbf{q}^{k+1,i} = \arg \max_{\|\boldsymbol{q}\| \leq \mathfix{R_{\mathrm{pol}}}} \left\{ \boldsymbol{q}  \cdot \boldsymbol{z}^{(2)}_i - L(x_i,\boldsymbol{q}) \right\}, \quad \forall i = 1, \ldots, M,
\]
where \( \boldsymbol{z}^{(2)}_i \) is the evaluated gradient of the value function at the point \( x_i \). Therefore, by modeling the updated policy \(\boldsymbol{Q}^{(k+1)}\) as the mean of a GP conditioned on the observations of \(\boldsymbol{Q}\) at the collocation points, we obtain
\begin{equation*}
\boldsymbol{Q}^{(k+1)}(x) = \boldsymbol{K}_{\boldsymbol{Q}}(x, X)\,
\boldsymbol{K}_{\boldsymbol{Q}}(X, X)^{-1}\,
\mathbf{q}^{k+1}, \quad x \in \mathbb{T}^d,
\end{equation*}
where \( \mathbf{q}^{k+1} \) denotes the vector of optimal control values at the sampled locations. 

The above three steps are iteratively repeated until a stopping criterion is met.

\subsection{\texorpdfstring{\mathfix{The Time-Dependent MFG Problem}}{The Time-Dependent MFG Problem}}
\label{Appendix4}
\mathfix{In this subsection, we give the details for solving the inverse problem associated with the following time-dependent MFG system, following Section~\ref{MFG Forward Problem Time Dependent}:}
\begin{equation}
\begin{cases}
-\partial_t u-\nu \Delta u+H(x,t, \nabla u) =F(m)+ V(x,t), & \forall (x, t)\in \mathbb{T}^d \times(0, T), \\
\partial_t m-\nu \Delta m-\operatorname{div}\left(m D_p H(x,t, \nabla u)\right)=0, & \forall (x, t)\in \mathbb{T}^d \times(0, T), \\
m(x, 0)=m_0(x), \quad u(x, T)=U_T(x), & \forall x\in \mathbb{T}^d.
\end{cases}
\label{eq:MFG_Systemtappendix}
\end{equation}
We can solve the problem using the following steps.

{
\textbf{Step 1.} In the first step, we solve the FP equation. We select \mathfix{\(M^{\mathrm{FP}}\)} collocation points \(\{(x_i,t_i)\}_{i=1}^{\mathfix{M^{\mathrm{FP}}}} \subset \mathbb{T}^d \times [0,T)\), where the first \mathfix{\(M_{\Omega}^{\mathrm{FP}}\)} points are located in the interior \(\mathbb{T}^d \times (0,T)\), and the remaining \mathfix{\(M^{\mathrm{FP}} - M_{\Omega}^{\mathrm{FP}}\)} points lie on the initial slice \(\mathbb{T}^d \times \{0\}\).
}
Using a GP to approximate the density function \( m \), and incorporating observations \( \boldsymbol{m}^o \), the problem is formulated as:
\begin{align}
\begin{cases}
\displaystyle \inf_{m \in \mathcal{M}} & \alpha_m \|m\|_{\mathcal{M}}^2 + \alpha_{m^o} \left|[\boldsymbol{\phi}^o, m] - \boldsymbol{m}^o \right|^2, \\
\text{s.t.} & \partial_t m(x_i, t_i) - \nu \Delta m(x_i, t_i) - \operatorname{div}(m \boldsymbol{Q}^{(k)})(x_i, t_i) = 0, \quad \forall i = 1, \dots, \mathfix{M_{\Omega}^{\mathrm{FP}}}, \\
& m(x_i,0) = m_0(x_i), \quad \forall i = \mathfix{M_{\Omega}^{\mathrm{FP}}+1}, \dots, \mathfix{M^{\mathrm{FP}}}. 
\end{cases}
\label{OptGPtdProb_Cpt}
\end{align}
\mathfix{Let \(\boldsymbol{\delta}^{m, \Omega} = (\delta_{(x_1,t_1)}, \dots, \delta_{(x_{M_{\Omega}^{\mathrm{FP}}},t_{M_{\Omega}^{\mathrm{FP}}})})\) and \(\boldsymbol{\delta}^{m, \partial\Omega} = (\delta_{(x_{M_{\Omega}^{\mathrm{FP}}+1},t_{M_{\Omega}^{\mathrm{FP}}+1})}, \dots, \delta_{(x_{M^{\mathrm{FP}}},t_{M^{\mathrm{FP}}})})\). Denote \(\boldsymbol{\delta}^m := (\boldsymbol{\delta}^{m, \Omega}, \boldsymbol{\delta}^{m, \partial\Omega})\).}
{
For the HJB step below, when the FP and HJB interior collocation sets are not necessarily identical, we also introduce the evaluation functionals of the density at the interior HJB collocation points,
\[
\boldsymbol{\delta}^{m,\mathrm{HJB}}
:=
\left(\delta_{(x_j,t_j)}\right)_{j=1}^{M_{\Omega}^{\mathrm{HJB}}},
\]
where \(\{(x_j,t_j)\}_{j=1}^{M_{\Omega}^{\mathrm{HJB}}}\) denotes the interior HJB collocation set used in Step~2.
}
We rewrite   \eqref{OptGPtdProb_Cpt} as 
\begin{equation}
    \begin{aligned}
& \begin{cases} \inf _{\boldsymbol{\rho}} \begin{cases}\inf _{m \in\mathcal{M}}\alpha_m\|m\|_{\mathcal{M}}^2 \\
\text { s.t. } \left[\boldsymbol{\delta}^m, m\right]=\boldsymbol{\rho}^{(1)},[\boldsymbol{\delta}^{m, \Omega} \circ \partial_{t}, m]=\boldsymbol{\rho}^{(2)},[\boldsymbol{\delta}^{m, \Omega} \circ \nabla, m]=\boldsymbol{\rho}^{(3)}, \\ \quad\quad [\boldsymbol{\delta}^{m, \Omega} \circ \Delta, m]=\boldsymbol{\rho}^{(4)},[\boldsymbol{\phi}^o, m]=\boldsymbol{\rho}^{(5)}{,[\boldsymbol{\delta}^{m,\mathrm{HJB}},m]=\boldsymbol{\rho}^{(6)}},
\end{cases} \\
\quad\quad\quad+\alpha_{m^o}|\boldsymbol{\rho}^{(5)}-\boldsymbol{m}^o|^2 \\
 \text { s.t. } \rho_i^{(2)} - \nu \rho_i^{(4)} - \boldsymbol{\rho}_i^{(3)}\cdot\boldsymbol{Q}^{(k)}(x_i, t_i) - \rho_i^{(1)}\operatorname{div}( \boldsymbol{Q}^{(k)})(x_i, t_i)= 0, \quad \forall i=1, \ldots, \mathfix{M_{\Omega}^{\mathrm{FP}}},
 \\
 \quad\quad\rho_i^{(1)}=m_0(x_i),\quad \forall i=\mathfix{M_{\Omega}^{\mathrm{FP}}+1}, \ldots, \mathfix{M^{\mathrm{FP}}},
 \end{cases}
\end{aligned}
\label{OptGPProb_Cpttltd}
\end{equation}
where $\boldsymbol{\rho}=(\boldsymbol{\rho}^{(1)},\boldsymbol{\rho}^{(2)},\boldsymbol{\rho}^{(3)},\boldsymbol{\rho}^{(4)},\boldsymbol{\rho}^{(5)}{,\boldsymbol{\rho}^{(6)}}).$
Denote $
\boldsymbol{\phi}^m=\left(\boldsymbol{\delta}^m,\boldsymbol{\delta}^{m, \Omega} \circ \partial_{t}, \boldsymbol{\delta}^{m, \Omega} \circ \nabla, \boldsymbol{\delta}^{m, \Omega} \circ \Delta, \boldsymbol{\phi}^o{,\boldsymbol{\delta}^{m,\mathrm{HJB}}}\right)
$. Let $K_m$ be the kernel associated with the RKHS $\mathcal{M}$. Let $m^\dagger$ be the solution to the first-level minimization problem. Thus, $
m^{\dagger}(x,t)=K_m((x,t), \boldsymbol{\phi}^m) K_m(\boldsymbol{\phi}^m, \boldsymbol{\phi}^m)^{-1} \boldsymbol{\rho}.
$
Consequently, the RKHS norm of \( m^\dagger \) is given by
$
\left\|m^{\dagger}\right\|_{\mathcal{M}}^2=\boldsymbol{\rho}^T K_m(\boldsymbol{\phi}^m, \boldsymbol{\phi}^m)^{-1} \boldsymbol{\rho}.
$
{
The additional block \(\boldsymbol{\rho}^{(6)}\) stores the density values at the HJB interior collocation points. After solving the FP subproblem at iteration \(k\), we write
\[
\boldsymbol{m}_{\mathrm{HJB}}^{(k)}
:=
\left(\boldsymbol{\rho}^{(6)}\right)^{(k)}
=
\left(m^{(k)}(x_j,t_j)\right)_{j=1}^{M_{\Omega}^{\mathrm{HJB}}}.
\]
The notation \(m_{\mathrm{HJB},j}^{(k)}\) denotes the \(j\)-th component of \(\boldsymbol{m}_{\mathrm{HJB}}^{(k)}\).} Hence, \eqref{OptGPProb_Cpttltd} can be reformulated as the following finite-dimensional optimization problem:
\begin{align}
\label{OptGPProb_rhotV}
\begin{cases}
\inf_{\boldsymbol{\rho}}  \alpha_m\boldsymbol{\rho}^T K_m(\boldsymbol{\phi}^m, \boldsymbol{\phi}^m)^{-1} \boldsymbol{\rho} +\alpha_{m^o}|\boldsymbol{\rho}^{(5)}- \boldsymbol{m}^o|^2\\
 \text { s.t. } \rho_i^{(2)} = \nu \rho_i^{(4)} + \boldsymbol{\rho}_i^{(3)}\cdot\boldsymbol{Q}^{(k)}(x_i, t_i) + \rho_i^{(1)}\operatorname{div}( \boldsymbol{Q}^{(k)})(x_i, t_i) , \quad \forall i=1, \ldots, \mathfix{M_{\Omega}^{\mathrm{FP}}},
 \\
\quad\quad \rho_i^{(1)}=m_0(x_i),\quad \forall i=\mathfix{M_{\Omega}^{\mathrm{FP}}+1}, \ldots, \mathfix{M^{\mathrm{FP}}}.
\end{cases}
\end{align}
{
If the FP and HJB interior collocation sets coincide and are ordered identically, the block \(\boldsymbol{\rho}^{(6)}\) is identified with the corresponding interior point-evaluation part of \(\boldsymbol{\rho}^{(1)}\). In that special case, the duplicated HJB-evaluation block need not be assembled separately in the numerical implementation. Equivalently, \(\boldsymbol{m}_{\mathrm{HJB}}^{(k)}\) is set equal to that block of interior density-values.
}
\mathfix{The problem in \eqref{OptGPProb_rhotV} is a convex quadratic optimization problem with linear equality constraints. If the feasible set is nonempty and the objective is strictly convex on it, the minimizer exists and is unique. The minimizer is obtained by solving the corresponding finite-dimensional KKT linear system.} 

\textbf{Step 2}. For the second step, we solve the HJB equation.
We employ the GP framework to approximate the value function \( u \) and the {cost field} \( V \). We select \mathfix{\(M^{\mathrm{HJB}}\)} collocation points \(\{(x_j,t_j)\}_{j=1}^{\mathfix{M^{\mathrm{HJB}}}} \subset \mathbb{T}^d \times (0,T]\), with the first \mathfix{\(M_{\Omega}^{\mathrm{HJB}}\)} points lying in the interior \(\mathbb{T}^d \times (0,T)\) and the remaining \mathfix{\(M^{\mathrm{HJB}} - M_{\Omega}^{\mathrm{HJB}}\)} points on the terminal slice \(\mathbb{T}^d \times \{T\}\).
 We solve
\begin{align}
\begin{cases}
\inf_{(u, V) \in \mathcal{U} \times \mathcal{V}} & \alpha_u\|u\|_{\mathcal{U}}^2 +\alpha_v \|V\|_{\mathcal{V}}^2+ \alpha_{v^o}| [\Psi, V] - \boldsymbol{V}^o|^2 \\
\text{s.t.} & -\partial_t u(x_j, t_j) - \nu \Delta u(x_j, t_j) + \boldsymbol{Q}^{(k)}(x_j, t_j) \cdot \nabla u(x_j, t_j)\\
& \quad\quad\quad\quad= L(x_j, t_j, \boldsymbol{Q}^{(k)}(x_j, t_j)) + V(x_j, t_j) + {F\!\left(m_{\mathrm{HJB},j}^{(k)}\right)}, \quad \forall j = 1, \dots, \mathfix{M_{\Omega}^{\mathrm{HJB}}},  \\
& u(x_j, T) = U_T(x_j), \quad \forall j = \mathfix{M_{\Omega}^{\mathrm{HJB}}+1}, \dots, \mathfix{M^{\mathrm{HJB}}}. 
\end{cases}
\label{OptHJBtdProb_Cptu1d}
\end{align}
{
\mathfix{Let \(\boldsymbol{\delta}^{u, \Omega} = (\delta_{(x_1,t_1)}, \dots, \delta_{(x_{M_{\Omega}^{\mathrm{HJB}}},t_{M_{\Omega}^{\mathrm{HJB}}})})\) and  \(\boldsymbol{\delta}^{u, \partial\Omega} = (\delta_{(x_{M_{\Omega}^{\mathrm{HJB}}+1},t_{M_{\Omega}^{\mathrm{HJB}}+1})}, \dots, \delta_{(x_{M^{\mathrm{HJB}}},t_{M^{\mathrm{HJB}}})})\). Denote \(\boldsymbol{\delta}^u := (\boldsymbol{\delta}^{u, \Omega}, \boldsymbol{\delta}^{u, \partial\Omega})\). We reformulate problem~\eqref{OptHJBtdProb_Cptu1d} as the two-level optimization problem}
}
\begin{equation}
    \begin{aligned}
& \begin{cases} \inf _{\boldsymbol{z},\boldsymbol{v}} \begin{cases}\inf _{(u,V) \in\mathcal{U}\times \mathcal{V}}\alpha_u\|u\|_{\mathcal{U}}^2+\alpha_v\|V\|_{\mathcal{V}}^2 \\
\text { s.t. } [\boldsymbol{\delta}^u, u]=\boldsymbol{z}^{(1)},[\boldsymbol{\delta}^{u, \Omega} \circ \partial_{t}, u]=\boldsymbol{z}^{(2)},[\boldsymbol{\delta}^{u, \Omega} \circ \nabla, u]=\boldsymbol{z}^{(3)},  \\ \quad
\quad [\boldsymbol{\delta}^{u, \Omega} \circ \Delta, u]=\boldsymbol{z}^{(4)},
[\boldsymbol{\delta}^{u, \Omega}, V]=\boldsymbol{v}^{(1)},[\Psi, V]=\boldsymbol{v}^{(2)},\end{cases}  \\
\quad\quad\quad+\alpha_{v^o}|\boldsymbol{v}^{(2)}-\boldsymbol{V}^{o}|^2 \\
 \text { s.t. } -z^{(2)}_j-\nu z^{(4)}_j+\boldsymbol{Q}^{(k)}(x_j, t_j)\cdot\boldsymbol{z}^{(3)}_j\\\quad\quad\quad\quad=L(x_j, t_j, \boldsymbol{Q}^{(k)}(x_j, t_j))+v^{(1)}_j+{F\!\left(m_{\mathrm{HJB},j}^{(k)}\right)}, \quad \forall j=1, \ldots, \mathfix{M_{\Omega}^{\mathrm{HJB}}}, \\ \quad\quad z^{(1)}_j=U_T(x_j),\quad \forall j=\mathfix{M_{\Omega}^{\mathrm{HJB}}+1}, \ldots, \mathfix{M^{\mathrm{HJB}}}, \end{cases}
\end{aligned}
\label{OptGPProb_Cpttlu}
\end{equation}
{
where $\boldsymbol{z}=(\boldsymbol{z}^{(1)},\boldsymbol{z}^{(2)},\boldsymbol{z}^{(3)},\boldsymbol{z}^{(4)})$ and $\boldsymbol{v}=(\boldsymbol{v}^{(1)},\boldsymbol{v}^{(2)})$. Let \(\boldsymbol{\delta}^{V,\Omega}:=\boldsymbol{\delta}^{u,\Omega}\) denote the same collocation evaluations, now acting on \(V\). Denote
\[
\boldsymbol{\phi}^u := \left(\boldsymbol{\delta}^u,\boldsymbol{\delta}^{u, \Omega} \circ \partial_t,\boldsymbol{\delta}^{u, \Omega} \circ \nabla,\boldsymbol{\delta}^{u, \Omega} \circ \Delta\right),
\qquad
\boldsymbol{\phi}^{V} := \left(\boldsymbol{\delta}^{V,\Omega},\Psi\right).
\]
In the general derivation here, \(V\) is treated as a function of \((x,t)\), so \(\boldsymbol{\delta}^{V,\Omega}\) consists of space-time evaluations and \(K_v\) is a kernel on \(\mathbb{T}^d\times[0,T]\). In the numerical experiment of Section~\ref{Timedependentexperiment}, we specialize to the time-independent case \(V(x,t)\equiv V(x)\).} Let $K_u$ and $K_v$ be kernels associated with the RKHSs $\mathcal{U}$ and $\mathcal{V}$, respectively. Let $(u^\dagger, V^\dagger)$ be the solution to the first level minimization problem for $u$ and $V$ in \eqref{OptGPProb_Cpttlu} given $\boldsymbol{z}$ and $\boldsymbol{v}$. Then,
\[
\begin{aligned}
        u^\dagger(x,t) = K_u((x,t), \boldsymbol{\phi}^u) K_u(\boldsymbol{\phi}^u, \boldsymbol{\phi}^u)^{-1} \boldsymbol{z}\quad \text{ and }\quad
        V^\dagger(x,t) = K_v((x,t), \boldsymbol{\phi}^{V}) K_v(\boldsymbol{\phi}^{V}, \boldsymbol{\phi}^{V})^{-1} \boldsymbol{v}.
\end{aligned}
\]
Consequently, 
\begin{align*}
\|u^\dagger\|_{\mathcal{U}}^2 = \boldsymbol{z}^T K_u(\boldsymbol{\phi}^u, \boldsymbol{\phi}^u)^{-1} \boldsymbol{z} \quad\text{ and }\quad
        \|V^\dagger\|_{\mathcal{V}}^2 = \boldsymbol{v}^T K_v(\boldsymbol{\phi}^{V}, \boldsymbol{\phi}^{V})^{-1} \boldsymbol{v}.
\end{align*}
Substituting these expressions into the original objective, we reformulate~\eqref{OptGPProb_Cpttlu} as the following finite-dimensional optimization problem:
\begin{align}
\label{OptGPProb_ztv}
\begin{cases}
\inf _{\boldsymbol{z},\boldsymbol{v}} \alpha_u\boldsymbol{z}^T K_u\left(\boldsymbol{\phi}^u, \boldsymbol{\phi}^u\right)^{-1} \boldsymbol{z}+\alpha_v\boldsymbol{v}^T K_v(\boldsymbol{\phi}^{V}, \boldsymbol{\phi}^{V})^{-1} \boldsymbol{v}+\alpha_{v^o}|\boldsymbol{v}^{(2)}-\boldsymbol{V}^{o}|^2\\
 \text { s.t. } z_j^{(2)}=-\nu z_j^{(4)}+\boldsymbol{Q}^{(k)}(x_j, t_j)\cdot\boldsymbol{z}^{(3)}_j\\\quad\quad\quad\quad\quad-(L(x_j, t_j, \boldsymbol{Q}^{(k)}(x_j, t_j))+v^{(1)}_j+{F\!\left(m_{\mathrm{HJB},j}^{(k)}\right)}),\quad \forall j=1, \ldots, \mathfix{M_{\Omega}^{\mathrm{HJB}}},
 \\
\quad\quad z^{(1)}_j=U_T(x_j),\quad \forall j=\mathfix{M_{\Omega}^{\mathrm{HJB}}+1}, \ldots, \mathfix{M^{\mathrm{HJB}}}.
\end{cases}
\end{align}
\mathfix{The problem in \eqref{OptGPProb_ztv} is a convex quadratic optimization problem with linear equality constraints. If the feasible set is nonempty and the objective is strictly convex on it, the minimizer exists and is unique. The minimizer can be obtained by solving the linear system given by the Lagrange multiplier method. We omit the details.}

\textbf{Step 3.} The final step involves updating the policy. \mathfix{Let \(\chi:=\{(x_j,t_j)\}_{j=1}^{M_{\Omega}^{\mathrm{HJB}}}\) denote the interior HJB collocation points in \(\mathbb{T}^d \times (0, T)\).} We compute the values of the updated policy at the collocation points as follows:
\[
\mathbf{q}^{k+1,j}=\arg \max _{\|\boldsymbol{q}\| \leq \mathfix{R_{\mathrm{pol}}}}\left\{\boldsymbol{q} \cdot \boldsymbol{z}^{(3)}_j-L(x_j, t_j, \boldsymbol{q})\right\},  \quad \forall j = 1, \ldots, \mathfix{M_{\Omega}^{\mathrm{HJB}}}.
\]
Next, we approximate the policy function \(\boldsymbol{Q}^{(k+1)} \) by a GP and obtain the representer formula
\[
\boldsymbol{Q}^{(k+1)}(x,t)=\boldsymbol{K}_{\boldsymbol{Q}}((x,t),\chi)\,
\boldsymbol{K}_{\boldsymbol{Q}}(\chi,\chi)^{-1}\,
\mathbf{q}^{k+1},\quad\forall  (x,t)\in \mathbb{T}^d\times (0, T).
\]
{This process is repeated iteratively, successively refining the distribution, value function, and policy until a stopping criterion is met.}

\section*{Declarations}
\noindent\textbf{Conflict of interest.}
The authors declare that they have no conflict of interest.

\medskip
\noindent\textbf{Data availability.}
The authors confirm that the data supporting the conclusions of this study are contained in the article.

\bibliographystyle{plain}
\bibliography{reference1}
\end{document}